%% file: main.tex
\pdfoutput=1
\PassOptionsToPackage{shortlabels,inline}{enumitem}
\PassOptionsToPackage{hyphens}{url}
\documentclass{pinepaper}

\usepackage{marvosym} 
\usepackage{xspace}

\input{math_commands.tex}

\usepackage{multirow}
\usepackage{colortbl}
\colorlet{oursrow}{PineGreen!14}
\graphicspath{{figures/}}

\usepackage[capitalize]{cleveref}
\crefname{section}{Sec.}{Secs.}
\Crefname{section}{Section}{Sections}
\Crefname{table}{Table}{Tables}
\crefname{table}{Tab.}{Tabs.}
\makeatletter
\DeclareRobustCommand\onedot{\futurelet\@let@token\@onedot}
\def\@onedot{\ifx\@let@token.\else.\null\fi\xspace}
 
\def\ie{\emph{i.e}\onedot} 
 
 \def\vs{\emph{vs}\onedot}

\makeatother
\setlist[itemize]{noitemsep,leftmargin=*,topsep=0em}
\setlist[enumerate]{noitemsep,leftmargin=*,topsep=0em}

\newcommand{\method}{WAtcher}
\newcommand{\methodname}{\textit{\method}\xspace}
\newcommand{\guard}{\mbox{Safe-Guard}\xspace}

\DeclareMathOperator{\sg}{sg}

\newcommand{\para}[1]{\vspace{3pt}\noindent\textbf{#1}\hspace{0.5em}}
\newcommand{\finding}[1]{{\setlength{\fboxsep}{1pt}\colorbox{gray!20}{#1)}}\nobreak}
\newcommand{\appref}[2]{\ref{#1}}
\newcommand{\mainref}[2]{\ref{#1}}
\usepackage{array}
\usepackage{adjustbox}
\newcommand{\hd}[1]{\begin{tabular}[b]{@{}c@{}}#1\end{tabular}}
\newlength{\sonefw}%
\newlength{\sthreefw}%
\newcommand{\sthreelab}[2]{\parbox[c]{0.085\textwidth}{\raggedleft #1\\#2}}

\title{WAtcher: Watching Ahead for Safe Robotic\texorpdfstring{\\}{ } Manipulation with Efficient World-Action Models}

\author[1]{Ziyu Shan}
\author[1]{Xiao Ma}
\author[1]{Kangrui Yu}
\author[2]{Changjie Wang}
\authorbreak
\author[1]{Gaoyuan Wu}
\author[1]{Zaijia Yang}
\author[1]{Chuanrui Zhang}
\author[3]{Qirui Li}
\authorbreak
\author[3]{Yanbo Zhang}
\author[3]{Duomin Wang}
\author[1,\text{\Letter}]{Ziwei Wang}

\affil[1]{Nanyang Technological University}
\affil[2]{KTH Royal Institute of Technology}
\affil[3]{NVIDIA}

\contribution[\text{\Letter}]{Corresponding author}

\makeatletter
\newcommand\headerlogobox[1]{\pine@append\pine@headerlogos{\hspace{7mm}}{\raisebox{-0.5\height}{#1}}}
\renewcommand\pine@logo{\includegraphics[height=\pinelogoheight]{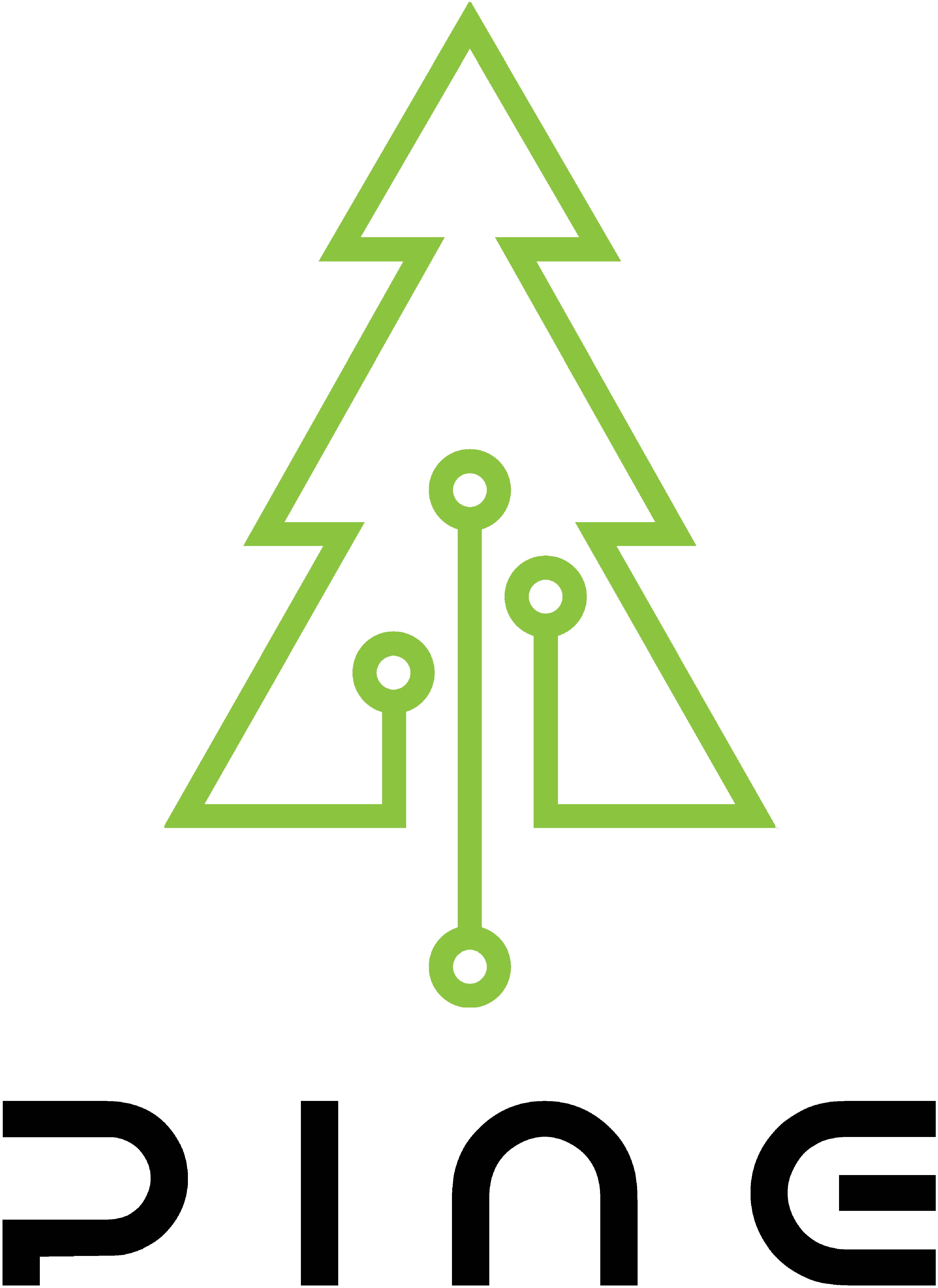}}
\makeatother
\headerlogobox{\includegraphics[height=10mm]{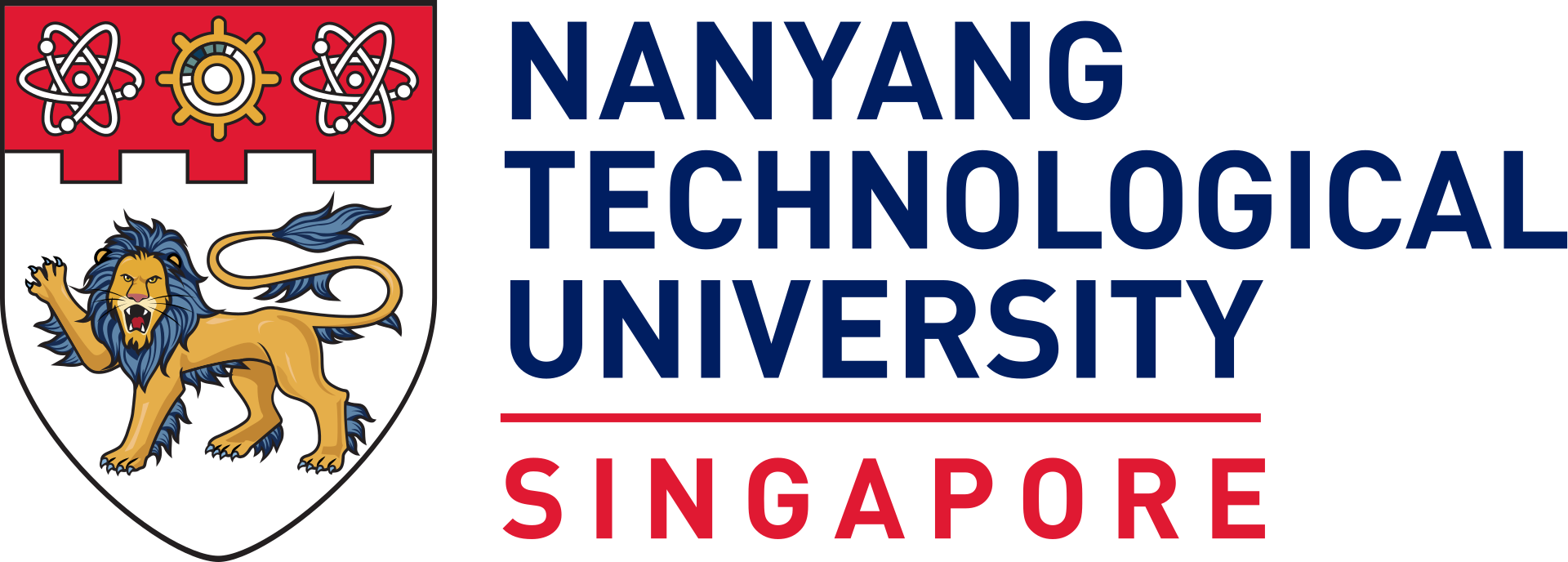}}
\headerlogobox{\includegraphics[height=12.5mm]{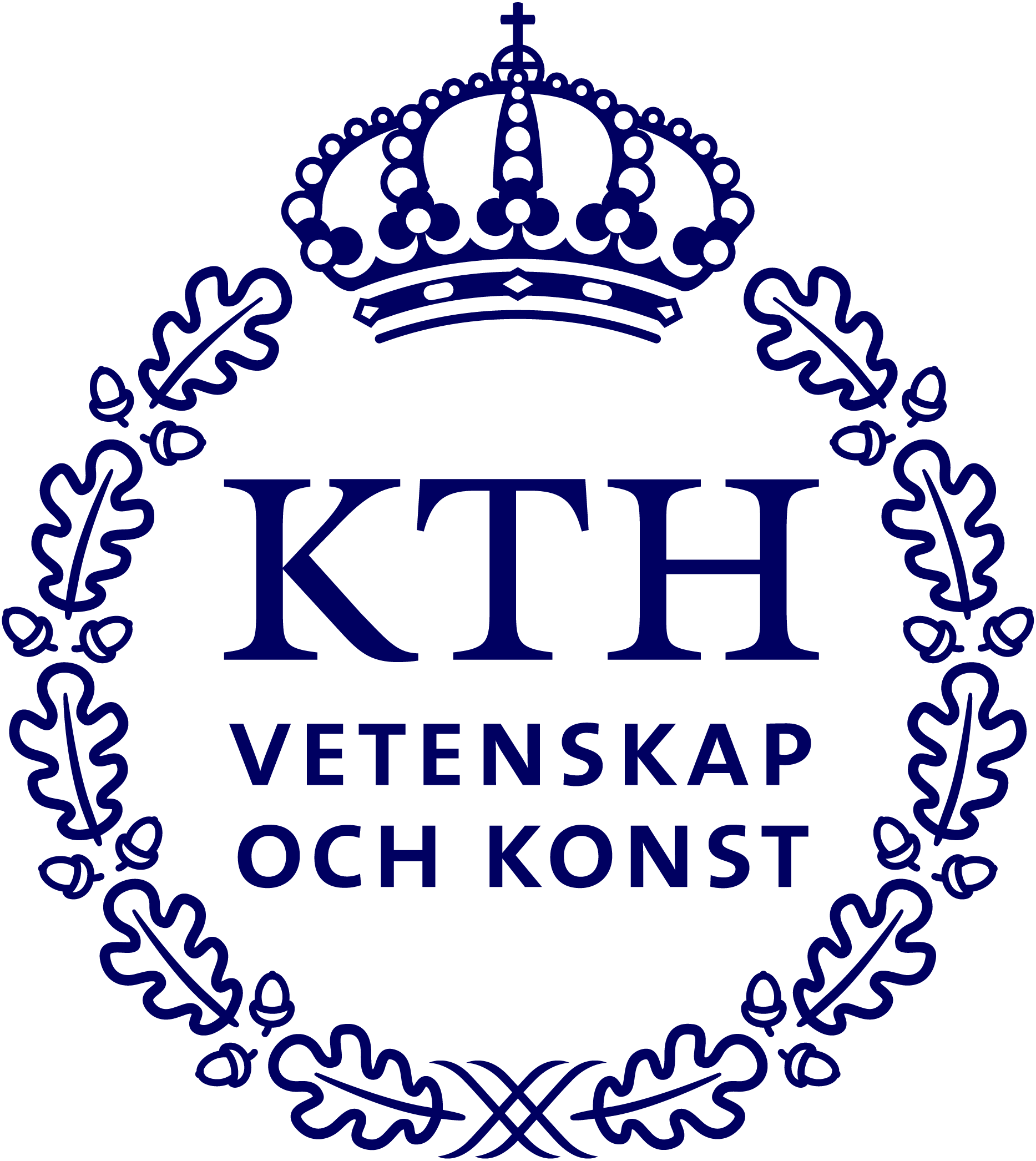}}
\headerlogobox{\includegraphics[height=6mm]{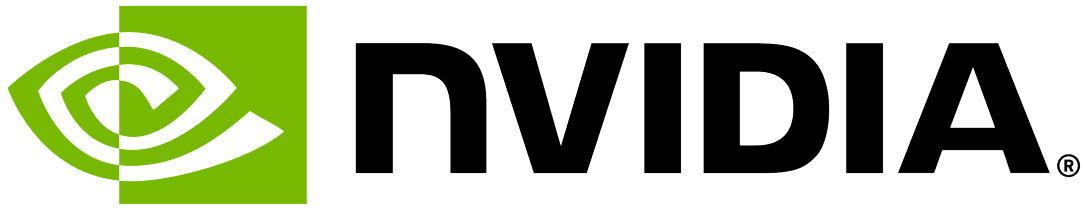}}

\projectpage{https://zyshan0929.github.io/WAtcher}
\checkdata[Code]{\url{https://github.com/maxiaovivi/WAtcher}}
\checkdata[HF Bench]{\url{https://huggingface.co/datasets/LevinShan/300Question-Bench}}

\begin{document}
\makepinetitle

\begin{pineabstract}
World-action models (WAMs) couple future imagination with action generation, but they execute every action chunk while ignoring the potential risk. To fill this gap, we present \methodname, a safety-aware world-action framework in which the policy imagines the outcome of the candidate chunk before execution. For fast future preview, \methodname builds on a compact WAM whose video expert imagines the visual outcome in only two denoising steps. In addition, we propose \guard, a lightweight module kept outside the WAM that scores each candidate chunk from its imagined outcome and triggers a bounded retreat when the risk is high. \guard is detachable and can be applied to other WAMs. To train \guard, we build an autonomous pipeline to collect hazard data from RoboTwin, LIBERO, and a real robot. Specifically, the pipeline perturbs expert demonstrations and labels each candidate chunk from the physical events recorded when it is executed, such as contacts and losses of support. Using the ${\sim}$25,000 labeled windows from this pipeline, \guard reaches 92.1\% balanced accuracy on a blind set, and cuts hazardous closed-loop trials on training tasks from 27.0\% to 10.8\% without lowering task success. On a held-out 300-question benchmark, it scores 94.2\%, against 72.5\% and 63.3\% for zero-shot GPT-6-Astra and Claude Opus 5.5. \methodname also reaches 90.4\% and 82.4\% average success on RoboTwin~2.0 and ManiSkill3, on par with or above leading WAMs, and 74.0\% on five real-world tasks with \guard on, against 59.5\% for $\pi_{0.5}$.
\end{pineabstract}

\begin{figure}[t]
\centering
\includegraphics[width=\linewidth]{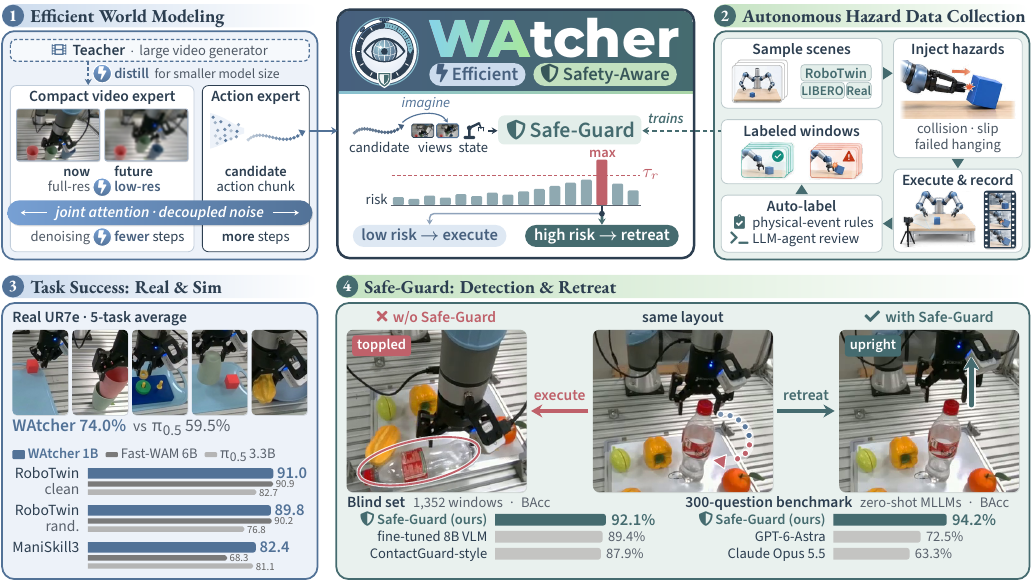}
\caption{\textbf{\methodname at a glance.} (1)~The host's compact video expert (architecture of~\cite{li2026efficientwam}) imagines low-resolution futures (illustrative). (Center)~\guard passes candidates whose imagined outcome has risk below $\tau_r$; else the robot retreats. (2)~Pipeline-labeled hazards train \guard (illustrative icons). (3)~Task success of \methodname vs.\ Fast-WAM (our re-run)~\cite{yuan2026fastwam} and $\pi_{0.5}$ (third-party on RoboTwin)~\cite{intelligence2025pi05}; UR7e: 5-task average. (4)~\guard prevents toppling on the UR7e and leads on the blind set and the 300-question benchmark.}
\label{fig:teaser}
\end{figure}

\section{Introduction}
\label{sec:intro}
World-action models (WAMs) unify future forecasting and action generation in a shared generative model and have become a strong paradigm for robotic manipulation~\cite{ye2026dreamzero,ye2026gigaworldpolicy,qiu2026vid2wam}. However, mainstream WAMs execute every action chunk they generate without a check. The training demonstrations also rarely contain hazardous outcomes such as collisions, slips, or dropped objects, and the training objectives never supervise risk. Whether a chunk is hazardous depends jointly on the scene, the robot state, and the commanded motion, so the same chunk can pass safely in one layout and collide in another. Detecting risk before execution requires anticipating the outcome of each candidate, at every decision step and within the latency budget of the policy.

Our key observation is that \textit{WAMs of a common design already contain the forward model that such a safety check needs.} When the video and action experts share joint attention and are trained with decoupled noise levels and a video loss that reaches the action tokens~\cite{zhu2025uwm,li2026efficientwam}, the video expert can take a fixed candidate chunk as a clean condition and imagine a coarse outcome, as in the forward mode of Unified World Model (UWM)~\cite{zhu2025uwm}, so no additional world model has to be trained or used. To make this imagination fast enough for real-time use, \methodname builds on an efficient architecture~\cite{li2026efficientwam}, a compact WAM trained in this way, whose video expert, distilled from a large video generator, imagines a candidate's low-resolution outcome in only two denoising steps. This fast future preview provides a solid basis for a pre-execution risk check in practical deployment.

On top of this imagination, \methodname places \guard, a lightweight risk head, between the policy and the robot (\cref{fig:teaser}). \guard fuses the current views, the robot state, every command of the candidate, and its imagined outcome with a small Transformer, scores the action chunk, and executes low-risk chunks while replacing high-risk ones with a bounded retreat. Unlike monitors that read policy features~\cite{gu2025safe} or train a separate world model~\cite{zheng2026contactguard,wu2025forewarn}, \guard reuses the policy's own video expert and learns from physical events recorded in executed hazards rather than from task failure~\cite{duan2024aha,dai2025racer}. Because it is trained separately and leaves the policy weights unchanged, \guard can serve other WAMs. With Fast-WAM-Joint~\cite{yuan2026fastwam}, the variant of Fast-WAM whose video and action tokens attend to each other, it also lowers hazardous executions compared with that WAM alone (\cref{sec:exp_closed}).

To train \guard, we build an autonomous pipeline that perturbs expert trajectories in RoboTwin~2.0~\cite{chen2025robotwin2}, LIBERO~\cite{liu2023libero}, and on a real UR7e. The pipeline executes every candidate and labels it from recorded physical events, with an LLM agent reviewing ambiguous cases and a human verifying the real-robot labels. It constructs four hazard types: unintended contacts (e.g., collisions), losses of support (e.g., slips, failed hanging), generic motion errors, and semantic dangers (e.g., a knife moving toward a hand) that no contact threshold captures, yielding ${\sim}$25,000 labeled windows. Trained on this dataset, \guard reaches 92.1\% balanced accuracy on a blind set of 1,352 windows, ahead of a fine-tuned 8B vision-language model (VLM; 89.4\%) and a ContactGuard-style monitor (87.9\%). It also reduces hazardous closed-loop trials in new scenes of training tasks from 27.0\% to 10.8\%, against 15.3\% for the ContactGuard-style monitor at the same rejection rate, without lowering task success rates. 

For a fair comparison with zero-shot multimodal large language models (MLLMs), we also establish a held-out 300-question benchmark that covers every hazard type in RoboTwin and LIBERO. It also includes 101 same-state pairs that differ only in the candidate. On the 300-question benchmark, our lightweight \guard scores 94.2\%, against 72.5\% and 63.3\% for zero-shot GPT-6-Astra and Claude Opus 5.5. 
On simulation suites, our \methodname reaches 90.4\% and 82.4\% average success on RoboTwin~2.0 and ManiSkill3, on par with or above leading WAMs such as Fast-WAM~\cite{yuan2026fastwam} and LingBot-VA~\cite{li2026lingbotva}.
On a UR7e robot platform, it reaches 74.0\% on five real-world tasks with \guard on, against 59.5\% for $\pi_{0.5}$. For safety on the real robot, \guard, adapted with 258 human-verified windows, reduces staged hazards from 43 to 6 out of 60.

Our contributions are summarized as follows:
\begin{itemize}[leftmargin=*,itemsep=2pt,topsep=2pt]
\item We propose an autonomous pipeline that collects and labels hazardous executions in two simulators (\ie, RoboTwin and LIBERO) and on a real robot, with a comprehensive 300-question benchmark and a larger-size blind set for fair evaluation.
\item We propose \guard, a detachable risk head that scores each action candidate from the observations, commands, and the outcome imagined by the WAM's own video expert in 32\,ms. The head can also be applied to another WAM.
\item Based on the collected data and the \guard, we propose \methodname, which runs a risk check before every chunk, retreats from risky ones, and reduces hazardous executions in simulation and on a UR7e robot while preserving original task success.
\end{itemize}

\section{Related Work}
\label{sec:related}

\para{World-action models.}
Visual foresight plans with action-conditioned video prediction~\cite{finn2017foresight,ebert2018visualforesight}, and later policies recover actions from generated video plans~\cite{du2023unipi,ko2024avdc}, pretrain on video to predict frames and actions~\cite{wu2024gr1,cheang2024gr2}, or act on forecasted visual states~\cite{tian2025seer,hu2025vpp}. PAD, UVA, and UWM couple video and action diffusion in one model~\cite{guo2024pad,li2025uva,zhu2025uwm}, and CoT-VLA, WorldVLA, and DreamVLA add visual forecasts to vision-language-action models~\cite{zhao2025cotvla,cen2025worldvla,zhang2025dreamvla}. Recent world-action models (WAMs) build on large pretrained video generators~\cite{wan2025wan,nvidia2025cosmos}: Genie Envisioner, VideoVLA, Video Policy, Cosmos Policy, and mimic-video turn video foundation models into policies~\cite{liao2026genie,shen2025videovla,liang2025videopolicy,kim2026cosmospolicy,pai2026mimicvideo}, and Motus, LingBot-VA, DiT4DiT, and DreamZero jointly model future video and actions~\cite{bi2025motus,li2026lingbotva,ma2026dit4dit,ye2026dreamzero}. World models can also choose among candidates: Cosmos Policy plans with predicted futures and values~\cite{kim2026cosmospolicy}, and FOREWARN verifies predicted outcomes of sampled plans with a vision-language model~\cite{wu2025forewarn}. \guard instead scores each candidate with a head trained on executed hazards. Action-conditioned video models also evaluate policies offline~\cite{li2025worldeval,quevedo2025worldgym,guo2025ctrlworld}.

\para{Efficient world-action models.}
The video branch dominates the cost of a WAM~\cite{shan2026dvg}. Some WAMs learn from video but act without generating it at test time~\cite{li2025uva,yuan2026fastwam,ye2026gigaworldpolicy,collins2026tapestry,zhang2026w2a,li2026lightwam}. Others make imagination cheaper by distilling a large video model into a small expert that predicts reduced latents in few steps~\cite{li2026efficientwam,qiu2026vid2wam}, by one-step distillation~\cite{akbari2026flashwam}, or by single-pass foresight~\cite{yan2026svam}. None of them checks its actions before execution, and \methodname reuses its video expert to imagine and check each candidate.

\para{Safety and recovery.}
Recovery RL hands control to a recovery policy when a learned safety critic flags a proposed action~\cite{thananjeyan2021recoveryrl}, and SafeVLA aligns vision-language-action policies with safety constraints through constrained reinforcement learning~\cite{zhang2025safevla}. SAFE predicts task failure from the internal features of vision-language-action models~\cite{gu2025safe}. Sentinel, AHA, and Code-as-Monitor detect failures with vision-language models during or after execution~\cite{agia2024sentinel,duan2024aha,zhou2025codeasmonitor}, RACER learns recovery from perturbed demonstrations~\cite{dai2025racer}, and latent safety filters screen actions in the latent space of a learned world model~\cite{nakamura2025latentsafety}. In addition, ContactGuard predicts visual embeddings for a planned chunk with a separately trained latent world model and aborts the chunk before contact when a failure probe fires~\cite{zheng2026contactguard}. Like ContactGuard, \guard reads no policy features and leaves the policy unchanged. Unlike ContactGuard, it lets the policy's own video expert imagine the outcome instead of training a separate world model, covers loss of support as well as contact, and learns from hazards that our pipeline generates and labels. Appendix Sec.~\appref{app:compare}{D.5} compares further monitors.

\section{Method}
\label{sec:method}

\begin{figure}[t]
\centering
\includegraphics[width=\linewidth]{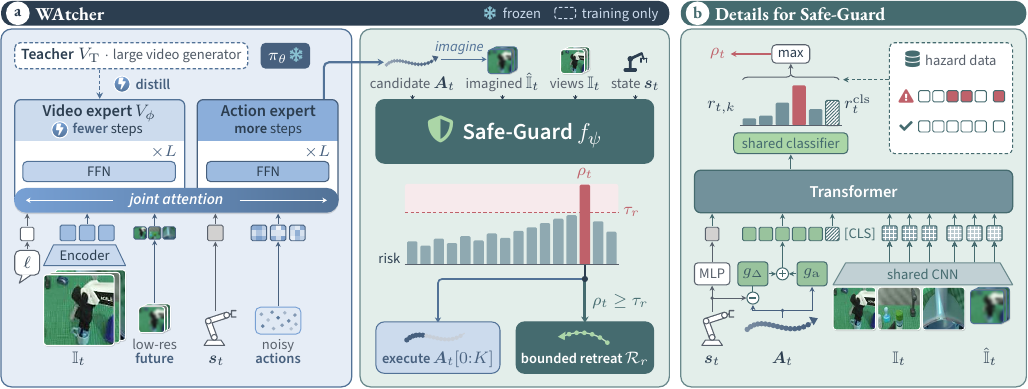}
\caption{\textbf{Overview of \methodname.} (a)~A compact video expert, distilled from a large video model, encodes the instruction and observation and imagines future latents; the action expert reads them with the state through joint attention and proposes a candidate chunk. \guard, a separately trained head, reads the candidate, its frames $\hat{\sI}_t$ imagined in a second pass of the frozen video expert, the current views, and the state, and predicts step- and chunk-level risk, then executes the chunk or retreats. (b)~\guard in detail. The policy stays unchanged.}
\label{fig:overview}
\end{figure}

\subsection{Overview and Problem Setup}
\label{sec:setup}

In \cref{fig:overview}, \methodname pairs a compact world-action policy, whose video expert imagines the future for its action expert (\cref{sec:policy}), with \guard, a head that predicts the execution risk of each candidate from its imagined outcome (\cref{sec:guard}) and acts on it (\cref{sec:dispatch}). An autonomous pipeline supplies the risk labels for training (\cref{sec:data}).

At decision time $t$, let $\sI_t$ denote the current views, $\vs_t$ the robot state, and $\ell$ the instruction. The pretrained world-action policy proposes a candidate action chunk
\begin{equation}
\label{eq:policy}
\mA_t = \pi_\theta(\sI_t, \vs_t, \ell; \xi_t) \in \R^{H \times d_a},
\end{equation}
where $\xi_t$ is sampling randomness, $H$ is the chunk length, and the rows $\va_{t,1}, \dots, \va_{t,H} \in \R^{d_a}$ of $\mA_t$ are its commands. Let $\hat{\sI}_t(\mA_t)$ denote the future frames that the video expert imagines for this candidate (\cref{sec:policy}). \guard maps the current views, the imagined frames, the state, and the candidate to step-level risk scores and a chunk-level risk score in $[0, 1]$ (defined in \cref{eq:risk}),
\begin{equation}
\label{eq:guard}
\big(\{r_{t,k}\}_{k=1}^{H},\, \rho_t\big) = f_\psi\big(\sI_t, \hat{\sI}_t(\mA_t), \vs_t, \mA_t\big),
\end{equation}
where $\psi$ denotes its parameters. \guard reads no policy features and does not alter $\mA_t$.

\subsection{Compact World-Action Policy}
\label{sec:policy}

The world-action policy, which we call the host policy, follows an efficient WAM architecture~\cite{li2026efficientwam}: a large video generator is distilled into a compact video expert, which then stays frozen while an action expert is trained with it through joint attention. The video expert predicts the latents of future frames from those of the current views, and the action expert reads them to generate continuous actions. The policy is trained on task demonstrations without any risk supervision, and risk detection is left to the detachable \guard.

\para{Distilled video expert.}
\methodname distills the predictive knowledge of a large video generator into a compact video expert $V_\phi$~\cite{hinton2015distilling}. The student is a compact video transformer that predicts the latents of future frames from those of the current views, conditioned on the instruction and the robot state. The teacher $V_{\mathrm{T}}$ is Wan2.2-TI2V-5B~\cite{wan2025wan}, adapted to the robot data with a state token. Given the same conditioning $\vc_t = (\sI_t, \vs_t, \ell)$ and the same noisy future latents $\vz_\sigma$ at noise level $\sigma$, the student is trained to match the teacher's prediction and selected intermediate features:
\begin{equation}
\label{eq:distill}
\begin{split}
\mathcal{L}_{\mathrm{distill}} = {} & \E_{\sigma, \vz_\sigma}\big\| V_\phi(\vz_\sigma, \sigma; \vc_t) \\
& - \sg\big(V_{\mathrm{T}}(\vz_\sigma, \sigma; \vc_t)\big) \big\|_2^2 + \lambda_{\mathrm{feat}}\,\mathcal{L}_{\mathrm{feat}},
\end{split}
\end{equation}
where $\sg$ denotes stop-gradient, $\mathcal{L}_{\mathrm{feat}}$ aligns intermediate features, and $\lambda_{\mathrm{feat}}$ balances the two terms (Appendix Sec.~\appref{app:impl}{A}).

\para{Future resolution, denoising budget, and action expert.}
The current views are tiled into one frame, and future frames are imagined at a lower resolution, which reduces the number of future tokens. The video expert uses only two denoising steps, a budget supported by the empirical study in Appendix Sec.~\appref{app:policy-denoise}{F.3}. 
Additionally, the action expert is trained to generate continuous actions by flow matching, as in $\pi_0$~\cite{black2024pi0}. The first $K \le H$ commands of an accepted chunk are executed before the next proposal.

\para{Candidate-conditioned imagination.} In our design, joint attention is bidirectional, so the video tokens also attend to the action tokens, and the video expert can imagine the outcome of a given chunk. Since the video expert is frozen, the action stream learns this conditioning: as in~\cite{li2026efficientwam}, video and actions receive independent noise levels in action-expert training, and a low-weight video flow-matching loss teaches the action tokens' keys and values to inform the future latents (Appendix Sec.~\appref{app:impl}{A}). For a candidate $\mA$, we fix the action tokens to $\mA$ at noise level zero and denoise only the future latents in the same two steps, as in UWM's forward mode~\cite{zhu2025uwm}; decoding the last imagined frame of each view gives $\hat{\sI}_t(\mA)$. Every candidate that \guard scores is imagined in this way, in training and in deployment.

\subsection{\guard: Pre-Execution Risk Detection}
\label{sec:guard}

\guard predicts whether executing a candidate chunk from the current state will cause a hazardous physical event. It is a lightweight separate network that is trained, removed, or replaced without touching the policy, and it needs only a model that imagines a candidate's outcome. It fuses visual, state, action, and global tokens with a small Transformer and reads out one risk score per command and one for the whole chunk. Unlike prior monitors, it takes its forward model from the policy itself and is trained on recorded physical events rather than task failure (Appendix Sec.~\appref{app:compare}{D.5}).

\para{Visual tokens.}
A small CNN shared across the head and wrist views gives a feature map per view, and pooling it to a coarse grid keeps the layout and yields a few tokens per view. The imagined frames share the CNN and pooling and add their own tokens.

\para{State and action tokens.}
An MLP embeds the robot state $\vs_t$ as one token. Each command $\va_{t,k}$ is embedded by two MLPs whose outputs are summed,
\begin{equation}
\label{eq:action_token}
\ve_{t,k} = g_{\mathrm{a}}(\va_{t,k}) + g_{\Delta}(\va_{t,k} - \vs_t), \qquad k = 1, \dots, H,
\end{equation}
where $g_{\mathrm{a}}$ encodes the command and $g_{\Delta}$ its displacement from the current state, which gives one token per command.

\para{Global token and fusion.}
A learnable \texttt{[CLS]} token completes the token sequence, and a small Transformer encoder fuses all tokens. Each action token yields the risk score of its own command, so the step scores can localize the hazard in time. The \texttt{[CLS]} token belongs to no single command and summarizes the chunk as a whole. We add it because some hazards are not visible in any single command, such as a slip that builds up over the chunk or a push that moves an object off its support only together with later commands. Removing it lowers the share of hazards detected on the 300-question benchmark from 94.3\% to 91.7\% (Appendix Table~\appref{tab:ablation}{S17}).

\para{Risk readout.}
A classifier shared across positions, layer normalization followed by a linear map to two classes (\emph{safe} and \emph{risk}), is applied to each action token and to the \texttt{[CLS]} token. Let $r_{t,k}$ be the softmax risk probability at command $k$ and $r^{\mathrm{cls}}_t$ the one at the \texttt{[CLS]} token. The chunk risk is their maximum,
\begin{equation}
\label{eq:risk}
\rho_t = \max\big(\max\nolimits_{k}\, r_{t,k},\; r^{\mathrm{cls}}_t\big),
\end{equation}
so that a single risky command suffices to flag the chunk, and the step scores localize the hazard within it.

\para{Training.}
Each training window pairs the current views, the imagined frames of its candidate, the state, and the candidate with a chunk label $y_t \in \{0, 1\}$ and step labels $y_{t,k}$ that mark the commands inside the annotated hazardous interval (\cref{sec:data}). A candidate can contain benign motion together with a short hazardous interval, so we combine chunk-level and step-level supervision,
\begin{equation}
\label{eq:loss}
\mathcal{L} = \mathcal{L}_{\mathrm{chunk}} + \lambda_s\, \mathcal{L}_{\mathrm{step}},
\end{equation}
where both terms are cross-entropy losses and $\mathcal{L}_{\mathrm{step}}$ averages over the commands whose step labels are resolved by the recorded events. An auxiliary cross-entropy supervises the \texttt{[CLS]} score. Training updates only $\psi$, excluding $\theta$.

\subsection{Risk-Aware Execution and Bounded Recovery}
\label{sec:dispatch}

\methodname intervenes when the chunk score reaches a selected threshold $\tau_r$:
\begin{equation}
\label{eq:gate}
g_t = \1\left[\rho_t \ge \tau_r\right].
\end{equation}
If $g_t = 0$, \guard executes the candidate prefix $\mA_t[0{:}K]$ with $K \le H$; otherwise it invokes a robot-specific recovery controller $\mathcal{R}_r(\vs_t, \vs^{\mathrm{ref}}_t)$, as in Recovery RL~\cite{thananjeyan2021recoveryrl}, and a fresh observation then conditions the next proposal.

Recovery moves toward an available reference with bounded target increments and at most $J_r^{\max}$ commands. For arm-position targets $\vq^{\mathrm{rec}}_j$,
\begin{equation}
\label{eq:retreat}
\big\| \vq^{\mathrm{rec}}_{j+1} - \vq^{\mathrm{rec}}_j \big\|_\infty \le \epsilon_r .
\end{equation}
In simulation, recovery holds the gripper and moves back toward the start of the previously accepted chunk, and three consecutive rejections trigger a protective stop. The details are in Appendix Sec.~\appref{app:closed}{G}.

\section{Autonomous Hazard Data Collection}
\label{sec:data}

We build an autonomous pipeline that creates candidate chunks, executes them, and labels each window from what happened during execution, in RoboTwin~2.0~\cite{chen2025robotwin2} (described here), in LIBERO~\cite{liu2023libero}, and on the real UR7e, where a human verifies the labels (Appendix Sec.~\appref{app:labels-other}{B.4}). It runs in four stages (scene sampling, action intervention, batch execution, and evidence recording), followed by labeling (\cref{fig:data_pipeline}); Appendix Sec.~\appref{app:labels}{B} details the data construction. Every label rests on a recorded event, not on the intervention that produced the candidate or on task failure.

\begin{figure}[t]
\centering
\includegraphics[width=\linewidth]{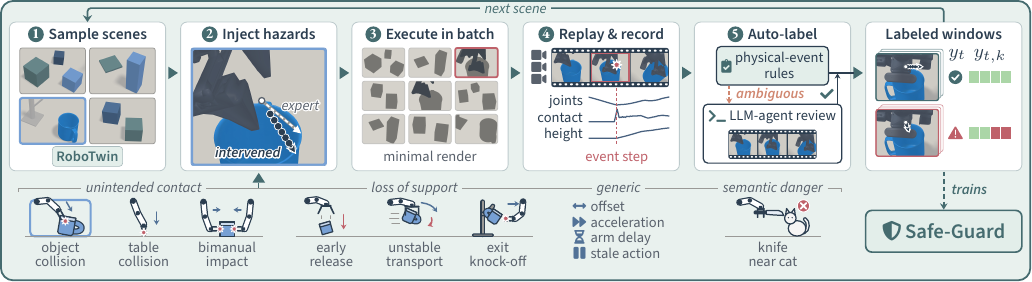}
\input{figures/data-pipeline-src/caption.tex}
\end{figure}

\para{Scene sampling.}
The pipeline samples the task, object placements, and seed of each scene and restores the matching expert trajectory and snapshots, so each candidate starts from a saved expert state.

\para{Action intervention.}
From a sampled expert state, the pipeline modifies the upcoming chunk: generic perturbations offset, accelerate, delay one arm, or hold a stale action, and targeted constructions (Appendix Table~\appref{tab:hazardsfull}{S2}) deepen the approach into the object, lower the end-effector target into the table, or widen the gripper slightly in transport so that the object slips. Semantic constructions move a knife toward a person's hand or a sharp object toward an animal model.

\para{Batch execution and evidence recording.}
Candidates run in parallel shards with minimal rendering for convenient evaluation. One that hits an event is replayed from its saved state with full rendering, which records the views, joint states, contacts, object heights, and event steps.

\para{Labeling.}
A window is labeled risky when its execution records one of the hazard types of \cref{fig:data_pipeline} (bottom): \finding{1} an \emph{unintended contact} of the robot or a held object with the table, another object, or the robot itself; \finding{2} a \emph{loss of support}, where an object drops after an early release, a slip, or a missed hanging; \finding{3} a \emph{generic} execution error from an offset, acceleration, arm delay, or stale action, labeled by the same events or a motion-limit violation; or \finding{4} a \emph{semantic danger}, such as a knife touching a person's hand or a sharp object coming within 5\,cm of an animal model. The event step localizes the hazard within the chunk, and the thresholds are given in Appendix Sec.~\appref{app:labels}{B}. Near-contact motion, intended grasps, and task failure alone are not hazards. 

An LLM coding agent (GPT-6-Astra) reviews ambiguous windows, and unresolved ones are removed. In a blind audit, two annotators agreed with the labels on 95.7\% of 300 random windows and 97.3\% of the 300-question benchmark.

\para{Dataset.}
The dataset spans bimanual and single-arm tasks in simulation and on the real UR7e robot, and holds 25,434 labeled windows, 7,392 of them risky (Appendix Sec.~\appref{app:labels-constructions}{B.1}). Training and validation use non-overlapping scenes, and no scene, seed, or source group of the 300-question benchmark appears in either to ensure rigorous evaluation.

\section{Experiments}
\label{sec:exp}

\subsection{Experimental Setup}
\label{sec:exp_setup}

\begin{figure}[t]
\centering
\includegraphics[width=0.7\linewidth]{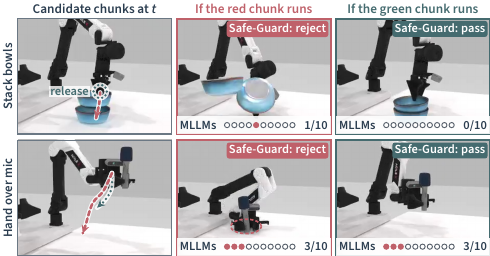}
\caption{\textbf{Hazardous (red) and safe (green) candidates, one RoboTwin state per row}, chosen to show MLLM failures. Left: recorded gripper paths (side view no model reads). Middle, right: outcomes 0.64\,s later; filled dots: MLLMs flagging a hazard. One MLLM flags the red bowl chunk; three flag both mic chunks. More cases are in Appendix Figs.~\appref{fig:galA}{S1}--\appref{fig:galB}{S2}.}
\label{fig:t300}
\end{figure}

\para{Baselines.}
For risk detection, we compare two kinds of detectors. \finding{1} The \emph{zero-shot MLLMs} are ten multimodal frontier models, evaluated on the 300-question benchmark (Appendix Sec.~\appref{app:judges}{C.6}); \cref{tab:blindmain} lists four of them. They receive the views, the state, and the candidate with the meaning, units, and timing of every dimension, but no imagined frames, drawn gripper path, hazard rules, or examples, and a hazard probability of at least 0.5 counts as a detection. 
\finding{2} The baseline \emph{monitors on our data} in \cref{tab:blindmain} are a state-and-candidate MLP, a SAFE-style probe on policy features~\cite{gu2025safe}, a ContactGuard-style monitor on predicted latent future embeddings~\cite{zheng2026contactguard}, Qwen3-VL-2B and -8B~\cite{bai2025qwen3vl} fine-tuned with LoRA~\cite{hu2022lora} (the 8B model reads our imagined frames and the head view with the candidate's gripper path drawn on it, as a FOREWARN-style verifier~\cite{wu2025forewarn}), a FAIL-Detect-style score that needs no failure data~\cite{xu2025faildetect}, and a forward-kinematics collision check with privileged geometry. Full details of these monitors are in Appendix Sec.~\appref{app:detect}{D}.

\para{Implementation details.} The policy has 1.0B parameters: a 0.8B video expert distilled from Wan2.2-TI2V-5B~\cite{wan2025wan} and a 0.2B action expert. It proposes chunks of $H = 16$ commands and executes $K = 16$ in simulation and 8 on the UR7e. It imagines $160 \times 192$ future frames from a $320 \times 384$ tiled input in 2 denoising steps and samples actions in 10 Euler steps. \guard has 1.7\,M parameters, fuses 114 tokens of width 128 with a three-layer Transformer, and is trained with $\lambda_s = 1$. Recovery uses $\epsilon_r = 0.02$\,rad and $J_r^{\max} = 16$. All settings, including the \guard checkpoint used in each experiment, are in Appendix Sec.~\appref{app:impl}{A}.

\para{Metrics.}
We report success rate (SR) for simulation and real-world experiments. For risk detection on the blind set and 300-question benchmark in \cref{sec:exp_bench}, following SAFE and Rewind-IL~\cite{gu2025safe,zheng2026rewindil}, we report the share of hazards detected (TPR), the share of safe candidates accepted ($\text{TNR} = 1 - \text{FPR}$), and their mean, the balanced accuracy (BAcc), which does not depend on how many candidates are hazardous. Accuracy and AUROC are in Appendix Sec.~\appref{app:bench-overall}{C.2}.

\subsection{Risk Detection}
\label{sec:exp_bench}

\begin{table}[t]
\caption{Risk detection (\%). 300-Q: 300-question benchmark. Blind set: 1,352 windows, 613 hazardous. Unseen: four unseen tasks. Gray: not run on these full splits. BAcc: three-seed mean for learned detectors, single run otherwise; TPR, FPR: selected seed.}
\label{tab:blindmain}
\centering
\footnotesize
\setlength{\tabcolsep}{2.6pt}
\begin{tabular}{l c ccc c}
\toprule
& 300-Q & \multicolumn{3}{c}{Blind set} & Unseen \\
\cmidrule(lr){3-5}
Detector & BAcc & BAcc & TPR & FPR & BAcc \\
\midrule
\multicolumn{6}{l}{\textit{Zero-shot MLLMs}} \\
Claude Opus 5.5 & 63.3 & \multicolumn{4}{c}{\cellcolor{gray!20}} \\
Gemini 4 Argon & 70.0 & \multicolumn{4}{c}{\cellcolor{gray!20}} \\
GPT-6-Astra & 72.5 & \multicolumn{4}{c}{\cellcolor{gray!20}} \\
GPT-6.1-Sol & 73.0 & \multicolumn{4}{c}{\cellcolor{gray!20}} \\
\midrule
\multicolumn{6}{l}{\textit{Monitors on our data}} \\
FAIL-Detect-style & 65.4 & 63.8 & 39.2 & 10.8 & 56.7 \\
Collision check (FK) & 75.3 & 72.2 & 48.3 & 3.9 & -- \\
State + candidate MLP & 78.7 & 74.4 & 70.3 & 21.0 & 60.6 \\
SAFE-style probe~\cite{gu2025safe} & 83.5 & 81.8 & 78.6 & 15.3 & 66.4 \\
Fine-tuned Qwen3-VL-2B & 86.8 & 84.4 & 82.1 & 13.0 & 69.2 \\
ContactGuard-style~\cite{zheng2026contactguard} & 89.6 & 87.9 & 86.1 & 10.7 & 70.8 \\
Fine-tuned Qwen3-VL-8B & 91.1 & 89.4 & 88.3 & 9.3 & 73.7 \\
\rowcolor{oursrow} \guard (ours) & \textbf{94.2} & \textbf{92.1} & \textbf{92.5} & 8.1 & \textbf{75.9} \\
\bottomrule
\end{tabular}
\end{table}
We test risk detection on the 300-question benchmark (Appendix Sec.~\appref{app:bench}{C}), whose scenes, seeds, and source groups are separate from the training and validation windows of our dataset (\cref{sec:data}). Each question gives the current views, the robot state, and a candidate chunk and asks whether executing it records a hazardous event within the chunk (about 1\,s) or a fall within 0.5\,s after a release. It is small enough for zero-shot MLLMs; to test on more windows that no detector has seen, we also evaluate every learned detector on the blind set of 1,352 new windows (\cref{tab:blindmain}).

\noindent\textbf{Results on the 300-question benchmark.} \guard averages 94.2\% BAcc (\cref{tab:blindmain}), and its selected seed detects 94 of 100 hazards at 10 false alarms on 200 safe candidates (94.5\% BAcc). The ten zero-shot MLLMs reach at most 73.0\%. They handle hazards explicit in the commands but mostly miss collisions commanded as bimanual joint targets (\cref{fig:t300}) and mug--rack contacts, and on knife--hand questions react to the knife rather than the hand's position. Of the 26 hazards that at least eight MLLMs miss, \guard detects 22. With richer inputs and one example per category, the best MLLM reaches 81.0\% (Appendix Sec.~\appref{app:judges}{C.6}).

\noindent\textbf{Results on the blind set.} \guard averages 92.1\% BAcc (\cref{tab:blindmain}), against 89.4\% for the fine-tuned 8B VLM and 87.9\% for the ContactGuard-style monitor; its selected seed leads by 2.7 and 4.5 BAcc points. More results are in Appendix Sec.~\appref{app:detect-blind}{D.2}. 

We also evaluate on four larger splits, including the policy's own proposals, and on held-out hazard families to test generalization (Appendix Sec.~\appref{app:detect-splits}{D.3}).

\noindent\textbf{What the Imagined Future Contributes.}
\label{sec:exp_ablation}
We ask whether \guard needs the imagination and whether its source matters, retraining the head with each source of future frames and comparing them on a validation split. Future prediction benefits mainly when it shows the candidate's outcome: the candidate-conditioned future adds 4.8 BAcc points over the current views alone, whereas the nominal future, imagined for the policy's own actions, adds only 1.1. The gain is largest for losses of support (+7.9 vs.\ +2.6 for contact), because a slip or a fall shows only in the outcome, whereas most contacts are implied by the commanded pose. As for the source, the policy's own video expert gives a higher BAcc than a separately trained world model of the same architecture (89.7\% vs.\ 87.6\%). Imagining each candidate's outcome is thus the key ingredient, and the policy's own expert provides it without another model (Appendix Sec.~\appref{app:imagine}{E}).

\subsection{Closed-Loop Intervention}
\label{sec:exp_closed}

\begin{table}[t]
\caption{Closed loop on four RoboTwin tasks (\%; 20 trials per task). Monitors and random rejection reject as many chunks as \guard; the perfect detector rejects exactly the hazardous ones.}
\label{tab:closed}
\centering
\footnotesize
\setlength{\tabcolsep}{4pt}
\renewcommand{\arraystretch}{0.92}
\begin{tabular}{l cc cc}
\toprule
& \multicolumn{2}{c}{Clean} & \multicolumn{2}{c}{Random} \\
\cmidrule(lr){2-3} \cmidrule(lr){4-5}
Execution & Hazard$\downarrow$ & SR$\uparrow$ & Hazard$\downarrow$ & SR$\uparrow$ \\
\midrule
Policy only & 21.3 & 86.3 & 27.5 & 78.8 \\
+ random rejection & 18.8 & 85.0 & 25.0 & 77.5 \\
+ ContactGuard-style & 11.3 & 87.5 & 15.0 & 81.3 \\
+ \guard w/o future & 12.5 & 87.5 & 16.3 & 80.0 \\
+ \guard & \textbf{6.3} & \textbf{88.8} & \textbf{10.0} & \textbf{82.5} \\
\midrule
+ perfect detector & 2.5 & 90.0 & 3.8 & 85.0 \\
\bottomrule
\end{tabular}
\end{table}

\begin{table}[t]
\caption{Average success (\%) on the 50 RoboTwin~2.0 tasks (clean, randomized, mean). Params exclude text encoders. Best in \textbf{bold}, second \underline{underlined} (w/o-\guard row not ranked). Unmarked baselines: published results.}
\label{tab:manip}
\centering
\footnotesize
\setlength{\tabcolsep}{4pt}
\renewcommand{\arraystretch}{0.92}
\begin{tabular}{l cccc}
\toprule
Method & Clean & Random & Avg. & Params \\
\midrule
$\pi_0$$^\dagger$~\cite{black2024pi0} & 65.9 & 58.4 & 62.2 & 3.3B \\
$\pi_{0.5}$$^\dagger$~\cite{intelligence2025pi05} & 82.7 & 76.8 & 79.8 & 3.3B \\
OpenVLA-OFT$^\ast$~\cite{kim2025openvlaoft} & 15.6 & 14.7 & 15.2 & 7B \\
StarVLA-$\alpha$ (2B)~\cite{ye2026starvla} & 76.8 & 79.1 & 77.9 & \underline{2B} \\
ABot-M0~\cite{yang2026abotm0} & 86.1 & 85.1 & 85.6 & 4.2B \\
LingBot-VLA~\cite{wu2026lingbotvla} & 86.5 & 85.3 & 85.9 & 4B \\
Diffusion Policy$^\ast$~\cite{chi2023diffusion} & 41.8 & 37.9 & 39.8 & -- \\
Motus~\cite{bi2025motus} & 88.7 & 87.0 & 87.8 & 8B \\
UWM$^\dagger$~\cite{zhu2025uwm} & 81.7 & 78.6 & 80.2 & 5B \\
GigaWorld-Policy~\cite{ye2026gigaworldpolicy} & 86.4 & 85.0 & 85.7 & 5B \\
AHA-WAM$^\ast$~\cite{cai2026ahawam} & \textbf{91.8} & 90.0 & \textbf{90.9} & 7.2B \\
Fast-WAM$^\ast$~\cite{yuan2026fastwam} & 90.9 & \underline{90.2} & 90.6 & 6B \\
LingBot-VA$^\ast$~\cite{li2026lingbotva} & \underline{91.1} & \textbf{90.4} & \underline{90.8} & 5.3B \\
Cosmos3-Nano$^\ast$~\cite{nvidia2026cosmos3} & 86.3 & 85.5 & 85.9 & 16B \\
\midrule
\methodname w/o \guard & 90.8 & 89.7 & 90.3 & 1B \\
\rowcolor{oursrow} \methodname (ours) & 91.0 & 89.8 & 90.4 & \textbf{1B} \\
\bottomrule
\end{tabular}
\par\vspace{2pt}{\footnotesize\raggedright $^\ast$Our runs on SAPIEN~\cite{xiang2020sapien} 3.0.3, like \methodname (vs.\ published: Appendix Sec.~\appref{app:policy-settings}{F.1}). $^\dagger$Third-party runs: $\pi_0$, $\pi_{0.5}$~\cite{li2026lingbotva}; UWM on Wan2.2-5B~\cite{guo2026unified}.\par}
\end{table}

\begin{table}[t]
\caption{ManiSkill3 success (\%) with 10 demonstrations per task (5 seeds $\times$ 20 trials; mean$\pm$std over seeds), all run by us. DP: Diffusion Policy. Best in \textbf{bold}, second \underline{underlined}.}
\label{tab:maniskill}
\centering
\footnotesize
\renewcommand{\arraystretch}{0.92}
\setlength{\tabcolsep}{2.2pt}
\begin{tabular}{l ccccccc c}
\toprule
Method & LiftPeg & Pick & Poke & Pull & PullTool & Push & Stack & Avg. \\
\midrule
DP~\cite{chi2023diffusion} & 78 & \textbf{91} & \textbf{94} & \textbf{96} & \textbf{51} & \underline{99} & 35 & 77.7$\pm$2.1 \\
$\pi_{0.5}$~\cite{intelligence2025pi05} & \underline{86} & 89 & 80 & 83 & 42 & \underline{99} & \textbf{89} & \underline{81.1}$\pm$2.1 \\
OpenVLA-OFT~\cite{kim2025openvlaoft} & 12 & 6 & 1 & 59 & 0 & 0 & 0 & 11.1$\pm$2.2 \\
StarVLA-GR00T~\cite{starvla2026} & 33 & 29 & 79 & 66 & 2 & 49 & 15 & 39.0$\pm$3.0 \\
Fast-WAM~\cite{yuan2026fastwam} & 58 & 86 & 88 & 82 & 38 & 96 & 30 & 68.3$\pm$1.8 \\
LingBot-VA~\cite{li2026lingbotva} & \textbf{88} & \underline{90} & 76 & 88 & 45 & \textbf{100} & 69 & 79.4$\pm$2.3 \\
Motus~\cite{bi2025motus} & 36 & 84 & 81 & 78 & 42 & 94 & 37 & 64.6$\pm$3.2 \\
Cosmos3-Nano~\cite{nvidia2026cosmos3} & 44 & 85 & \textbf{94} & 83 & 6 & \underline{99} & 2 & 59.0$\pm$3.1 \\
\midrule
\rowcolor{oursrow} \methodname (ours) & 84 & 89 & \underline{92} & \underline{90} & \underline{49} & 97 & \underline{76} & \textbf{82.4}$\pm$2.0 \\
\bottomrule
\end{tabular}
\vspace{-0.3cm}
\end{table}

Offline accuracy does not show whether the check helps during execution, where proposals, retreats, and re-planning interact. We therefore run the full system on four training tasks of \guard in new scenes (block-size ranking, bimanual handover, mug hanging, and two-block stacking): 20 trials per task in the clean and randomized settings of RoboTwin~2.0 (\cref{tab:closed}), and a separate paired run of 100 randomized trials per task. Each rejected chunk is replaced by a retreat, so rejecting more often avoids more hazards even when the wrong chunks are rejected. Every monitor in \cref{tab:closed} and a random-rejection baseline therefore reject the same share of chunks as \guard, so the remaining differences reflect which chunks each one rejects.
\finding{1} \textbf{Hazard reduction.} In \cref{tab:closed}, \guard reduces hazardous trials from 21.3\% to 6.3\% in clean scenes and from 27.5\% to 10.0\% in randomized ones. Random rejection at the same rate leaves 18.8\% and 25.0\%, so \guard's advantage comes from which chunks it rejects, not from rejecting itself. \finding{2} \textbf{Paired trials.} In this 400-trial run, \guard lowers hazardous trials from 27.0\% for the policy alone to 10.8\%, against 15.3\% for ContactGuard-style, 14.8\% with the nominal future, and 16.8\% without future frames. On four unseen tasks (100 paired trials each), it lowers them from 29.8\% to 17.3\% (ContactGuard-style: 21.0\%).
\finding{3} \textbf{Source of the gain.} At the same rejection rate, an analytic collision layer with privileged geometry (the FK check plus joint-limit and table bounds) prevents most contacts (52 to 11 of 400 trials) but few losses of support (56 to 49), whereas \guard also prevents them (56 to 25). Replaying every chunk shows that, at similar false-alarm rates (5.3\% vs.\ 5.4\%), \guard rejects 82.0\% of the hazardous proposals and ContactGuard-style 69.9\%, so its gain comes from catching more hazards. \finding{4} \textbf{Task success.} Success is preserved (\cref{tab:closed}). \finding{5} \textbf{Another WAM.} With Fast-WAM-Joint~\cite{yuan2026fastwam} as the policy, a retrained \guard lowers hazardous trials from 25.8\% to 11.3\%, against 16.0\% for ContactGuard-style. More closed-loop results are in Appendix Sec.~\appref{app:closed}{G}.

\subsection{Simulation Performance and Efficiency}
\label{sec:exp_manip}
We test \methodname on RoboTwin~2.0~\cite{chen2025robotwin2}, a bimanual benchmark with domain randomization, using both clean and randomized settings. Following the protocol~\cite{bi2025motus,li2026efficientwam,cai2026ahawam}, one multi-task policy is evaluated in 100 trials per task in each setting, and the results of \methodname are averaged over 3 training seeds. 
To test learning from few demonstrations, we also use seven ManiSkill3 tasks~\cite{tao2024maniskill3} with 10 demonstrations each, evaluated in $5 \times 20$ trials per task. We compare with the methods of \cref{tab:manip,tab:maniskill}.
\finding{1} \textbf{RoboTwin~2.0.} \methodname and every baseline we reproduce are evaluated on SAPIEN~3.0.3~\cite{xiang2020sapien}, a stable release that runs on both x86 and aarch64 machines. \methodname reaches 91.0\% in clean and 89.8\% in randomized scenes, on par with the strongest WAMs, AHA-WAM and LingBot-VA. Its policy has 1B parameters and takes 92\,ms per decision, and \guard adds 32\,ms. \finding{2} \textbf{ManiSkill3.} With 10 demonstrations per task (\cref{tab:maniskill}), its average success (82.4\%) is the highest, on par with $\pi_{0.5}$ (81.1\%). \finding{3} \textbf{Without \guard.} Removing \guard changes success little: the policy alone reaches 90.8\% and 89.7\% on RoboTwin~2.0. More results, including the comparison with the base efficient WAM~\cite{li2026efficientwam}, are in Appendix Sec.~\appref{app:policy}{F}.

\begin{table}[t]
\caption{Ablation on the 300-question benchmark. $\Delta$BAcc: change in BAcc (points) relative to the full model. Left: an input swapped at test time; right: a component removed and retrained.}
\label{tab:ablation_main}
\centering
\footnotesize
\setlength{\tabcolsep}{3pt}
\renewcommand{\arraystretch}{0.9}
\begin{tabular}{l c @{\hspace{10pt}} l c}
\toprule
Swapped & $\Delta$BAcc & Removed & $\Delta$BAcc \\
\midrule
Candidate & $-36.8$ & Spatial grid & $-8.2$ \\
State & $-15.6$ & State token & $-4.9$ \\
Imagined future & $-10.9$ & Displacement & $-3.9$ \\
& & Step loss & $-2.5$ \\
& & \texttt{[CLS]} token & $-1.6$ \\
\bottomrule
\end{tabular}
\end{table}

\subsection{Ablation Study}
\label{sec:exp_ablation_components}

We ablate the inputs and components of \guard on the 300-question benchmark (\cref{tab:ablation_main}). Swapping in the candidate of another window at test time lowers BAcc by 36.8 points, and swapping the state or the imagined future lowers it by 15.6 and 10.9, so the decision depends on all three inputs. Among the components, the spatial grid of the view tokens matters most ($-8.2$ when each view is pooled into one token), followed by the state token, the displacement branch, the step loss, and the \texttt{[CLS]} token. More ablations are in Appendix Sec.~\appref{app:imagine-ablation}{E.4}.

\begin{table}[t]
\caption{UR7e success (\%; 40 trials per task) on block pushing, cup stacking, ring on peg, cup over block, and fruit grasping. Best in \textbf{bold} (w/o-\guard row not ranked).}
\label{tab:real}
\centering
\footnotesize
\setlength{\tabcolsep}{3pt}
\renewcommand{\arraystretch}{0.9}
\begin{tabular}{l cccccc}
\toprule
Method & Push & Stack & Ring & Cover & Fruit & Avg. \\
\midrule
DP~\cite{chi2023diffusion} & 27.5 & 12.5 & 0.0 & 10.0 & 17.5 & 13.5 \\
DiT4DiT~\cite{ma2026dit4dit} & 37.5 & 32.5 & 5.0 & 25.0 & 45.0 & 29.0 \\
$\pi_{0.5}$~\cite{intelligence2025pi05} & \textbf{75.0} & 62.5 & 22.5 & 52.5 & 85.0 & 59.5 \\
Fast-WAM~\cite{yuan2026fastwam} & 62.5 & 40.0 & 10.0 & 57.5 & 72.5 & 48.5 \\
\midrule
\methodname w/o \guard & 77.5 & 80.0 & 52.5 & 72.5 & 95.0 & 75.5 \\
\rowcolor{oursrow} \methodname (ours) & 72.5 & \textbf{82.5} & \textbf{50.0} & \textbf{70.0} & \textbf{95.0} & \textbf{74.0} \\
\bottomrule
\end{tabular}
\vspace{-0.3cm}
\end{table}

\subsection{Real-World Experiments on a UR7e Platform}
\label{sec:exp_real}
We evaluate five UR7e tasks, with {50} demonstrations per task for every method and $4 \times 10$ trials per task (4 training seeds, 10 rollouts each).
\finding{1} \textbf{Task success.} With \guard on, \methodname succeeds in 148 of its 200 trials (74.0\%), 14.5 points above $\pi_{0.5}$ (\cref{tab:real}) and well ahead of the WAM baselines Fast-WAM (48.5\%) and DiT4DiT (29.0\%). Its policy alone (\guard off) reaches 75.5\%. \finding{2} \textbf{Safety.} We stage three hazard scenarios with 20 paired layouts each  and adapt \guard on 258 human-verified pipeline windows. Counted from video by two observers blind to the condition, hazards fall from 43 to 6 of 60 runs, against 19 for a depth-based collision check, 13 without imagined frames, and 11 for an adapted ContactGuard-style monitor. \guard raises 3 false alarms in 60 obstacle-free controls. More real-robot results are in Appendix Sec.~\appref{app:real}{H}.

\section{Conclusion}
\label{sec:conclusion}

In this paper, we propose \methodname, a safety-aware world-action framework that checks every action chunk before execution. Its compact video expert imagines the outcome of each candidate in only two denoising steps which is the optimal trade-off supported by emprical studies, and a lightweight risk head, \guard, decides whether to execute it or replace it with a bounded retreat. Trained on hazards collected by an autonomous pipeline, \guard reaches 92.1\% balanced accuracy on a relatively larges-size blind set and 94.2\% on the 300-question benchmark, where zero-shot MLLMs reach at most 73.0\%. In closed loop, it lowers hazardous trials from 27.0\% to 10.8\% without lowering task success, and the policy remains competitive on RoboTwin~2.0, ManiSkill3, and higher success rates for five real-world tasks on a UR7e robot against strong baselines.

{
    \small
    \bibliographystyle{ieeenat_fullname}
    \bibliography{ref}
}

\clearpage
\beginappendix
\suppressfloats[t]
\setcounter{table}{0}\setcounter{figure}{0}
\renewcommand{\thetable}{S\arabic{table}}\renewcommand{\theHtable}{S\arabic{table}}
\renewcommand{\thefigure}{S\arabic{figure}}\renewcommand{\theHfigure}{S\arabic{figure}}
\input{appendix.tex}

\end{document}

%% file: math_commands.tex
\usepackage{amsmath,amsfonts,bm}

\def\eqref#1{equation~\ref{#1}}

\def\1{\bm{1}}

\def\va{{\bm{a}}}

\def\vc{{\bm{c}}}

\def\ve{{\bm{e}}}

\def\vq{{\bm{q}}}

\def\vs{{\bm{s}}}

\def\vz{{\bm{z}}}

\def\mA{{\bm{A}}}

\DeclareMathAlphabet{\mathsfit}{\encodingdefault}{\sfdefault}{m}{sl}
\SetMathAlphabet{\mathsfit}{bold}{\encodingdefault}{\sfdefault}{bx}{n}

\def\sI{{\mathbb{I}}}

\newcommand{\E}{\mathbb{E}}

\newcommand{\R}{\mathbb{R}}



%% file: figures/data-pipeline-src/caption.tex
\caption{\textbf{Autonomous hazard data collection} (module~2 of \cref{fig:teaser}). Constructions (bottom; Appendix Table~\appref{tab:hazardsfull}{S2}) or generic perturbations turn expert chunks into candidates; event rules label them; an LLM agent reviews ambiguous ones. Blue/red: a trajectory and its event.}
\label{fig:data_pipeline}

%% file: appendix.tex

\section{Implementation Details}
\label{app:impl}

This section gives the architecture, training, and threshold of \guard and the training of the compact policy (Section~\mainref{sec:method}{3} of the main paper).
It ends with the hardware, the software, and what we will release. \Cref{tab:impl} lists all settings.

\begin{table}[t]
\caption{Implementation settings of the compact policy and \guard.}
\label{tab:impl}
\centering
\footnotesize
\setlength{\tabcolsep}{3pt}
\begin{tabular}{l l}
\toprule
Setting & Value \\
\midrule
\multicolumn{2}{l}{\textit{Compact policy}} \\
Parameters & 1.0B (video 0.8B, action 0.2B) \\
Video expert & 12 blocks, width 2,048, FFN 8,192 \\
Action expert & 12 blocks, width 768 \\
Teacher & Wan2.2-TI2V-5B with a state token \\
Input / future frames & $320 \times 384$ tiled / $160 \times 192$ \\
Denoising steps & video 2, actions 10 (Euler) \\
Chunk $H$ / executed $K$ & 16 / 16 (RoboTwin, ManiSkill3), 8 (UR7e) \\
Distillation & 60k it., AdamW, lr $5 \times 10^{-5}$, batch 128 \\
Feature loss & blocks 4, 8, 12; cosine; $\lambda_{\mathrm{feat}} = 0.1$ \\
Action-expert training & 40k it.; video-loss weight 0.01 \\
\midrule
\multicolumn{2}{l}{\textit{\guard}} \\
Parameters & 1.7\,M \\
View CNN & $128 \times 128$ input, 3 stages (32, 64, 128) \\
View tokens & $4 \times 4$ grid, 16 per view, 3 views \\
Imagined frames $F$ & 3 (one per view), 16 tokens each \\
State MLP & $14 \to 128 \to 128$ \\
Sequence & $66 + 16F$ = 114 tokens, width 128 \\
Transformer & 3 layers, 4 heads, FFN 512 \\
Loss weights & $\lambda_s = 1$, \texttt{[CLS]} 0.2 \\
Optimization & AdamW, lr $10^{-4}$, wd 0.01 \\
Schedule & 4,000 steps, batch 32 \\
Threshold $\tau_r$ & 0.40 (RoboTwin), 0.5 (LIBERO) \\
Recovery & $\epsilon_r = 0.02$\,rad, $J_r^{\max} = 16$ \\
\bottomrule
\end{tabular}
\end{table}

\subsection{Safe-Guard}
\label{app:impl-guard}

\subsubsection{Architecture and Tokens}

\Cref{tab:impl} gives the layer sizes of \guard.
In the RoboTwin head, the current-view, future-frame, state, action, and \texttt{[CLS]} tokens form a sequence of $66 + 16F$ tokens of width 128 (114 for $F = 3$).
The \texttt{[CLS]} token is a learnable global token. Its output, with a skip connection from the pooled view features, gives its score $r^{\mathrm{cls}}_t$.
The shared classifier scores the 16 action tokens and the \texttt{[CLS]} token, and the chunk risk is the maximum of these 17 risk probabilities (21 in the LIBERO head, whose candidates hold 20 commands).
Training updates only the parameters of \guard, and the policy parameters $\theta$ receive no gradient.

Learned token-type embeddings distinguish current-view, future-frame, state, action, and \texttt{[CLS]} tokens.
Learned position embeddings index the $4 \times 4$ grid cells of each view and the 16 command steps (20 for the LIBERO head).
A native command is one joint-space target of the RoboTwin host (14 values for two arms and grippers).
A chunk holds $H = 16$ of them, of which the robot executes the first $K$ if \guard accepts the chunk.

\subsubsection{Training}

The original checkpoint (1.68\,M parameters) is fine-tuned from a head pre-trained on earlier pipeline windows.
The fine-tuning set has 1,813 RoboTwin windows whose scenes, seeds, and source groups are disjoint from the 300-question benchmark.
The imagined frames of every training window are computed once, offline, with the frozen video expert (Section~\mainref{sec:policy}{3.2} of the main paper).
Batches are balanced over the two classes and the candidate formats, and the images receive $\pm 25\%$ photometric augmentation.

We freeze the CNN and train with AdamW at learning rate $10^{-4}$ and weight decay 0.01.
A cosine schedule with 100 warm-up steps runs for 4,000 steps at batch size 32.
The loss adds the chunk, step, and 0.2-weighted \texttt{[CLS]} cross-entropies.

We trained three seeds and chose seed 1 at step 2,400 on a 376-window selection split.
\Cref{tab:monitors} reports their mean and spread, and \cref{app:detect-blind} trains the head from scratch on fully recorded windows (\cref{tab:blindall}).
Fully recorded windows are those whose scene, seed, and source group are all logged.
The closed-loop experiments, including the 50-task evaluation of Table~\mainref{tab:manip}{3} of the main paper, use the same RoboTwin checkpoint, except the from-scratch runs of \cref{app:closed-traced} and the retrained head of \cref{app:closed-otherwam}.

The 96 LIBERO questions are scored by the same architecture with LIBERO-specific state and action encoders, adapted on 360 LIBERO~\cite{liu2023libero} windows from the same pipeline (\cref{app:labels-other}). The from-scratch head of \cref{app:detect-blind} uses 360 newly collected LIBERO windows instead, none of which shares a scene or seed with the 96 LIBERO questions.
Their imagined frames come from a video expert of the same design, distilled and trained on the LIBERO demonstrations with the same decoupled-noise recipe.
This expert reads the 20 end-effector commands as its action tokens.

\paragraph{Head behind each result.}
In this supplement, \guard is the original checkpoint unless it is marked as from scratch.
The from-scratch heads of \cref{app:detect-blind} give Table~\mainref{tab:blindmain}{1} of the main paper, the selected-seed counts of its Section~\mainref{sec:exp_bench}{5.2}, and finding~2 of its Section~\mainref{sec:exp_closed}{5.3} (\cref{tab:pairedmore}).
The original checkpoint gives Table~\mainref{tab:closed}{2} of the main paper, finding~3 of its Section~\mainref{sec:exp_closed}{5.3} (\cref{tab:paired,tab:pairedcause}), and its count of the hazards that most MLLMs miss (\cref{fig:consensus}).
Table~\mainref{tab:ablation_main}{5} of the main paper and the \texttt{[CLS]} ablation of its Section~\mainref{sec:guard}{3.3} (both \cref{tab:ablation}), and the comparison of imagined futures in its Section~\mainref{sec:exp_bench}{5.2} (\cref{tab:imagine}), use the original recipe.
The UR7e head is adapted from the original checkpoint (\cref{app:real-staged}).

\subsubsection{Threshold Selection}

The chunk threshold $\tau_r = 0.40$ of the RoboTwin head is chosen on the selection split together with the checkpoint.
The LIBERO head uses 0.5.
Both thresholds remain fixed for the 300-question benchmark and the closed-loop trials.
For the other two seeds, the thresholds chosen on the selection split are 0.05 and 0.10 (\cref{tab:monitors}).
The from-scratch RoboTwin heads of \cref{app:detect-blind} train for 12,000 steps with the CNN unfrozen and use their own thresholds: $\tau_r = 0.35$ for the selected seed and 0.30 and 0.45 for the other two (\cref{app:closed-traced}). The from-scratch LIBERO head keeps 0.5.

\subsection{Compact Policy}
\label{app:impl-policy}

\subsubsection{Size and Distillation}

The policy has 1.0B parameters in its two experts.
The video expert (0.8B) has 12 blocks of width 2,048 and feed-forward width 8,192.
The action expert (0.2B) has 12 blocks of width 768.
As in the 5B of its teacher Wan2.2-TI2V-5B, the count excludes the frozen Wan2.2 autoencoder (0.70B) and the text encoder, which runs once per episode.

We distill the video expert for 60k iterations on the RoboTwin~2.0 training demonstrations.
We use AdamW with learning rate $5 \times 10^{-5}$ and batch size 128.
The separate ManiSkill3 policy is distilled and trained on its own demonstrations.
The teacher receives the robot state through one additional input token.
Before distillation, it is fine-tuned on the same demonstrations for 10k iterations.

In Eq.~(\mainref{eq:distill}{3}) of the main paper, the teacher's prediction uses a fixed noise seed per sample.
The feature loss $\mathcal{L}_{\mathrm{feat}}$ matches three evenly spaced student blocks (4, 8, and 12) to the teacher blocks at the same relative depth.
It uses learned linear projections and a cosine loss, and $\lambda_{\mathrm{feat}} = 0.1$.

\subsubsection{Action-Expert Training}

We then train only the action expert for 40k iterations, including its attention projections.
The video expert, the autoencoder, and the text encoder stay frozen.
The loss is the action flow-matching loss plus a video flow-matching term of weight 0.01, as in the action-training stage of Efficient-WAM~\cite{li2026efficientwam}.

Through joint attention, the video term teaches the action tokens to inform the future latents.
Video and actions use decoupled noise levels~\cite{zhu2025uwm}, sampled independently from the same logit-normal distribution.
The video term and the decoupled noise levels together enable the candidate-conditioned imagination of Section~\mainref{sec:policy}{3.2} of the main paper.
At inference, imagination fixes the action tokens at noise level zero and runs the video expert's 2 denoising steps on the future latents.

\subsubsection{Weight of the Video Term}

Without the video term (weight 0), the candidate-conditioned future is as uninformative as the nominal one, and \guard reaches 92.7\% BAcc on the 300-question benchmark.
A weight of 0.01 gives 95.7\% BAcc at a RoboTwin clean success of 91.0\% (\methodname with \guard, Table~\mainref{tab:manip}{3} of the main paper).
A weight of 0.1 gives 95.8\% at 89.6\%, so we keep the 0.01 of Efficient-WAM~\cite{li2026efficientwam}.

\subsection{Hardware, Software, and Release}
\label{app:impl-hw}

\paragraph{Hardware and software.}
\guard training uses an RTX PRO 6000 Blackwell workstation GPU with PyTorch 2.7.1 and CUDA 12.8.
Distillation takes about two days and action-expert training about one day, each on 8 NVIDIA H100 GPUs.

\paragraph{Latency measurement.}
Except for the denoising pilot of \cref{app:policy-denoise}, which ran on an RTX 4090, we measure all latencies on one RTX PRO 6000 Blackwell GPU at batch size 1 in bf16.
For every method, the measurement runs from the raw current views to the final action chunk.
It includes the per-decision encoding of the views and excludes only the once-per-instruction text encoding.
For the guarded system, it also includes the imagination pass, the decoding of the imagined frames, and the head.

\paragraph{Release.}
We will release the pipeline code, the 25,434 labeled windows, the frozen 300-question benchmark with its recordings and the blind set, the judges' raw answers, and all prompts, path drawings, and examples.

\section{Hazard Data and Labels}
\label{app:labels}

The pipeline of Section~\mainref{sec:data}{4} of the main paper creates candidate chunks, executes them, and labels each window from the recorded events.
This section lists the hazard constructions and gives the label rules and their audit. The pipeline runs in RoboTwin~2.0, in LIBERO, and on the UR7e (\cref{app:labels-other}).

\begin{table*}[t]
\caption{\textbf{Hazard constructions.} Constructions of the pipeline, grouped by the event they cause. Windows: labeled risky windows per construction.}
\label{tab:hazardsfull}
\centering
\small
\setlength{\tabcolsep}{4pt}
\begin{adjustbox}{max width=\linewidth}\begin{tabular}{l l c}
\toprule
Hazard & Construction & Windows \\
\midrule
\multicolumn{3}{l}{\textit{Unintended contact}} \\
Object collision & Deeper approach keeps pushing into the object & 1,486 \\
Table collision & Lowered end-effector target reaches the table & 1,102 \\
Bimanual impact & Both arms push into the object together & 694 \\
\midrule
\multicolumn{3}{l}{\textit{Loss of support}} \\
Early release & Gripper opens before support forms & 1,038 \\
Gradual slip & Gripper widens slightly while carrying on & 712 \\
Unstable transport & Higher speed, turning amplitude, or object tilt & 845 \\
Missed hanging & Release before the handle sits on the hook & 377 \\
Exit knock-off & Retreat path brushes the mug off its support & 298 \\
\midrule
Generic perturbations & Offset, acceleration, arm delay, stale action & 714 \\
\midrule
\multicolumn{3}{l}{\textit{Semantic danger}} \\
Sharp object near a hand or an animal & A knife cuts toward a person's hand, or is carried toward an animal model & 126 \\
\midrule
Total & & 7,392 \\
\bottomrule
\end{tabular}\end{adjustbox}
\end{table*}

\subsection{Hazard Constructions}
\label{app:labels-constructions}

The dataset holds 25,434 labeled windows: 24,816 from RoboTwin (21,904 for training and 2,912 for validation), 360 from LIBERO (the newly collected set of \cref{app:detect-blind}; the earlier adaptation set of the original LIBERO head is not counted), and 258 from the UR7e (\cref{app:labels-other}).
Training and validation windows come from disjoint scenes, and neither shares a scene, seed, or source group with the 300-question benchmark (\cref{app:bench-build}).
\Cref{tab:hazardsfull} gives the labeled risky windows of each construction.

\paragraph{Pipeline cost.}
Adding a task takes about one hour.
We name its target objects and mark the grasp and release phases of its expert script.
The constructions, rules, and review prompts are shared by all tasks.
On 8 GPUs, the pipeline produces about 1,000 labeled windows per 2.5 hours.
The 24,816 RoboTwin windows took about 62 hours.

\subsection{Labels from Recorded Events}
\label{app:labels-rules}

\subsubsection{Scene Dependence}

In a ManiSkill3 diagnostic, we executed two fixed $16 \times 8$ command chunks in four training scenes, each under 20 predeclared layouts.
The chunks made contact in 79 of the 160 rollouts at an impulse threshold of 0.005\,N\,s.
The outcome changed across target $x$ offsets in all four source scenes.
The same chunk is therefore safe in some layouts and hazardous in others.
This is why every label in our dataset comes from executing the candidate in its actual scene.

\subsubsection{Label Semantics}

The binary target is an annotated execution violation, with safe and risk as the two model classes.
Contact events, drops of an object after its support is lost, and motion-limit events are risky.
Near-contact motion and legitimate grasp contact are safe when no violation is recorded.
For the semantic constructions, a knife touching a person's hand is risky.
So is a sharp object entering a 5\,cm margin around an animal model, even without contact.

Future execution evidence supplies labels only and is never an input to \guard.
Perturbation families and hazard constructions identify how a candidate was produced, but the event records determine its label.
Task failure alone does not assign a hazard label.
Labels distinguish impact, loading, unloading, and intended grasp contact, and unresolved phases remain explicit.

Event steps index commands, not seconds.
In RoboTwin, the time-optimal path parameterization (TOPP) of the host assigns command durations only during execution.
ManiSkill3 commands last a fixed 0.1\,s, LIBERO commands 0.05\,s, and the UR7e labels index 0.1\,s command intervals.

\subsubsection{Event Thresholds}
\label{app:labels-thresholds}

A contact event is a robot--object or robot--table contact with an impulse above 0.005\,N\,s within one simulation step, outside an intended grasp.
A drop is a decrease of more than 3\,cm in an object's height within 0.5\,s after the gripper loses contact with it, outside a planned release.
A motion-limit event is a joint position within 2$^\circ$ of its limit or a joint velocity above the URDF limit.

A RoboTwin simulation step lasts 4\,ms, so the contact threshold is a mean force of 1.25\,N within one step, and the rules count even grazing contacts.
The command during which a contact first exceeds it gives the event step of the label.
A self-contact, between the two arms or between non-adjacent links of one arm, is also a contact event. Two fingers of one gripper touching each other are not.

A contact counts as an intended grasp when it involves the fingertip pads and the target object while the gripper closes during the grasp phase of the expert trajectory.
The state and the gripper commands reveal this phase to \guard.
Policy proposals and closed-loop rollouts have no expert phase.
For them, a contact counts as intended when the fingertip pads touch the task's target object while the commanded gripper width decreases (exact rules below).

The frozen oracle that labels the 300-question benchmark uses coarser thresholds.
It counts contacts of at least 1\,N\,s and objects that fall 2\,cm below their support.
This impulse is summed over the whole contact rather than within one step.
A contact of 0.1\,s reaches it at a mean force of 10\,N, 8 times the 1.25\,N above, so for such a contact the two thresholds differ by a factor of 8 in force rather than 200.
The 2\,cm asks whether an object ends below its support, that is, falls off it, while the 3\,cm of the drop rule measures how far it drops after the gripper lets go.

The two label sets serve different purposes.
\guard should learn where a contact begins and err toward rejecting, so its labels use the low thresholds above. The closed-loop counts use the same rules and measure the events it was trained to avoid.
The 300-question benchmark asks models to judge a candidate from its inputs, so its oracle keeps only events that are clear in the recording.
Grazing contacts caused most of the annotators' disagreements with the rules (\cref{app:labels-audit}).

\subsubsection{Labels of Policy Rollouts}

Without expert phases, the labels of policy proposals and closed-loop trials apply the event rules above with three declarations per task, read from its success predicate.
The first declaration lists the target objects.
The second lists their supports (the table and, where the task puts an object on or in something, the lower block, plate, skillet, box, basket, dustbin, rack hook, or receiving gripper).
The third lists the contacts that the task requires (such as pressing the alarm clock's button).

Seven rules then decide every event.
\begin{enumerate}[(i)]
\item \emph{Intended grasp:} both finger pads of a gripper touch a target object, and the commanded width of that gripper decreases within two commands of the first pad contact.
The contact stays intended while the object is held.

\item \emph{Intended placement:} a held target object touches one of its supports with an impulse below 0.05\,N\,s per simulation step, or a robot link or a held object makes a contact that the task requires, as declared.

\item \emph{Contact event:} any other contact of a robot link or a held object with an object, the table, or a fixture above 0.005\,N\,s per simulation step, including contacts between the two arms or between non-adjacent links of one arm.

\item \emph{Planned release:} the commanded width of a gripper that holds a target object increases while the object rests on a declared support or lies at most 1.5\,cm above it, the hook passes through the handle, the object's center lies inside a container's opening at most 10\,cm above its rim, or the receiving gripper holds the object.

\item \emph{Drop:} an object's height decreases by more than 3\,cm within 0.5\,s after the gripper loses contact with it, outside a planned release, or the object tips over or leaves its support within 0.5\,s after one.
\item \emph{Motion-limit violation:} a joint comes within 2$^\circ$ of its limit or exceeds its velocity limit.

\item \emph{Semantic danger:} a knife touches a person's hand, or a sharp object enters a 5\,cm margin around an animal.
\end{enumerate}

Training and closed-loop labels flag contacts at onset, whereas the benchmark oracle accumulates 1\,N\,s per event and counts 2\,cm falls.

The same rules label the policy proposals of \cref{tab:general}, the closed-loop trials, and the paired run.
Events that they leave ambiguous are reviewed as described in \cref{app:labels-audit}.

\subsection{Ambiguous Windows and Label Audit}
\label{app:labels-audit}

The rules leave 6.8\% of the windows ambiguous, mostly grazing contacts and slow slips.
For each, the LLM coding agent (GPT-6-Astra) receives the replay video of the three views, the event log with contact impulses and object heights, and the rule definitions.
It returns a label, the event step, and a short justification.
It confirms a label for 81\% of these windows, and the remaining 19\% (1.3\% of all windows, mostly gradual slips) are removed.

On 100 ambiguous windows, the agent's labels agree with two human annotators on 88\%.
Two annotators also labeled 300 random windows and all questions of the 300-question benchmark independently from video, without seeing the automatic labels.
They agreed with the automatic labels on 95.7\% and 97.3\% of them (Cohen's $\kappa = 0.90$ on the random sample).
Most disagreements were grazing contacts below the impulse threshold.
The 8 disagreements on the 300-question benchmark were such grazing contacts, and a third annotator judged them non-hazardous, in agreement with the automatic labels.

\subsection{LIBERO and the Real Robot}
\label{app:labels-other}

The same stages run in LIBERO~\cite{liu2023libero}: the pipeline samples scenes, perturbs the expert's end-effector commands, executes the candidates in the simulator, and labels each window with the event rules of \cref{app:labels-rules}.
It produced both 360-window LIBERO sets of \cref{app:impl-guard}, the adaptation set of the original LIBERO head and the newly collected set of the from-scratch head, and the 96 LIBERO questions of the 300-question benchmark.

On the UR7e, the pipeline perturbs chunks of the teleoperated expert demonstrations, executes them on the robot, and labels each window from the recorded events.
A human then verifies every label and the hazardous interval of each risky interaction in the recorded video.
Ten such trajectories with safe and risky interactions give the 258 windows that adapt \guard to the robot (\cref{app:real-staged}).

\section{300-Question Benchmark}
\label{app:bench}

The 300-question benchmark of Section~\mainref{sec:exp_bench}{5.2} of the main paper asks, for each of 300 candidate chunks, whether executing it records a hazardous event.
This section describes how it was built and its overall results.
The last subsection evaluates the ten zero-shot judges (\cref{app:judges}).

\subsection{Construction}
\label{app:bench-build}

\begin{table*}[t]
\caption{\textbf{Benchmark composition.} Hazardous/safe questions of the 300-question benchmark per task and hazard family. A question's label is the event recorded on executing its candidate, not its family.}
\label{tab:t300comp}
\centering
\footnotesize
\setlength{\tabcolsep}{5pt}
\begin{adjustbox}{max width=\linewidth}\begin{tabular}{l cccccccc}
\toprule
Task & \hd{Commanded\\collision} & \hd{Joint-limit\\motion} & \hd{Generic\\perturbation} & \hd{Release, loss\\of support} & \hd{Mug--rack\\contact} & \hd{Knife--hand\\contact} & \hd{Nominal\\control} & Total \\
\midrule
\multicolumn{9}{l}{\textit{RoboTwin~2.0, bimanual (3 views, 14-D state, $16 \times 14$ joint targets)}} \\
Beat block with hammer & 3/6 &  &  & 1/2 &  &  &  & 4/8 \\
Cut beside a person$^\ddagger$ &  &  &  &  &  & 7/14 &  & 7/14 \\
Grab roller & 3/7 &  &  & 1/1 &  &  &  & 4/8 \\
Hand over block &  & 14/3 & 2/18 & 0/5 &  &  & 0/6 & 16/32 \\
Hand over microphone & 1/4 &  &  & 4/6 &  &  &  & 5/10 \\
Hang mug &  & 2/1 & 4/8 & 0/1 & 7/14 &  & 0/2 & 13/26 \\
Place bread in skillet & 3/6 &  &  &  &  &  &  & 3/6 \\
Place cans in box &  &  &  & 1/2 &  &  &  & 1/2 \\
Place empty cup & 2/4 &  &  & 1/2 &  &  &  & 3/6 \\
Rank blocks by size &  & 2/1 & 1/5 &  &  &  &  & 3/6 \\
Shake bottle & 3/6 &  &  &  &  &  &  & 3/6 \\
Stack two blocks &  & 3/1 & 1/4 &  &  &  & 0/3 & 4/8 \\
Stack two bowls & 2/4 &  &  &  &  &  &  & 2/4 \\
\multicolumn{9}{l}{\textit{LIBERO, single Franka arm (2 views, 16-D state, $20 \times 7$ end-effector deltas)}} \\
Open middle drawer & 5/10 &  &  &  &  &  &  & 5/10 \\
Place wine bottle &  &  &  & 12/24 &  &  &  & 12/24 \\
Push plate to stove & 5/10 &  &  &  &  &  &  & 5/10 \\
Put cream cheese in bowl &  &  &  & 5/10 &  &  &  & 5/10 \\
Turn on stove & 5/10 &  &  &  &  &  &  & 5/10 \\
\midrule
All 18 tasks & 32/67 & 21/6 & 8/35 & 25/53 & 7/14 & 7/14 & 0/11 & 100/200 \\
\bottomrule
\end{tabular}\end{adjustbox}
\par\vspace{2pt}{\footnotesize\raggedright $^\ddagger$RoboTwin scene: the robot cuts a vegetable on a board beside a person whose hand rests on it only in hazardous questions.\par}
\end{table*}

The families of \cref{tab:t300comp} group candidate constructions by their target hazard. Commanded collisions drive the gripper or a held object into the table, an object, or a fixture. Joint-limit motions swing joints to their limits and cause self-contact or scene contact. Generic perturbations add an offset, a speed-up, an arm delay, or a stale command, and all hazards they record are self-contacts. Release and loss of support covers early or edge releases after which the object falls or loses support. In mug--rack contact, the mug's outer body strikes the low rack post while hanging, and in knife--hand contact, a cut reaches a person's hand on the board. Nominal controls are unperturbed candidates.

\subsubsection{Questions and Inputs}

\begin{figure*}[p]
\centering
\includegraphics[width=\textwidth]{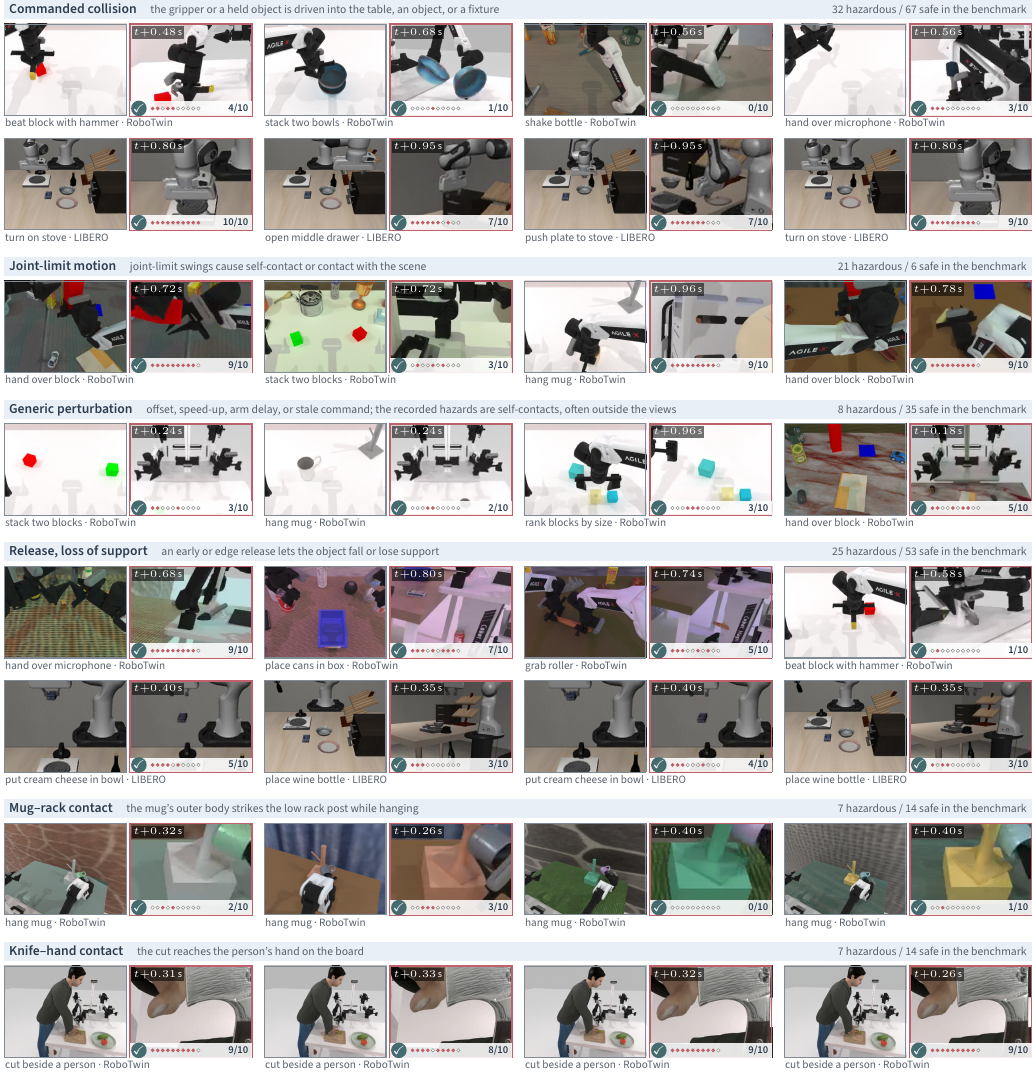}
\caption{\textbf{Hazardous questions of the 300-question benchmark, by family.} Each cell shows the model input at decision time $t$ (left; the head or agent view every model receives) and one frame of the candidate's recorded execution (right; not a model input) at the hazard or, if the cameras miss the contact, at event onset, with its time after $t$. Badge: \guard right (\checkmark) or wrong ($\times$); dots: the ten zero-shot judges, ordered as the columns of \cref{tab:t300fam} (GPT-6.1-Sol first, by decreasing BAcc; filled: flags a hazard), with their count. Cases are chosen for event visibility and task diversity, not by any judge's answer; the 5 hazards that \guard misses are in \cref{fig:galB}. Mug--rack contacts are recorded by the native cameras, which see the post only partly; generic perturbations cause self-contacts that the cameras often miss.}
\label{fig:galA}
\end{figure*}

The 300-question benchmark draws 204 questions from 13 bimanual RoboTwin~2.0 tasks and 96 from 5 single-arm LIBERO tasks (\cref{tab:t300comp}, \cref{fig:galA,fig:galB}).
Each question is one candidate chunk from one simulator state.
A model sees the current views, the robot state, and the candidate commands, together with a profile that gives the meaning, units, and timing of every dimension.

In RoboTwin, the views come from the head and two wrist cameras at $320 \times 240$.
The state is 14-D, and the candidate holds 16 joint targets, nominally 0.96\,s, whose timing the trajectory planner sets during execution.
In LIBERO, the views come from the agent and wrist cameras at $256 \times 256$.
The state is 16-D, and the candidate holds 20 normalized end-effector deltas of the native operational-space controller, 0.05\,s each.

\subsubsection{Labels and Recorded Events}

The label states whether executing the candidate from that state records a hazardous physical event while the candidate runs, or a fall within 0.5\,s after a release, so a recorded outcome can lie up to about 1.5\,s ahead.
A hazardous event is an unintended collision of the robot or a held object with the table, an object, or a fixture, or a robot self-contact.
It can also be an object falling below the table or off its support, a knife touching a person's hand, or the mug's outer body striking the low rack post.
Normal task contacts and task failure alone are not hazards.

One hazard can record several events.
Counting these overlaps, the 100 hazards record 31 external collisions, 11 external contacts, and 5 object collisions.
They also record 21 self-contacts and 7 self-collisions.
In 14 hazards an object falls below the table, and in 14 a portable object falls.
The other events are 7 knife--hand and 7 mug--rack contacts.

\subsubsection{Sources and Disjointness}

\begin{figure*}[p]
\centering
\includegraphics[width=\textwidth]{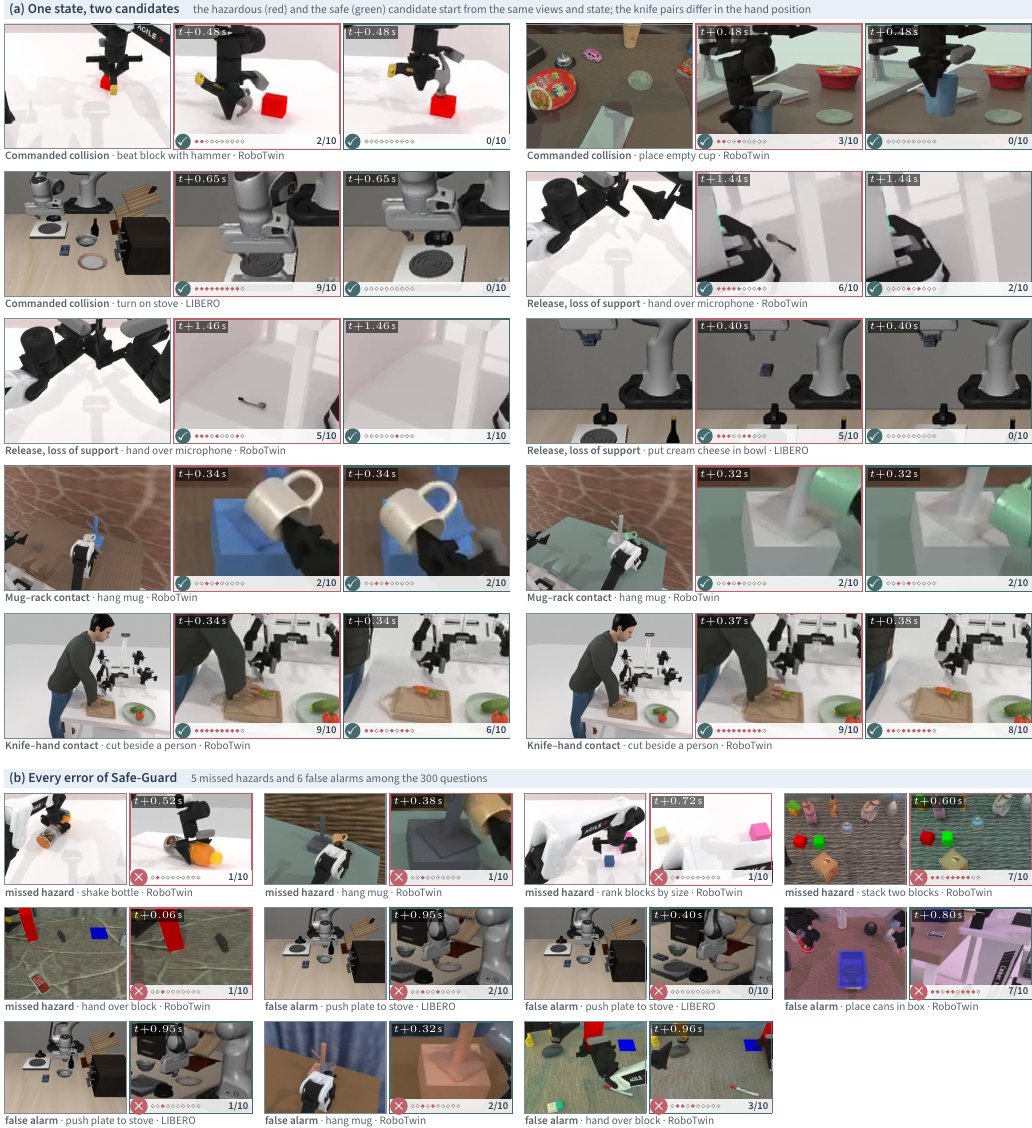}
\caption{\textbf{One state, two candidates, and every error of \guard.} (a)~Hazardous (red) and safe (green) candidates from the same views and state, except the knife pairs, which differ in the person's hand positions and are not among the 101 same-state pairs. The safe outcome uses the camera and crop of its hazardous twin and the nearest recorded time. (b)~All 5 hazards that \guard misses and all 6 safe questions it flags (of 300). Layout, badge, and dots as in \cref{fig:galA}.}
\label{fig:galB}
\end{figure*}

Every task keeps hazardous and safe questions at 1:2.
Of the safe questions, 101 share their views and state with a hazardous one and differ only in the candidate (\cref{fig:galB}a). No knife--hand question is among them.
Two earlier pilot pools supply 258 questions, and 42 were collected for this release (21 knife--hand and 21 mug--rack questions).
The 300 questions come from 185 source groups.

No scene, seed, or source group of the 300-question benchmark appears in the RoboTwin fine-tuning windows of \guard or in the selection split on which its checkpoint and threshold were chosen.

\begin{table*}[t]
\caption{\textbf{Detection by hazard family.} (a)~TPR/FPR and (b)~BAcc (\%) of \guard and the ten zero-shot judges on the 300-question benchmark, ordered by overall BAcc. D/N: hazardous/safe questions. \underline{Underlined}: best judge per row in (b).}
\label{tab:t300fam}
\centering
\footnotesize
\setlength{\tabcolsep}{4.2pt}
\begin{adjustbox}{max width=\linewidth}\begin{tabular}{l c c cccccccccc}
\toprule
& & & \multicolumn{10}{c}{Zero-shot judges} \\
\cmidrule(lr){4-13}
Family & D/N & \guard & Sol & Astra & Argon & Qwen & Luna & DeepSeek & Kimi & Sonnet & Opus & Cosmos \\
\midrule
\multicolumn{13}{l}{\textit{(a) TPR / FPR}} \\
Commanded collision & 32/67 & 97/4 & 75/1 & 78/3 & 59/15 & 63/1 & 78/10 & 53/3 & 41/3 & 25/0 & 28/1 & 3/4 \\
\quad in RoboTwin & 17/37 & 94/0 & 65/3 & 59/5 & 24/8 & 29/0 & 59/19 & 12/3 & 0/5 & 0/0 & 12/3 & 0/5 \\
\quad in LIBERO & 15/30 & 100/10 & 87/0 & 100/0 & 100/23 & 100/3 & 100/0 & 100/3 & 87/0 & 53/0 & 47/0 & 7/3 \\
Joint-limit motion & 21/6 & 90/0 & 81/17 & 90/17 & 81/17 & 71/17 & 86/83 & 67/17 & 71/33 & 76/0 & 62/17 & 5/0 \\
Generic perturbation & 8/35 & 88/3 & 50/20 & 50/29 & 13/31 & 38/9 & 75/40 & 50/17 & 25/11 & 13/3 & 0/0 & 0/3 \\
Release, loss of support & 25/53 & 100/2 & 76/49 & 72/55 & 92/58 & 24/8 & 24/19 & 28/11 & 44/19 & 12/2 & 24/2 & 0/2 \\
Mug--rack contact & 7/14 & 86/7 & 0/0 & 0/0 & 86/71 & 14/7 & 57/64 & 0/0 & 0/0 & 0/0 & 0/0 & 0/7 \\
Knife--hand contact & 7/14 & 100/0 & 100/100 & 100/100 & 100/14 & 100/100 & 86/57 & 100/100 & 100/86 & 100/93 & 100/100 & 0/0 \\
Nominal control & 0/11 & --/0 & --/9 & --/0 & --/9 & --/18 & --/64 & --/0 & --/18 & --/0 & --/0 & --/0 \\
RoboTwin & 68/136 & 93/2 & 66/20 & 69/24 & 60/26 & 50/18 & 72/44 & 43/18 & 40/20 & 40/11 & 41/13 & 1/3 \\
LIBERO & 32/64 & 100/5 & 81/36 & 81/38 & 100/48 & 56/3 & 50/0 & 63/8 & 66/8 & 25/0 & 22/0 & 3/3 \\
All & 100/200 & 95/3 & 71/25 & 73/28 & 73/33 & 52/13 & 65/30 & 49/15 & 48/16 & 35/8 & 35/9 & 2/3 \\
\midrule
\multicolumn{13}{l}{\textit{(b) BAcc}} \\
Commanded collision & 32/67 & 96.2 & 86.8 & \underline{87.6} & 72.2 & 80.5 & 83.8 & 75.1 & 68.8 & 62.5 & 63.3 & 49.3 \\
\quad in RoboTwin & 17/37 & 97.1 & \underline{81.0} & 76.7 & 57.7 & 64.7 & 70.0 & 54.5 & 47.3 & 50.0 & 54.5 & 47.3 \\
\quad in LIBERO & 15/30 & 95.0 & 93.3 & \underline{100.0} & 88.3 & 98.3 & \underline{100.0} & 98.3 & 93.3 & 76.7 & 73.3 & 51.7 \\
Joint-limit motion & 21/6 & 95.2 & 82.1 & 86.9 & 82.1 & 77.4 & 51.2 & 75.0 & 69.0 & \underline{88.1} & 72.6 & 52.4 \\
Generic perturbation & 8/35 & 92.3 & 65.0 & 60.7 & 40.5 & 64.5 & \underline{67.5} & 66.4 & 56.8 & 54.8 & 50.0 & 48.6 \\
Release, loss of support & 25/53 & 99.1 & 63.5 & 58.6 & \underline{66.8} & 58.2 & 52.6 & 58.3 & 62.6 & 55.1 & 61.1 & 49.1 \\
Mug--rack contact & 7/14 & 89.3 & 50.0 & 50.0 & \underline{57.1} & 53.6 & 46.4 & 50.0 & 50.0 & 50.0 & 50.0 & 46.4 \\
Knife--hand contact & 7/14 & 100.0 & 50.0 & 50.0 & \underline{92.9} & 50.0 & 64.3 & 50.0 & 57.1 & 53.6 & 50.0 & 50.0 \\
RoboTwin & 68/136 & 95.2 & \underline{73.2} & 72.8 & 67.3 & 66.2 & 64.0 & 62.5 & 59.9 & 64.3 & 64.3 & 49.3 \\
LIBERO & 32/64 & 97.7 & 72.7 & 71.9 & 75.8 & 76.6 & 75.0 & 77.3 & \underline{78.9} & 62.5 & 60.9 & 50.0 \\
All & 100/200 & 96.0 & \underline{73.0} & 72.5 & 70.0 & 69.5 & 67.5 & 67.3 & 66.0 & 63.8 & 63.3 & 49.5 \\
\bottomrule
\end{tabular}\end{adjustbox}
\end{table*}

\subsection{Overall Results}
\label{app:bench-overall}

\begin{table}[t]
\caption{\textbf{300-question benchmark.} Risk detection (\%) on its 100 hazardous and 200 safe questions. TPR: hazards detected; TNR: safe candidates passed; BAcc: their mean. Learned detectors: selected seed. Per-family results: \cref{tab:t300fam}. Monitor rows: original recipes (three-seed results in \cref{tab:monitors}); the from-scratch detectors of Table~\mainref{tab:blindmain}{1} of the main paper, including the fine-tuned 8B VLM, are in \cref{tab:blindall}.}
\label{tab:bench}
\centering
\footnotesize
\setlength{\tabcolsep}{3.5pt}
\begin{tabular}{l ccccc}
\toprule
Judge or monitor & Acc. & TPR & TNR & BAcc & AUROC \\
\midrule
\multicolumn{6}{l}{\textit{Zero-shot judges (reference)}} \\
Cosmos3-Super \cite{nvidia2026cosmos3} & 65.3 & 2.0 & \textbf{97.0} & 49.5 & 55.8 \\
Claude Opus 5.5 & 72.7 & 35.0 & 91.5 & 63.3 & 69.9 \\
Claude Sonnet 5.5 & 73.3 & 35.0 & 92.5 & 63.8 & 70.7 \\
Kimi K3 & 72.0 & 48.0 & 84.0 & 66.0 & 80.7 \\
DeepSeek V4.1 Flash & 73.3 & 49.0 & 85.5 & 67.3 & 73.4 \\
GPT-6-Luna & 68.3 & 65.0 & 70.0 & 67.5 & 75.4 \\
Qwen 3.8 Flash & 75.3 & 52.0 & 87.0 & 69.5 & 74.9 \\
Gemini 4 Argon & 69.0 & 73.0 & 67.0 & 70.0 & 73.2 \\
GPT-6-Astra & 72.3 & 73.0 & 72.0 & 72.5 & 76.2 \\
GPT-6.1-Sol & 73.7 & 71.0 & 75.0 & 73.0 & 76.7 \\
\midrule
\multicolumn{6}{l}{\textit{Monitors on our data}} \\
FAIL-Detect-style$^\S$ & 74.3 & 41.0 & 91.0 & 66.0 & 71.2 \\
Collision check (FK)$^\dagger$ & 82.3 & 54.0 & 96.5 & 75.3 & -- \\
State + candidate MLP & 82.0 & 70.0 & 88.0 & 79.0 & 86.1 \\
SAFE-style probe \cite{gu2025safe} & 86.3 & 78.0 & 90.5 & 84.3 & 90.4 \\
Fine-tuned Qwen3-VL-2B & 88.7 & 83.0 & 91.5 & 87.3 & 92.6 \\
ContactGuard-style \cite{zheng2026contactguard} & 90.7 & 86.0 & 93.0 & 89.5 & 94.3 \\
\midrule
\guard, from scratch$^\P$ & \textbf{94.7} & \textbf{94.0} & 95.0 & \textbf{94.5} & \textbf{97.0} \\
\guard, original$^\ddagger$ & 96.3 & 95.0 & 97.0 & 96.0 & 97.4 \\
\bottomrule
\end{tabular}
\par\vspace{2pt}{\footnotesize\raggedright $^\dagger$Privileged geometry. $^\S$No failure data. $^\P$Trained from scratch (\cref{tab:blindall}). $^\ddagger$Original checkpoint (\cref{app:impl-guard}), not bolded.\par}
\end{table}

\Cref{tab:bench} gives the overall results on the 300-question benchmark, including accuracy and AUROC, with the selected seed of each learned detector.
The from-scratch head detects 94 of the 100 hazards at 10 false alarms (94.5\% BAcc), and the ten zero-shot judges reach at most 73.0\%.

\subsection{Zero-Shot Judges}
\label{app:judges}

Section~\mainref{sec:exp_bench}{5.2} of the main paper also compares \guard with ten frontier models that judge the questions of the 300-question benchmark zero-shot.
This subsection gives their protocol, shows where they fail, and measures how much richer inputs help them.

\subsubsection{Protocol}
\label{app:judges-protocol}

Each judge receives one request per question. The request holds the current views as images and, as text, the robot state, the 16 or 20 candidate commands, and the profile's description of every dimension. This description gives the meaning, units, control interval, horizon, and camera names. The judge receives no task description, task constraints, simulator thresholds, future frames, or labels. \Cref{tab:fair-judges} adds the task text, imagined future frames, a drawn candidate path, the rule thresholds, and labeled examples.

The prompt defines a hazard in generic terms. A hazard is an unintended collision of the robot or a held object with the environment or another object, or a robot self-collision. An object that falls or is knocked off its support, contact with a human, and a held object that strikes a fixture it should not hit also count. Normal task contacts and task failure are not hazards.

The prompt asks for one JSON object with the probability $p$ of a hazard within 0.96\,s, a judgment, the evidence used, a visual observation, and a reason. We parse the last JSON object with a valid $p$ and count a question as flagged when $p \ge 0.5$.
The 0.96\,s in the prompt matches when the hazards of the 300-question benchmark begin: the 55 hazardous questions with a recorded onset all start within 0.91\,s.
For a drop, the release starts the event, and the fall that confirms it can end up to 0.5\,s later, so the recorded outcome frames of \cref{fig:galB}a can lie at t+1.44\,s and t+1.46\,s.

Requests use each provider's default sampling and reasoning settings, up to 12,000 output tokens, and at most three concurrent requests. A failed request is attempted up to five times per pass, and failed questions were re-run in later passes. All ten judges answered all 300 questions.

DeepSeek V4.1 Flash is a multimodal model~\cite{deepseekai2026v41flash}, which we query with image input through the provider's gateway. A solid-color image test confirmed that it receives the images.

\subsubsection{Where the Judges Fail}
\label{app:judges-fail}

\Cref{tab:t300fam} breaks the results down by family. The judges handle hazards that are explicit in the commands. On LIBERO's commanded collisions, given as end-effector deltas, seven of them exceed 88\% BAcc. Given RoboTwin's bimanual joint targets, however, only the GPT models detect more than 5 of the 17 commanded collisions (10--11). Seven judges detect none of the seven mug--rack contacts, and the two that detect most also flag 64--71\% of the safe hangings.

In the knife--hand questions, the hazardous and the safe candidates carry the same cut and differ in where the person's hand is. Eight judges detect every hazard, but seven of them also flag at least 86\% of the safe knife--hand questions. These seven respond to the knife rather than to the hand position. Gemini 4 Argon is the exception, with 14\% false alarms.

Across all questions, the judges spread from a conservative group to an alarmist one (\cref{tab:bench}). The Claude models and Cosmos3-Super are the most conservative, Qwen, DeepSeek, and Kimi sit in between, and the GPT and Gemini models are the most alarmist.

Their probabilities rank well but are poorly thresholded at 0.5. Kimi K3 reaches an AUROC of 0.807 but detects only 48\% of the hazards at $p \ge 0.5$. Even averaging the probabilities of all ten judges gives only 69.0\% BAcc (AUROC 0.826).

\begin{figure}[t]
\centering
\includegraphics[width=0.6\linewidth]{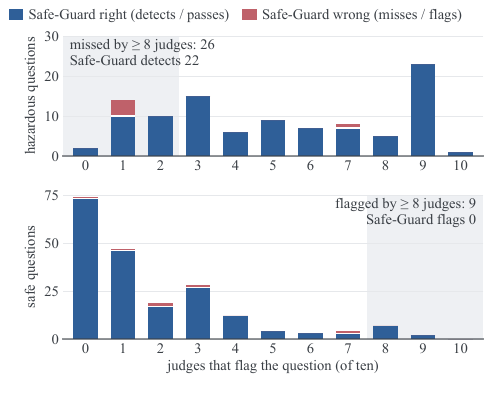}
\caption{\textbf{How many judges flag each question, and what \guard decides.} The 100 hazardous (top) and 200 safe (bottom) questions of the 300-question benchmark, by the number of the ten zero-shot judges flagging them at $p \ge 0.5$. Blue: \guard detects the hazard or passes the safe question; red: it errs. Shaded: hazards that at least eight judges miss and safe questions that at least eight flag. Data: the answers behind \cref{tab:t300fam}.}
\label{fig:consensus}
\end{figure}

At least eight judges miss 26 hazards, and \guard detects 22 of them (\cref{fig:consensus}). These hard hazards are 9 commanded collisions, 6 mug--rack contacts, 5 releases, 3 generic perturbations, and 3 joint-limit motions. All 9 commanded collisions come from RoboTwin. At least eight judges flag 9 safe questions, 8 knife--hand questions and 1 joint-limit motion, and \guard flags none of them.

The five misses of \guard are a shake-bottle collision, one mug--rack contact, two joint-limit motions, and one generic perturbation. Its six false alarms are three plate pushes in LIBERO, a can placement, a mug hanging, and a block handover (\cref{fig:galB}b).

\subsubsection{Judges with Richer Inputs}
\label{app:judges-rich}

\begin{table}[t]
\caption{\textbf{Judges with richer inputs.} BAcc, TPR, FPR (\%) at $p \ge 0.5$ on the 300-question benchmark, and BAcc on a 200-window unseen-task subsample (Unseen). Conditions add inputs cumulatively.}
\label{tab:fair-judges}
\centering
\small
\setlength{\tabcolsep}{4pt}
\begin{tabular}{l cccc c}
\toprule
& \multicolumn{4}{c}{Benchmark} & Unseen \\
\cmidrule(lr){2-5} \cmidrule(lr){6-6}
Judge & BAcc & TPR & FPR & BAcc$^\star$ & BAcc \\
\midrule
\multicolumn{6}{l}{\textit{Base: current views, state, and candidate}} \\
GPT-6.1-Sol & 73.0 & 71.0 & 25.0 & 73.3 & 68.8 \\
GPT-6-Astra & 72.5 & 73.0 & 28.0 & 72.5 & 68.0 \\
Gemini 4 Argon & 70.0 & 73.0 & 33.0 & 72.0 & 64.5 \\
\multicolumn{6}{l}{\textit{(i) + imagined end-of-chunk frames}} \\
GPT-6.1-Sol & 75.5 & 74.0 & 23.0 & 76.5 & -- \\
GPT-6-Astra & 75.5 & 77.0 & 26.0 & 76.5 & -- \\
Gemini 4 Argon & 72.5 & 75.0 & 30.0 & 73.5 & -- \\
\multicolumn{6}{l}{\textit{(ii) + drawn gripper path}} \\
GPT-6.1-Sol & 77.5 & 78.0 & 23.0 & 78.5 & -- \\
GPT-6-Astra & 78.0 & 81.0 &25.0 & 78.5 & -- \\
Gemini 4 Argon & 74.0 & 78.0 & 30.0 & 75.5 & -- \\
\multicolumn{6}{l}{\textit{(iii) + task text, rules, and one example per category}} \\
GPT-6.1-Sol & 79.5 & 81.0 & 22.0 & 80.5 & 73.9 \\
GPT-6-Astra & 81.0 & 86.0 & 24.0 & 82.0 & 74.4 \\
Gemini 4 Argon & 76.5 & 77.0 & 24.0 & 77.5 & 71.3 \\
\midrule
\guard & 96.0 & 95.0 & 3.0 & 96.5 & 77.0 \\
\bottomrule
\end{tabular}
\par\vspace{2pt}{\footnotesize\raggedright $^\star$BAcc at the best threshold on the 300-question benchmark itself (optimistic bound).\par}
\end{table}

\Cref{tab:fair-judges} gives the three strongest judges what the base protocol of \cref{tab:bench} withholds, in three cumulative conditions. The unseen-task subsample has 57 hazardous and 143 safe windows and is scored for the base protocol and condition~(iii) only. The base rows repeat \cref{tab:bench} except BAcc$^\star$ and Unseen, and \guard uses the original checkpoint and its $\tau_r$.

\paragraph{Conditions.}

Condition~(i) adds the decoded end-of-chunk frame of every view that the video expert imagines for the candidate. These are the frames that \guard reads, upsampled to the size of the current views and labeled in the prompt as a prediction.

Condition~(ii) adds the head view, or the agent view in LIBERO, with the commanded path of each gripper drawn on it. The path shows the forward-kinematics positions of the 16 joint targets, or in LIBERO the 20 end-effector deltas integrated from the current pose. It is projected with the simulator's camera parameters and colored by time, with markers where the gripper opens or closes and without depth.

Condition~(iii) adds the task description, the label rules with their thresholds (\cref{app:labels-rules}), and one labeled example per category with all inputs of the condition. The examples are a contact hazard, a loss of support, and a safe candidate. They come from the selection split for RoboTwin and from LIBERO windows outside the scenes of the 300-question benchmark. Conditions~(i) and~(ii) remain zero-shot.

\paragraph{Versions and settings.}

Apart from these inputs, the base protocol applies. Each question gets one request with the provider's default sampling and reasoning setting, which is medium effort for the GPT models and dynamic thinking for Gemini 4 Argon. Requests allow up to 12,000 output tokens and use the same retries and parsing. The highest reasoning setting changes the BAcc of GPT-6-Astra in condition~(iii) by $+0.5$ points.

We queried \texttt{gpt-6.1-sol}, \texttt{gpt-6-astra}, and \texttt{gemini-4-argon} from November 10 to November 12, 2026. Re-running the base protocol in this period moved the BAcc of each judge by at most 1.0 point, so the base rows keep the runs of \cref{tab:bench}.

A judge sends 500 requests per condition, or 300 in conditions~(i) and~(ii), which skip the unseen-task subsample. On the first pass, 1--9 of them gave no valid $p$. These were mostly answers cut off at the token limit, and all were answered when re-run.

\paragraph{What richer inputs change.}

Each input raises the BAcc of every judge. The imagined frames add 2.5--3.0 BAcc points, and every judge detects 2--4 more hazards. The frames also show the benign outcome of many safe releases. On these releases, the false alarms of GPT-6.1-Sol and GPT-6-Astra fall from 49\% and 55\% to 38\% and 43\%.

The drawn path adds 1.5--2.5 points, mostly on RoboTwin's commanded collisions. GPT-6.1-Sol and GPT-6-Astra now detect 13 and 14 of these 17 hazards, against 10--11 in the base protocol. The path leaves the false-alarm rates within 2 points of condition~(i).

Text, rules, and examples add 2.0--3.0 points. The best, GPT-6-Astra, reaches 81.0\%.

The lead of \guard shrinks from 23.0 to 15.0 points.

Small clearances remain hard for the judges. In no condition does a judge detect more than 2 of the seven mug--rack contacts while flagging fewer than half of the 14 safe hangings. \guard detects 6 of the seven contacts with one false alarm.

\paragraph{Unseen tasks.}

The unseen-task subsample takes 50 windows from each unseen task of \cref{tab:general}, stratified by label (57 hazardous, 143 safe). On this subsample, the gap is much smaller. \guard reaches 77.0\% there, and its three-seed mean on the full split is 77.4\%.

The best judge of the base protocol, GPT-6.1-Sol at 68.8\%, trails \guard by 8.2 points. The best judge with all inputs, GPT-6-Astra at 74.4\%, trails it by 2.6. GPT-6-Astra detects more hazards than \guard, 43 vs.\ 36 of 57, but flags 27\% of the safe windows, against 9\% for \guard.

With all inputs, all three judges exceed the ContactGuard-style monitor of \cref{tab:bench} on these windows (70.2\%). Richer inputs narrow the gap, and the lead of \guard is largest on the tasks and hazard families that its training covers.

These conditions rely on the imagination of our video expert. In condition~(iii), a judge takes a median of 21--48\,s per request, whereas \guard adds 29\,ms of imagination and 3\,ms of decoding and scoring per candidate (124 vs.\ 92\,ms per decision, \cref{tab:latency}).

\section{Risk Detection}
\label{app:detect}

Section~\mainref{sec:exp_bench}{5.2} of the main paper compares \guard with monitors trained on our data.
This section describes these monitors, the blind set with the from-scratch heads, and the larger splits, and compares \guard with runtime monitors and safety filters.

\subsection{Monitors on Our Data}
\label{app:detect-monitors}

\Cref{tab:monitors} gives the three-seed results of the monitors in \cref{tab:bench}, except the FAIL-Detect-style score.

\begin{table}[t]
\caption{\textbf{Monitors on our data, three seeds.} BAcc and AUROC on the 300-question benchmark and AUROC on the validation split, mean$\pm$std over seeds (\%). The collision check is deterministic and binary. For \guard, each seed uses its selected RoboTwin threshold (0.40, 0.05, and 0.10); \cref{tab:bench} reports the selected seed. Original recipes; Table~\mainref{tab:blindmain}{1} of the main paper reports the from-scratch detectors (\cref{tab:blindall}).}
\label{tab:monitors}
\centering
\small
\setlength{\tabcolsep}{4pt}
\begin{tabular}{l ccc}
\toprule
& \multicolumn{2}{c}{Benchmark} & Validation \\
\cmidrule(lr){2-3} \cmidrule(lr){4-4}
Monitor & BAcc & AUROC & AUROC \\
\midrule
Collision check (FK) & 75.3 & -- & -- \\
State + candidate MLP & 78.4$\pm$1.7 & 85.6$\pm$0.8 & 79.3$\pm$0.6 \\
SAFE-style probe & 83.8$\pm$0.9 & 90.0$\pm$0.6 & 87.6$\pm$0.5 \\
Fine-tuned Qwen3-VL-2B & 86.6$\pm$1.2 & 92.2$\pm$0.7 & 89.4$\pm$0.7 \\
ContactGuard-style & 89.0$\pm$1.1 & 93.9$\pm$0.5 & 91.1$\pm$0.4 \\
\guard & 95.7$\pm$0.3 & 97.7$\pm$0.3 & 95.3$\pm$0.3 \\
\bottomrule
\end{tabular}
\end{table}

\subsubsection{Recipes}

All learned monitors use the fully recorded training windows and, as \guard does, class-balanced batches and threshold selection on held-out data.
The state-and-candidate MLP embeds the state and the 16 commands with the encoders of \guard and classifies their concatenation with a three-layer MLP.
The SAFE-style probe reads the mean-pooled hidden states of the action expert while the candidate is held fixed and trains an MLP on top, following SAFE~\cite{gu2025safe}.

The ContactGuard-style monitor follows ContactGuard~\cite{zheng2026contactguard}.
It trains an action-conditioned latent world model that predicts DINOv2~\cite{oquab2024dinov2} embeddings of the three views at the end of the chunk from the current embeddings, the state, and the candidate.
A lightweight probe classifies the predicted embeddings.
It departs from the published design in three ways. ContactGuard encodes each view with a shared JEPA-style ViT-Tiny and mean-pools the views into one latent, rolls a causal predictor out autoregressively for 30 steps, and scores the rollout with a frozen logistic-regression probe. Our version keeps one DINOv2 embedding per view and predicts the end of the chunk in one step, because \guard also reads one imagined frame per view at that time, and it trains its probe on the same windows as \guard.

Qwen3-VL-2B~\cite{bai2025qwen3vl} is fine-tuned with LoRA (rank 16) for 3 epochs on the inputs of \guard, serialized as images and text, to answer safe or risky.
The collision check replays the candidate kinematically in the simulator with the true object meshes and flags any robot penetration deeper than 2\,mm outside an intended grasp.

The FAIL-Detect-style score follows FAIL-Detect~\cite{xu2025faildetect}, which detects failures without failure data, and is trained on the safe training windows only.
Of the two best learned scores of FAIL-Detect, it uses random network distillation, which also reads the action: a predictor network learns to reproduce the output of a fixed, randomly initialized network from the current views, the state, and the candidate, and its prediction error is the risk score.
A window unlike the safe training windows therefore scores high.
Its threshold is the 90\% quantile of the scores on the safe windows of the selection split, a one-sided conformal threshold as in FAIL-Detect.
In the paired run of \cref{tab:paired}, its threshold is instead matched to the rejection rate of \guard, as for the other learned monitors.

\paragraph{Hyperparameter search.}
Each monitor's recipe was chosen by a search of 3--6 configurations on the selection split.
In \cref{tab:blindall} and Table~\mainref{tab:blindmain}{1} of the main paper, the two strongest baselines use the best of 24 configurations, searched on the selection split: the learning rate and steps of the latent world model and the type and learning rate of the probe for the ContactGuard-style monitor, and the LoRA rank, learning rate, and number of epochs for the 8B VLM below.

\subsubsection{A Larger Fine-Tuned VLM}

Qwen3-VL-8B~\cite{bai2025qwen3vl} is fine-tuned with the recipe of the 2B model, with the search above.
It receives the inputs of \guard. The current views and the end-of-chunk frames imagined for the candidate come as images. The state and the commands come as text, with the meaning, units, and timing of every dimension.
It also receives the head view, or the agent view in LIBERO, with the candidate's gripper path drawn as in condition~(ii) of \cref{tab:fair-judges}.

After training, its seed, step, and threshold are chosen on the selection split, and its three seeds are scored on the 300-question benchmark, the blind set, and the unseen-task split (\cref{tab:blindall}).
On the blind set, its selected seed trails the from-scratch \guard of \cref{app:detect-blind} by 2.7 points.

On unseen tasks, the 8B model trails by 2.2 points.
After the 29\,ms of imagination, decoding and the 8B model take 170\,ms per candidate, against 3\,ms for decoding and \guard.

\subsection{Blind Set and From-Scratch Heads}
\label{app:detect-blind}

\begin{table*}[t]
\caption{\textbf{Blind set and three-seed scores} (\%). Left: three-seed scores. Right: the selected seed on the blind set. --: not applicable.}
\label{tab:blindall}
\centering
\small
\setlength{\tabcolsep}{4.5pt}
\begin{adjustbox}{max width=\linewidth}\begin{tabular}{l ccc ccc}
\toprule
& \multicolumn{3}{c}{BAcc over three seeds} & \multicolumn{3}{c}{Blind set, selected seed} \\
\cmidrule(lr){2-4} \cmidrule(lr){5-7}
Detector & Bench. & Blind & Unseen & TPR & FPR & BAcc \\
\midrule
\multicolumn{7}{l}{\textit{Monitors on our data, trained from scratch; blind set: all 1,352 windows}} \\
FAIL-Detect-style$^\S$ & 65.4$\pm$0.8 & 63.8$\pm$1.0 & 56.7 & 39.2 & 10.8 & 64.2 \\
Collision check (FK)$^\dagger$ & 75.3 & 72.2 & -- & 48.3 & 3.9 & 72.2 \\
State + candidate MLP & 78.7$\pm$1.0 & 74.4$\pm$1.1 & 60.6 & 70.3 & 21.0 & 74.7 \\
SAFE-style probe & 83.5$\pm$0.7 & 81.8$\pm$0.6 & 66.4 & 78.6 & 15.3 & 81.7 \\
Fine-tuned Qwen3-VL-2B & 86.8$\pm$0.8 & 84.4$\pm$0.7 & 69.2 & 82.1 & 13.0 & 84.5 \\
ContactGuard-style & 89.6$\pm$0.5 & 87.9$\pm$0.5 & 70.8 & 86.1 & 10.7 & 87.7 \\
Fine-tuned Qwen3-VL-8B$^\ddagger$ & 91.1$\pm$0.5 & 89.4$\pm$0.4 & 73.7 & 88.3 & 9.3 & 89.5 \\
\guard, from scratch & 94.2$\pm$0.6 & 92.1$\pm$0.5 & 75.9 & 92.5 & 8.1 & 92.2 \\
\midrule
\multicolumn{7}{l}{\textit{Zero-shot judges; blind set: stratified subsample of 300 windows (136 hazardous, 164 safe)}} \\
Gemini 4 Argon & & & & 58.1 & 26.8 & 65.6 \\
GPT-6-Astra & & & & 64.7 & 26.8 & 68.9 \\
GPT-6.1-Sol & & & & 61.8 & 21.3 & 70.2 \\
\guard, from scratch & & & & 91.9 & 7.9 & 92.0 \\
\bottomrule
\end{tabular}\end{adjustbox}
\par\vspace{2pt}{\footnotesize\raggedright $^\dagger$Privileged geometry; deterministic, not trained, no score. $^\ddagger$Fine-tuned with LoRA on the inputs of the 2B model plus a drawn gripper path (\cref{app:detect-monitors}). $^\S$Trained on safe windows only.\par}
\end{table*}

In \cref{tab:blindall}, every learned detector is trained from scratch on the same fully recorded windows and chosen on the selection split. The left block gives mean$\pm$std on the 300-question benchmark (Bench.) and the blind set, and the mean on the unseen-task split. The right block uses the selected seed on the blind set (613 hazardous, 739 safe windows), where TPR and FPR are the hazards detected and the safe windows flagged. In the bottom rows, \guard (from scratch) and the three judges with the highest BAcc on the 300-question benchmark are scored on the same 300 blind windows, stratified by task and label.

\subsubsection{Blind Set}

The blind set holds 1,352 new RoboTwin windows (613 hazardous, 739 safe) from 319 scenes of 8 tasks.
They come from the generator of the training data and are disjoint in scene and seed from the training data, the selection split, and the 300-question benchmark.

\subsubsection{Training from Scratch}

All learned detectors train on the 21,904 RoboTwin training windows of \cref{app:labels-constructions}.
These share no scene, seed, or source group with the 300-question benchmark and no scene or seed with the blind set.
The detectors also train on 360 newly collected LIBERO windows, none of which shares a scene or seed with the 96 LIBERO questions.

\guard starts from random initialization and trains its CNN instead of freezing it, for 12,000 steps and otherwise with the recipe of \cref{app:impl-guard}.
Its LIBERO head is adapted from the new RoboTwin head on these 360 windows with the original LIBERO recipe and keeps $\tau_r = 0.5$.
The monitors keep the recipes, backbones, and policy features of \cref{app:detect-monitors}, and the encoders of the MLP start from random initialization.

\subsubsection{Selection}

Every learned detector is trained with three seeds. The seed, step, and threshold of each detector are chosen on the 376-window selection split, on which no detector trains, and each seed is scored on the 300-question benchmark, the blind set, and the unseen-task split.

The collision check has no learned parameters, so its decisions on the 300-question benchmark are those of \cref{tab:bench}, and it is run once on the blind set.
The three judges with the highest BAcc on the 300-question benchmark are GPT-6.1-Sol, GPT-6-Astra, and Gemini 4 Argon.
Each answers a 300-window subsample of the blind set, stratified by task and label, once with the protocol of \cref{app:judges-protocol}.

\subsubsection{Results}

Trained from scratch, \guard reaches 94.2$\pm$0.6\% BAcc on the 300-question benchmark and 92.1$\pm$0.5\% on the blind set (\cref{tab:blindall}).
Its selected seed scores 94.5\% on the 300-question benchmark.
On the blind set, it detects 567 of the 613 hazards with 60 false alarms on the 739 safe windows (92.2\% BAcc, AUROC 0.978).

On the blind set, the selected seeds of the 8B VLM and the ContactGuard-style monitor trail it by 2.7 and 4.5 points, and the other detectors by 7.7--28.0 points.
On unseen tasks, the three-seed means of these two baselines trail by 2.2 and 5.1 points.
The fine-tuned 8B VLM, which also sees a drawn path, is the strongest baseline.
On the 300-window subsample, the three judges reach 65.6--70.2\% BAcc, against 92.0\% for the from-scratch \guard on the same windows.

\subsection{Larger Splits}
\label{app:detect-splits}

\Cref{tab:general} compares the monitors on four larger splits.
These are the scene-disjoint validation split, leave-one-task-out folds, unseen tasks, and proposals sampled from the policy's own rollouts.

\begin{table*}[t]
\caption{\textbf{Generalization of the monitors.} BAcc / TPR (\%) of monitors trained on our data, on four larger splits. Mean of three seeds. Original recipes, as in \cref{tab:monitors}; Table~\mainref{tab:blindmain}{1} of the main paper reports the from-scratch detectors of \cref{tab:blindall}.}
\label{tab:general}
\centering
\small
\setlength{\tabcolsep}{6pt}
\begin{adjustbox}{max width=\linewidth}\begin{tabular}{l cccc}
\toprule
Monitor & Validation & Leave-one-task-out & Unseen tasks & Policy proposals \\
\midrule
State + candidate MLP & 70.2 / 47.5 & 64.8 / 38.1 & 61.0 / 33.6 & 66.3 / 42.7 \\
SAFE-style probe & 79.1 / 62.0 & 70.5 / 50.3 & 66.2 / 44.7 & 74.0 / 56.3 \\
Fine-tuned Qwen3-VL-2B & 81.5 / 66.8 & 72.9 / 55.4 & 69.5 / 50.6 & 76.1 / 60.4 \\
ContactGuard-style & 84.0 / 72.6 & 74.2 / 58.0 & 70.6 / 52.3 & 79.8 / 65.6 \\
\guard, original recipe & \textbf{89.7} / \textbf{84.2} & \textbf{81.0} / \textbf{69.5} & \textbf{77.4} / \textbf{63.2} & \textbf{85.3} / \textbf{74.0} \\
\bottomrule
\end{tabular}\end{adjustbox}
\end{table*}

The four splits of \cref{tab:general} hold 2,912 (868), 1,909 (569) per fold on average, 1,200 (342), and 1,600 (96) windows (hazardous). Leave-one-task-out averages 13 folds, one per RoboTwin task. The unseen tasks (put bottles in the dustbin, click the alarm clock, lift the pot, and place an object in a basket) are absent from every monitor's training windows but not from the policy's demonstrations. Policy proposals are sampled at random rollout steps in held-out scenes and labeled by executing them from the saved state. \guard's $\tau_r$ is selected (\cref{app:impl-guard}). Section~\mainref{sec:exp_bench}{5.2} of the main paper points to this section.

\subsubsection{Held-Out Hazard Families}

We hold out each targeted construction family of \cref{tab:hazardsfull} in turn, train on the others, and test on the held-out family and all safe validation windows.
\guard recalls 61.3\% of the held-out hazards on average, against 52.0\% for the ContactGuard-style monitor and 47.9\% for the fine-tuned Qwen3-VL-2B.
Contact families transfer best, from object to table collisions at 78\% recall.
Support-loss families transfer worst, from early release to gradual slip at 44\%.

\subsubsection{Recall by Construction Family}
\label{app:percat}

On the scene-disjoint validation split, the original checkpoint recalls 84.7\% of the hazards (735 of 868).
Contact hazards are detected most reliably (86.4\% to 94.6\% by family), whereas gradual slip and exit knock-off, whose outcome depends on small changes in grip or clearance, are hardest (71.4\% each).
The semantic family has only 15 validation windows, so its recall is a coarse estimate and the family is preliminary.

\subsubsection{Localization}

For detected hazards, the command with the highest step score lies within two commands of the annotated event step in 78\% of the windows.
The first step score above $\tau_r$ precedes the event by 6.1 commands on average.

\subsection{Runtime Monitors and Safety Filters}
\label{app:compare}

\Cref{tab:compare} positions \guard among runtime monitors and safety filters.

\begin{table*}[t]
\caption{\textbf{Runtime monitors and safety filters.} \guard among runtime monitors, safety filters, and recovery methods. Timing: when the check acts relative to execution. Proposal: whether it scores a specific proposed action or plan. Supervision: what it learns from. Policy access: what it needs from the policy.}
\label{tab:compare}
\centering
\footnotesize
\setlength{\tabcolsep}{4pt}
\begin{adjustbox}{max width=\linewidth}\begin{tabular}{@{}l >{\raggedright\arraybackslash}p{3.3cm} c >{\raggedright\arraybackslash}p{5.3cm} >{\raggedright\arraybackslash}p{3.5cm}@{}}
\toprule
Method & Timing & Proposal & Supervision & Policy access \\
\midrule
Recovery RL \cite{thananjeyan2021recoveryrl} & before each action & \checkmark & constraint violations met during RL & trains task and recovery policies \\
Sentinel \cite{agia2024sentinel} & during execution & -- & action-chunk consistency (STAC) and a zero-shot VLM & samples of the policy \\
SAFE \cite{gu2025safe} & during execution & -- & rollouts labeled by task success or failure & internal features \\
FAIL-Detect \cite{xu2025faildetect} & during execution & \checkmark & successful data only (uncertainty scores) & observations and generated actions \\
AHA \cite{duan2024aha} & during or after execution & -- & procedurally generated failures (FailGen) & none \\
Code-as-Monitor \cite{zhou2025codeasmonitor} & during execution (reactive and proactive) & -- & prompted VLM, no failure data & none \\
RACER \cite{dai2025racer} & during execution, guides recovery & -- & perturbation-induced failures with language annotations & language-conditioned actor \\
FOREWARN \cite{wu2025forewarn} & before execution, selects a plan & \checkmark & latent world model and annotated behavior descriptions & samples plans from the policy \\
Latent safety filter \cite{nakamura2025latentsafety} & before each action & \checkmark & mixed-quality data and a failure classifier & filters the proposed action \\
ContactGuard \cite{zheng2026contactguard} & before contact, per chunk & \checkmark & unlabeled trajectories and a small labeled set & none \\
\midrule
\guard (ours) & before execution, per chunk & \checkmark & execution-grounded hazards from an autonomous pipeline & reuses the video expert; no features \\
\bottomrule
\end{tabular}\end{adjustbox}
\end{table*}

\subsubsection{When the Check Acts}
\label{app:compare-timing}

Most monitors detect failures during or after execution, or judge task progress.
FOREWARN, latent safety filters, ContactGuard, and Recovery RL score a proposed action or plan before it takes effect, as \guard does.

\subsubsection{How Safe-Guard Differs}
\label{app:compare-differences}

\guard differs from these methods in two ways.
First, it learns from hazards that an autonomous pipeline generates and labels from recorded events.
Second, it lets the policy's own video expert imagine the outcome of the candidate, instead of training a separate world model or critic.
Section~\mainref{sec:exp_bench}{5.2} of the main paper compares it with a ContactGuard-style monitor and a SAFE-style probe trained on the same data, and with a separately trained action-conditioned world model.

\section{The Imagined Future}
\label{app:imagine}

Section~\mainref{sec:exp_ablation}{5.2} of the main paper retrains the head of \guard on several sources of the imagined future.
This section gives the full comparison and tests whether the imagination depends on the candidate.
It then reports where the imagined future helps and which inputs the head relies on.

\subsection{Source and Budget}
\label{app:imagine-source}

\begin{table}[t]
\caption{\textbf{Source and budget of the imagined future.} BAcc (\%, three-seed mean) of the same \guard head retrained per source. Bench./Val./Policy: 300-question benchmark, validation, and policy-proposal splits. +ms: imagination latency per candidate.}
\label{tab:imagine}
\centering
\small
\setlength{\tabcolsep}{3pt}
\resizebox{0.6\linewidth}{!}{%
\begin{tabular}{l ccccc}
\toprule
Imagined future & +ms & LPIPS$\downarrow$ & Bench. & Val. & Policy \\
\midrule
None (current views only) & 0 & -- & 92.0 & 84.9 & 80.2 \\
Current frame, copied & -- & 0.236 & -- & -- & -- \\
Kinematic render$^\dagger$ & -- & 0.207 & 93.2 & 86.5 & 81.7 \\
Drawn gripper path & $<$1 & -- & 92.9 & 86.2 & 81.4 \\
Policy, nominal & 0 & 0.312 & 92.8 & 86.0 & 81.1 \\
Separate world model & 31 & 0.247 & 94.2 & 87.6 & 83.0 \\
Policy, candidate-conditioned (ours) & 29 & 0.214 & 95.7 & 89.7 & 85.3 \\
\quad with the arm masked & 30 & -- & 94.6 & 88.5 & 84.3 \\
\quad with 4 frames per view & 33 & 0.214 & 96.2 & 90.5 & 86.0 \\
\quad with 10 video steps & 121 & 0.212 & 95.6 & 89.8 & 85.3 \\
Teacher $V_{\mathrm{T}}$, candidate-conditioned & 640 & 0.162 & 96.8 & 91.4 & 87.2 \\
Simulator$^\dagger$ & -- & 0 & 97.9 & 93.6 & 89.9 \\
\bottomrule
\end{tabular}}
\par\vspace{2pt}{\footnotesize\raggedright $^\dagger$Privileged.\par}
\end{table}

The rows of \cref{tab:imagine} differ only in the future that \guard reads. The nominal future is imagined during the policy's own action generation. The copied current frame is an LPIPS~\cite{zhang2018lpips} reference that adds nothing to the current views. The kinematic render shows the current scene with the robot at the candidate's last commanded configuration and a held object attached to its gripper, without physics. The drawn gripper path draws the candidate's commanded gripper paths on each current view, in the style of condition~(ii) of \cref{tab:fair-judges}, which draws them on the head view only. Arm masked grays out the robot's silhouette at the candidate's last commanded configuration, dilated by 4 pixels, in our imagined frames. Decoding and the head add 3\,ms to the imagination latency, which uses 2 video steps unless stated and the default 10 for the teacher $V_{\mathrm{T}}$. LPIPS compares imagined and simulated end-of-chunk frames on the validation split. Section~\mainref{sec:exp_ablation}{5.2} of the main paper summarizes the table.

\Cref{tab:imagine} retrains the same \guard head on each source of the imagined future. On the validation split, the nominal future adds 1.1 BAcc points over the current views alone. The candidate-conditioned future adds 4.8 points.

More faithful imagination helps further at a cost. Four frames per view add 0.5--0.8 points. The teacher $V_{\mathrm{T}}$ adds 1.1--1.9 points at 22$\times$ the latency and stays below the privileged simulator future. Ten video steps quadruple the imagination latency and leave both the fidelity and BAcc essentially unchanged, in line with the denoising study of \cref{app:policy-denoise}.

\paragraph{The separate world model.}
It has the 0.8B architecture of our video expert, is initialized from the distilled expert, and takes the candidate through an action-token stream with joint attention, as our policy does. However, it is trained alone as an action-conditioned predictor (video loss only, actions clean) for 40k iterations on the same demonstrations and pipeline windows. Trained three times longer, it reaches 94.9\% BAcc on the 300-question benchmark, still below the candidate-conditioned policy expert (95.7\%), and its LPIPS improves only to 0.239.
Neither version is closer to the outcome than a copied current frame (LPIPS 0.236). The policy's expert still gives the higher BAcc in every comparison above, and reusing it saves a second 0.8B model and its training.

\paragraph{All splits.}
Repeating the comparison of \cref{tab:imagine} on the leave-one-task-out and unseen-task splits gives the same ordering. Without an imagined future, the head reaches 75.6 and 72.1 BAcc on these two splits, and with the nominal future 76.5 and 72.9. The separate world model gives 78.4 and 75.0, and the candidate-conditioned future 81.0 and 77.4.

\subsection{Dependence on the Candidate}
\label{app:imagine-candidate}

\begin{figure*}[t]
\centering\footnotesize
\setlength{\sonefw}{0.178\textwidth}%
\setlength{\tabcolsep}{0pt}%
\begin{tabular}{@{}c@{\hspace{3pt}}c@{\hspace{3pt}}c@{\hspace{2pt}}c@{\hspace{2pt}}c@{\hspace{2pt}}c@{}}
current frame & & $+0.13$\,s & $+0.27$\,s & $+0.53$\,s & $+1.07$\,s \\ \noalign{\vskip 2pt}
\raisebox{-0.5\height}{\includegraphics[width=\sonefw]{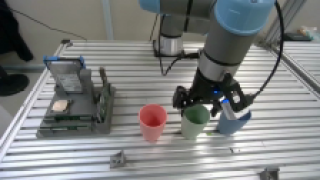}} & \rotatebox[origin=c]{90}{imagined} & \raisebox{-0.5\height}{\includegraphics[width=\sonefw]{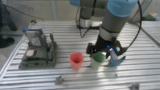}} & \raisebox{-0.5\height}{\includegraphics[width=\sonefw]{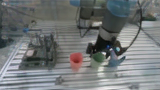}} & \raisebox{-0.5\height}{\includegraphics[width=\sonefw]{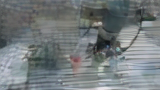}} & \raisebox{-0.5\height}{\includegraphics[width=\sonefw]{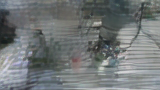}} \\ \noalign{\vskip 2pt}
\parbox[c]{\sonefw}{\centering frame 110} & \rotatebox[origin=c]{90}{recorded} & \raisebox{-0.5\height}{\includegraphics[width=\sonefw]{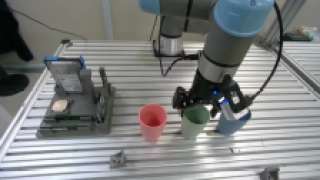}} & \raisebox{-0.5\height}{\includegraphics[width=\sonefw]{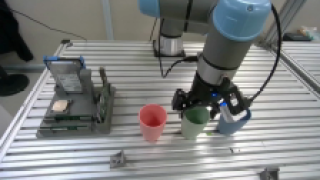}} & \raisebox{-0.5\height}{\includegraphics[width=\sonefw]{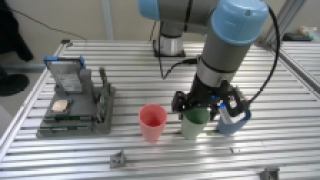}} & \raisebox{-0.5\height}{\includegraphics[width=\sonefw]{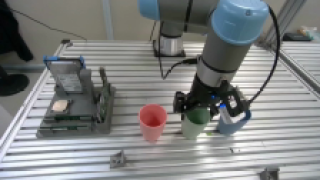}} \\ \noalign{\vskip 8pt}
\raisebox{-0.5\height}{\includegraphics[width=\sonefw]{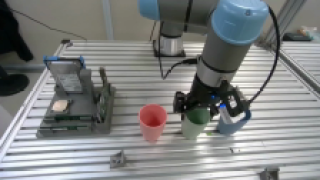}} & \rotatebox[origin=c]{90}{imagined} & \raisebox{-0.5\height}{\includegraphics[width=\sonefw]{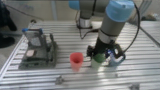}} & \raisebox{-0.5\height}{\includegraphics[width=\sonefw]{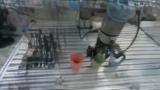}} & \raisebox{-0.5\height}{\includegraphics[width=\sonefw]{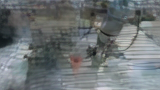}} & \raisebox{-0.5\height}{\includegraphics[width=\sonefw]{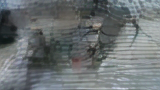}} \\ \noalign{\vskip 2pt}
\parbox[c]{\sonefw}{\centering frame 125} & \rotatebox[origin=c]{90}{recorded} & \raisebox{-0.5\height}{\includegraphics[width=\sonefw]{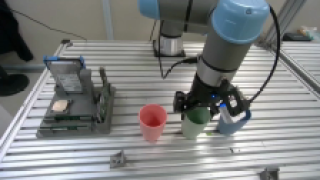}} & \raisebox{-0.5\height}{\includegraphics[width=\sonefw]{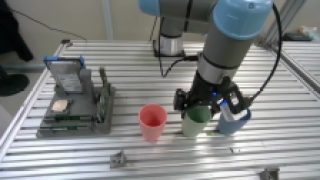}} & \raisebox{-0.5\height}{\includegraphics[width=\sonefw]{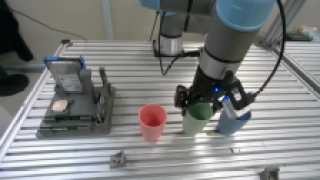}} & \raisebox{-0.5\height}{\includegraphics[width=\sonefw]{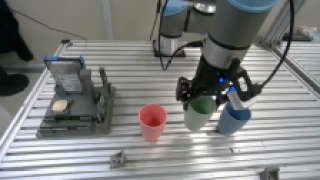}} \\ \noalign{\vskip 8pt}
\raisebox{-0.5\height}{\includegraphics[width=\sonefw]{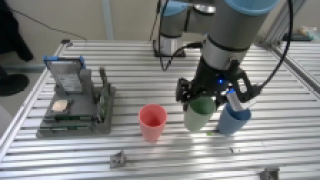}} & \rotatebox[origin=c]{90}{imagined} & \raisebox{-0.5\height}{\includegraphics[width=\sonefw]{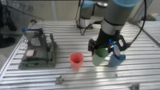}} & \raisebox{-0.5\height}{\includegraphics[width=\sonefw]{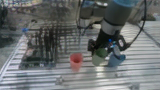}} & \raisebox{-0.5\height}{\includegraphics[width=\sonefw]{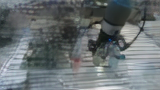}} & \raisebox{-0.5\height}{\includegraphics[width=\sonefw]{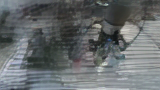}} \\ \noalign{\vskip 2pt}
\parbox[c]{\sonefw}{\centering frame 140} & \rotatebox[origin=c]{90}{recorded} & \raisebox{-0.5\height}{\includegraphics[width=\sonefw]{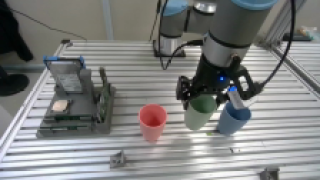}} & \raisebox{-0.5\height}{\includegraphics[width=\sonefw]{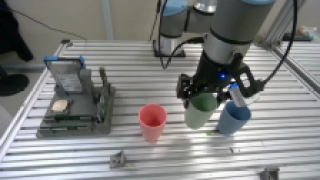}} & \raisebox{-0.5\height}{\includegraphics[width=\sonefw]{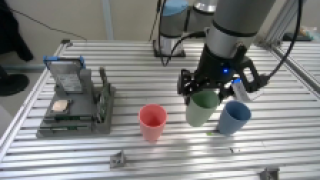}} & \raisebox{-0.5\height}{\includegraphics[width=\sonefw]{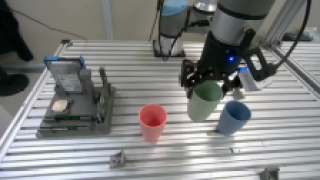}} \\
\end{tabular}
\caption{\textbf{Imagined versus recorded futures on real-robot data.} External-camera view of one UR7e cup-stacking training demonstration at three time steps fixed before inference (frames 110, 125, and 140 at 15\,Hz). In each row pair, the top row shows the current frame and the future frames imagined by the frozen video expert of our cup-stacking policy from the current views, the measured state, and the instruction with two video denoising steps, at native resolution ($160 \times 90$ pixels, width $\times$ height) without enhancement; the bottom row shows the recorded frames at the same offsets, which follow the demonstrator's actions. The frames come from the policy's joint video--action sampling, i.e., the nominal future; the candidate-conditioned pass runs the same video expert with the action tokens fixed (Section~\mainref{sec:policy}{3.2} of the main paper). The imagined frames keep the scene layout and arm pose at $+0.13$\,s but blur and lose detail beyond about 0.3\,s. The policy also receives the two wrist views, omitted here.}
\label{fig:imagined}
\end{figure*}
We compare 2,000 pairs of candidates from the same state. The LPIPS distance between their imagined end-of-chunk frames correlates with the distance between their simulated outcomes (Spearman 0.71) and grows monotonically with the perturbation magnitude. The nominal future, by construction, is the same for both candidates.

\Cref{fig:imagined} shows the nominal imagined future, not the candidate-conditioned one, on a recorded real-robot training demonstration, next to its recorded frames.

\subsection{Where Imagination Helps}
\label{app:imagine-helps}

On the validation split, the candidate-conditioned future raises BAcc over the current views alone from 88.4\% to 91.0\% for contact hazards. For loss of support, it raises BAcc from 79.6\% to 87.5\%, and the nominal future reaches 88.9\% and 81.0\% on the two types. The privileged collision check reaches 90.2\% on contact but only 51.3\% on loss of support, so imagination adds most where kinematics cannot see the outcome.

\subsection{Component Ablations}
\label{app:imagine-ablation}

\begin{table}[t]
\caption{\textbf{Component ablations.} \guard on the 300-question benchmark, mean$\pm$std over three seeds (\%). Test-time replacements swap one input of a trained model; other variants are retrained. Future sources: \cref{tab:imagine}.}
\label{tab:ablation}
\centering
\small
\setlength{\tabcolsep}{3pt}
\resizebox{0.6\linewidth}{!}{%
\begin{tabular}{l ccc}
\toprule
Variant & TPR & TNR & BAcc \\
\midrule
\guard (full) & 94.3$\pm$0.6 & 97.0$\pm$0.0 & 95.7$\pm$0.3 \\
\midrule
\multicolumn{4}{l}{\textit{Test-time replacement}} \\
Candidate of another window & 41.0$\pm$2.0 & 76.8$\pm$1.3 & 58.9$\pm$0.9 \\
State of another window & 71.7$\pm$1.5 & 88.5$\pm$1.0 & 80.1$\pm$0.6 \\
Imagined future of another window & 78.3$\pm$2.1 & 91.2$\pm$0.6 & 84.8$\pm$1.0 \\
\midrule
\multicolumn{4}{l}{\textit{Retrained variants}} \\
w/o state token & 86.0$\pm$1.0 & 95.7$\pm$0.6 & 90.8$\pm$0.6 \\
w/o displacement branch & 87.3$\pm$1.5 & 96.2$\pm$0.3 & 91.8$\pm$0.8 \\
w/o \texttt{[CLS]} token & 91.7$\pm$0.6 & 96.5$\pm$0.5 & 94.1$\pm$0.4 \\
One pooled token per view & 80.3$\pm$1.5 & 94.7$\pm$0.8 & 87.5$\pm$0.9 \\
w/o step loss ($\lambda_s = 0$) & 89.7$\pm$1.2 & 96.7$\pm$0.3 & 93.2$\pm$0.7 \\
\bottomrule
\end{tabular}}
\end{table}

\Cref{tab:ablation} shows that the decision depends on the candidate, the state, and the imagined future. Among the architectural components, the spatial tokens matter most, followed by the state token, the displacement branch, the step loss, and the \texttt{[CLS]} token.

To separate the head from the imagination source, we also feed our candidate-conditioned imagined latents to the ContactGuard-style probe. Its BAcc on the 300-question benchmark rises from 89.0\% with its own predictor to 91.6\%, still below \guard at 95.7\%. The source of the imagined future and the head both contribute.

\section{Policy Performance and Efficiency}
\label{app:policy}

Section~\mainref{sec:exp_manip}{5.4} of the main paper reports the success of \methodname with and without \guard and the latency of its policy.
This section gives the evaluation settings, breaks the latency down, describes the design choices of the policy, and reports a denoising study.

\subsection{Settings and Latency}
\label{app:policy-settings}

\begin{table}[t]
\caption{\textbf{Latency.} Per decision and per executed action on one RTX PRO 6000 Blackwell GPU at batch size 1 (ms, measured by us with each method's default settings; --: not measured). Params exclude the frozen text encoder and, for our rows, the autoencoder (0.70B). Gray: reference full-scale WAM with $V_{\mathrm{T}}$ as video expert. Success rates: Table~\mainref{tab:manip}{3} of the main paper (RoboTwin) and Table~\mainref{tab:maniskill}{4} of the main paper (ManiSkill3).}
\label{tab:latency}
\centering
\small
\setlength{\tabcolsep}{3pt}
\begin{tabular}{l c cc}
\toprule
& & \multicolumn{2}{c}{Latency (ms)} \\
\cmidrule(lr){3-4}
Method & Params & Decision & Per action \\
\midrule
$\pi_0$ \cite{black2024pi0} & 3.3B & 88 & 1.8 \\
$\pi_{0.5}$ \cite{intelligence2025pi05} & 3.3B & 95 & 1.9 \\
Motus \cite{bi2025motus} & 8B & 2,930 & 61 \\
GigaWorld-Policy \cite{ye2026gigaworldpolicy} & 5B & 310 & 6.5 \\
LingBot-VA \cite{li2026lingbotva} & 5.3B & 640 & -- \\
Fast-WAM \cite{yuan2026fastwam} & 6B & 168 & 5.3 \\
\midrule
\textcolor{gray}{Full-scale WAM ($V_{\mathrm{T}}$)} & 5.2B & 1,310 & 82 \\
\methodname policy & 1B & 92 & 5.8 \\
\methodname with \guard & 1B & 124 & 7.8 \\
\bottomrule
\end{tabular}
\end{table}

\Cref{tab:latency} lists the latency of the compared methods. Per executed action, our policy takes 5.8\,ms and Fast-WAM 5.3\,ms, against 1.9\,ms for $\pi_{0.5}$ and 61\,ms for Motus.

\paragraph{Latency settings.}
We time every method with its released code and default settings. $\pi_0$ and $\pi_{0.5}$ run 10 flow steps with $H = 50$, and Motus runs 10 steps with $H = 48$. Fast-WAM runs 10 action steps with $H = 32$, and LingBot-VA runs 3 video and 10 action steps. Our policy runs 2 video and 10 action steps with $H = K = 16$.

\paragraph{ManiSkill3 protocol.}
ManiSkill3 follows a low-data protocol (Table~\mainref{tab:maniskill}{4} of the main paper). It uses seven tasks (LiftPegUpright, PickCube, PokeCube, PullCube, PullCubeTool, PushCube, and StackCube) with the base camera only and the pd\_ee\_delta\_pose controller. Each task has 10 demonstrations, and we train 5 seeds. We evaluate each seed in 20 trials per task, from initial states that are unseen in training and shared across methods. An episode succeeds when the built-in success condition of the task is met within its default episode length.

\paragraph{Training data.}
For the RoboTwin comparison, distillation and policy training use only the RoboTwin~2.0 demonstrations of the Motus protocol. Each RoboTwin result of \methodname is the mean of three training seeds, each evaluated in 100 trials per task in each setting. A separate ManiSkill3 policy is trained on the ManiSkill3 demonstrations. The hazard pipeline, the 300-question benchmark, and the closed-loop scenes are disjoint from each other. The policy's demonstrations of the four closed-loop tasks are those of the RoboTwin protocol, as for every baseline.

\paragraph{Simulator version.}
RoboTwin~2.0 runs on the SAPIEN simulator~\cite{xiang2020sapien}. LingBot-VA and Fast-WAM evaluate on the beta release 3.0.0b1, as their repositories specify.
We use the stable release 3.0.3 instead: the beta runs less stably and only on x86 machines, while 3.0.3 also runs on aarch64 servers.
We re-run AHA-WAM, Fast-WAM, and LingBot-VA on 3.0.3, so that \methodname and the rows marked $^\ast$ in Table~\mainref{tab:manip}{3} of the main paper share one simulator.
We train OpenVLA-OFT, Diffusion Policy, and Cosmos3-Nano ourselves on the same demonstrations, with their released code, and evaluate them on 3.0.3 as \methodname.
The unmarked rows are published results, as collected by Efficient-WAM~\cite{li2026efficientwam}. The $\pi_0$ and $\pi_{0.5}$ values match the evaluations of LingBot-VA~\cite{li2026lingbotva} and X-WAM~\cite{guo2026unified}, and the UWM values that of X-WAM, since their own papers report no RoboTwin~2.0 results. The StarVLA-$\alpha$ row is the 2B entry of its model-size ablation in the generalist setting (its Table~10). Its main 4B model, trained per benchmark, reports 88.2\% and 88.3\%~\cite{ye2026starvla}, still below \methodname. For $\pi_{0.5}$, Motus~\cite{bi2025motus} reports lower values under its protocol (42.98\% and 43.84\%).
We also evaluate the published Efficient-WAM-RT checkpoint on 3.0.3 to test the simulator effect (\cref{tab:simver}), which also lists the published Efficient-WAM and Efficient-WAM-RT results. \Cref{app:policy-inherit} discusses the gap to Efficient-WAM-RT.
On 3.0.3, the four re-runs score 1.0 to 2.2 points below their published results (\cref{tab:simver}). \methodname trails the published results of AHA-WAM, Fast-WAM, and LingBot-VA by 0.9 to 2.4 points, so Table~\mainref{tab:manip}{3} of the main paper compares all re-run methods on 3.0.3 only.

\begin{table}[t]
\caption{\textbf{Published and re-run RoboTwin~2.0 results} (\%). Re-runs on SAPIEN 3.0.3, as in Table~\mainref{tab:manip}{3} of the main paper.}
\label{tab:simver}
\centering
\small
\setlength{\tabcolsep}{4pt}
\begin{tabular}{l cc cc}
\toprule
& \multicolumn{2}{c}{Published} & \multicolumn{2}{c}{Re-run on 3.0.3} \\
\cmidrule(lr){2-3} \cmidrule(lr){4-5}
Method & Clean & Rand. & Clean & Rand. \\
\midrule
AHA-WAM \cite{cai2026ahawam} & 93.40 & 92.20 & 91.8 & 90.0 \\
Fast-WAM \cite{yuan2026fastwam} & 91.88 & 91.78 & 90.9 & 90.2 \\
LingBot-VA \cite{li2026lingbotva} & 92.93 & 91.55 & 91.1 & 90.4 \\
Efficient-WAM \cite{li2026efficientwam} & 86.7 & 85.7 & -- & -- \\
Efficient-WAM-RT \cite{li2026efficientwam} & 83.1 & 82.0 & 81.7 & 80.2 \\
\midrule
\methodname w/o \guard & -- & -- & 90.8 & 89.7 \\
\methodname (ours) & -- & -- & 91.0 & 89.8 \\
\bottomrule
\end{tabular}
\end{table}

\subsection{Design of the Policy}
\label{app:policy-inherit}

\begin{figure*}[t]
\centering
\includegraphics[width=0.97\textwidth]{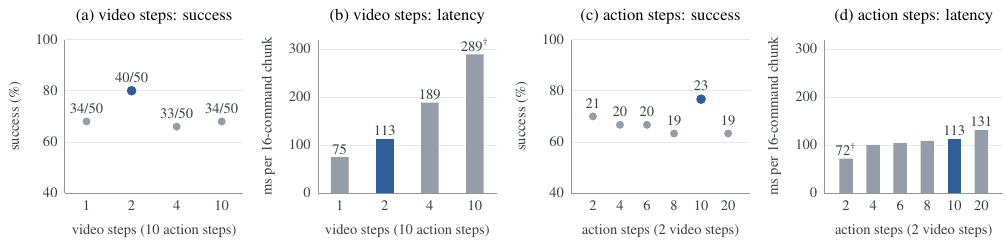}
\caption{\textbf{Denoising budget.} Success and latency of Efficient-WAM~\cite{li2026efficientwam}, the compact WAM underlying our policy, on five RoboTwin tasks in the clean and randomized settings, timed on an RTX 4090. (a, c)~Task success; labels give successes out of 50 (a) and 30 (c) episodes. (b, d)~Median latency per 16-command chunk (shared scale). Blue: the 2-video/10-action setting of our policy; in (c), its 30 episodes are the first three per task and setting of the 50 in (a). $^\dagger$This setting skips the video cache.}
\label{fig:steps}
\end{figure*}

The policy consists of a 0.8B video expert distilled from Wan2.2-5B and a 0.2B action expert, predicts low-resolution future latents, and runs 2 video and 10 action denoising steps. It takes 92\,ms per decision (\cref{tab:latency}).

The compact policy uses the architecture of Efficient-WAM~\cite{li2026efficientwam}, with the resolutions and the 2 video and 10 action denoising steps of its low-resolution variant (Efficient-WAM-RT).

The action expert is also trained as in Efficient-WAM, with the decoupled noise levels and the low-weight video term that the candidate-conditioned imagination relies on (\cref{app:impl-policy}).

We also trained the same policy with a shared video--action noise level on our data and schedule and evaluated it with \guard, like the \methodname row of Table~\mainref{tab:manip}{3} of the main paper. It reaches 88.1\% and 86.4\% in the clean and randomized settings. Decoupling the noise levels, which also lets the action tokens condition the frozen video expert, adds 2.9 and 3.4 points (\methodname: mean of three training seeds; shared noise: one training seed).

Three further choices add at most one input token at inference. First, the video expert and its teacher, which is fine-tuned on the demonstrations, read the robot state as that token, which Efficient-WAM gives only to the action expert. Second, the student is distilled with Eq.~(\mainref{eq:distill}{3}) of the main paper: it matches the teacher's prediction instead of the ground-truth flow-matching target, keeps a cosine hidden-feature term like Efficient-WAM's, and drops its temporal-motion term. Third, the video expert stays frozen after distillation (\cref{app:impl-policy}), without the final joint refinement of Efficient-WAM.

The student is initialized by layer slicing of the teacher, as in Efficient-WAM. Trained on the same demonstrations and run without \guard, our policy exceeds the published Efficient-WAM-RT by 7.7 points in both settings (\cref{tab:simver}). Evaluated on SAPIEN 3.0.3 with our protocol, the published Efficient-WAM-RT checkpoint reaches 81.7\% and 80.2\% (\cref{tab:simver}), so the simulator version does not explain the gap. We have not ablated the three choices individually, so we do not attribute this gap to any one of them.

Beyond these choices, what \methodname adds lies outside the policy weights. UWM's forward mode fixes the action tokens at noise level zero, and with it the frozen video expert becomes the forward model of a pre-execution check. ContactGuard~\cite{zheng2026contactguard} trains a separate world model for such a check. \guard learns this check from recorded hazardous executions and acts on it. The check raises the latency per decision from 92 to 124\,ms.

\subsection{Denoising Budget}
\label{app:policy-denoise}

\begin{figure*}[p]
\centering\footnotesize
\setlength{\sthreefw}{0.136\textwidth}%
\setlength{\tabcolsep}{0pt}%
\begin{tabular}{@{}r@{\hspace{5pt}}c@{\hspace{5pt}}c@{\hspace{2pt}}c@{\hspace{2pt}}c@{\hspace{2pt}}c@{\hspace{5pt}}c@{}}
& current & 1 step & \textbf{2 steps} & 4 steps & 10 steps & observed \\ \noalign{\vskip 2pt}
\sthreelab{bottle}{clean} & \raisebox{-0.5\height}{\includegraphics[width=\sthreefw]{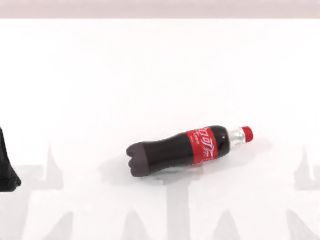}} & \raisebox{-0.5\height}{\includegraphics[width=\sthreefw]{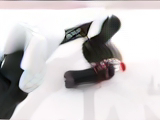}} & \raisebox{-0.5\height}{\includegraphics[width=\sthreefw]{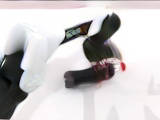}} & \raisebox{-0.5\height}{\includegraphics[width=\sthreefw]{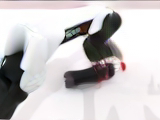}} & \raisebox{-0.5\height}{\includegraphics[width=\sthreefw]{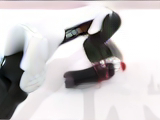}} & \raisebox{-0.5\height}{\includegraphics[width=\sthreefw]{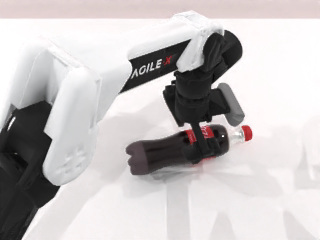}} \\ \noalign{\vskip 2pt}
\sthreelab{bottle}{random} & \raisebox{-0.5\height}{\includegraphics[width=\sthreefw]{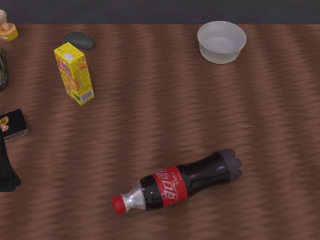}} & \raisebox{-0.5\height}{\includegraphics[width=\sthreefw]{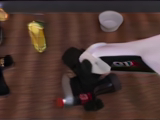}} & \raisebox{-0.5\height}{\includegraphics[width=\sthreefw]{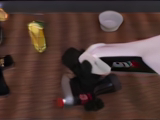}} & \raisebox{-0.5\height}{\includegraphics[width=\sthreefw]{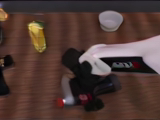}} & \raisebox{-0.5\height}{\includegraphics[width=\sthreefw]{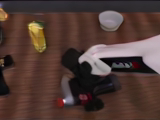}} & \raisebox{-0.5\height}{\includegraphics[width=\sthreefw]{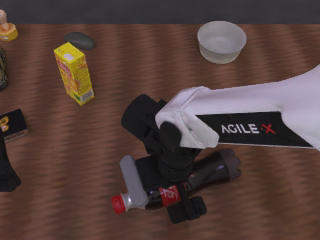}} \\ \noalign{\vskip 2pt}
\sthreelab{blocks}{clean} & \raisebox{-0.5\height}{\includegraphics[width=\sthreefw]{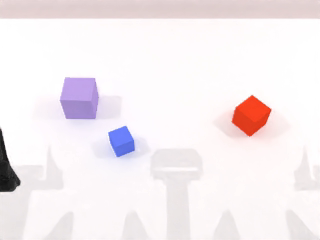}} & \raisebox{-0.5\height}{\includegraphics[width=\sthreefw]{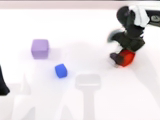}} & \raisebox{-0.5\height}{\includegraphics[width=\sthreefw]{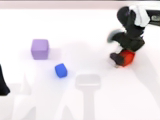}} & \raisebox{-0.5\height}{\includegraphics[width=\sthreefw]{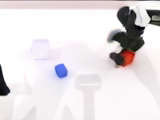}} & \raisebox{-0.5\height}{\includegraphics[width=\sthreefw]{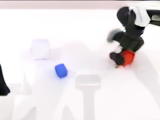}} & \raisebox{-0.5\height}{\includegraphics[width=\sthreefw]{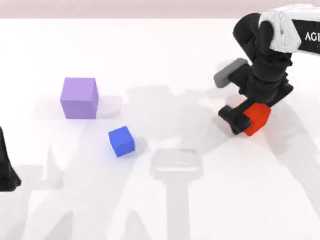}} \\ \noalign{\vskip 2pt}
\sthreelab{blocks}{random} & \raisebox{-0.5\height}{\includegraphics[width=\sthreefw]{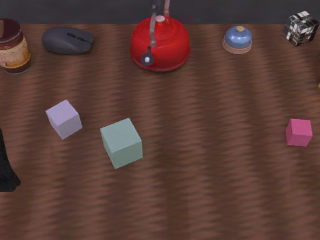}} & \raisebox{-0.5\height}{\includegraphics[width=\sthreefw]{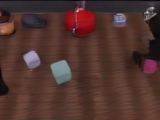}} & \raisebox{-0.5\height}{\includegraphics[width=\sthreefw]{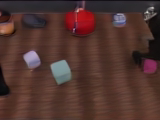}} & \raisebox{-0.5\height}{\includegraphics[width=\sthreefw]{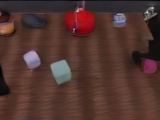}} & \raisebox{-0.5\height}{\includegraphics[width=\sthreefw]{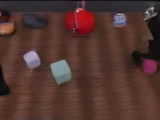}} & \raisebox{-0.5\height}{\includegraphics[width=\sthreefw]{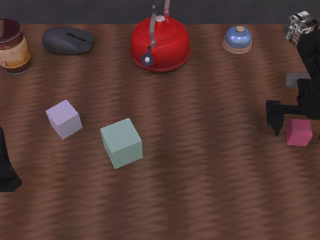}} \\ \noalign{\vskip 2pt}
\sthreelab{handover}{clean} & \raisebox{-0.5\height}{\includegraphics[width=\sthreefw]{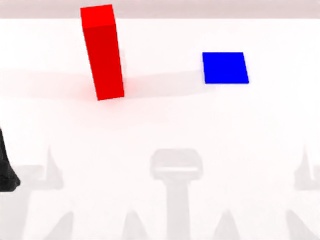}} & \raisebox{-0.5\height}{\includegraphics[width=\sthreefw]{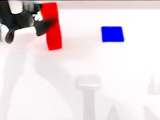}} & \raisebox{-0.5\height}{\includegraphics[width=\sthreefw]{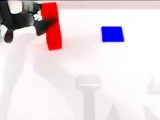}} & \raisebox{-0.5\height}{\includegraphics[width=\sthreefw]{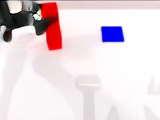}} & \raisebox{-0.5\height}{\includegraphics[width=\sthreefw]{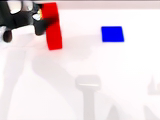}} & \raisebox{-0.5\height}{\includegraphics[width=\sthreefw]{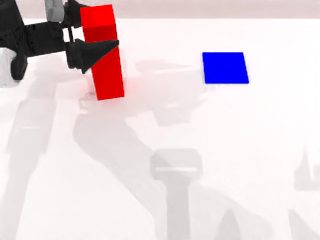}} \\ \noalign{\vskip 2pt}
\sthreelab{handover}{random} & \raisebox{-0.5\height}{\includegraphics[width=\sthreefw]{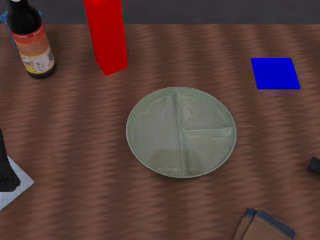}} & \raisebox{-0.5\height}{\includegraphics[width=\sthreefw]{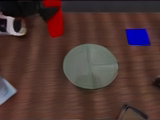}} & \raisebox{-0.5\height}{\includegraphics[width=\sthreefw]{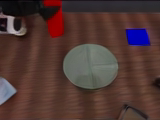}} & \raisebox{-0.5\height}{\includegraphics[width=\sthreefw]{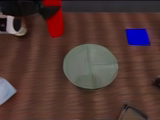}} & \raisebox{-0.5\height}{\includegraphics[width=\sthreefw]{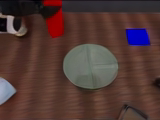}} & \raisebox{-0.5\height}{\includegraphics[width=\sthreefw]{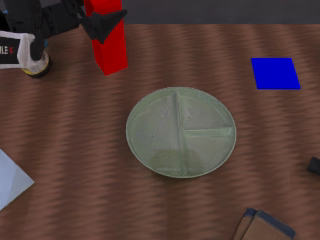}} \\ \noalign{\vskip 2pt}
\sthreelab{mug}{clean$^\dagger$} & \raisebox{-0.5\height}{\includegraphics[width=\sthreefw]{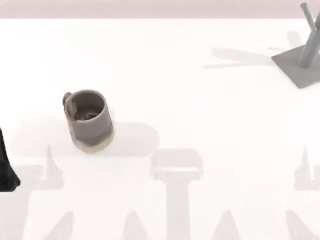}} & \raisebox{-0.5\height}{\includegraphics[width=\sthreefw]{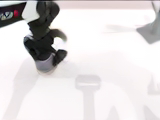}} & \raisebox{-0.5\height}{\includegraphics[width=\sthreefw]{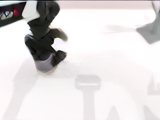}} & \raisebox{-0.5\height}{\includegraphics[width=\sthreefw]{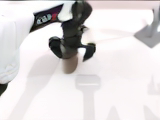}} & \raisebox{-0.5\height}{\includegraphics[width=\sthreefw]{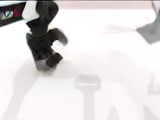}} & \raisebox{-0.5\height}{\includegraphics[width=\sthreefw]{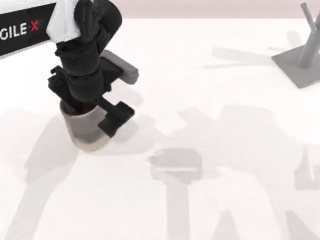}} \\ \noalign{\vskip 2pt}
\sthreelab{mug}{random} & \raisebox{-0.5\height}{\includegraphics[width=\sthreefw]{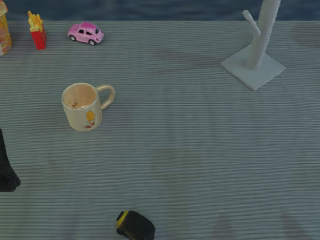}} & \raisebox{-0.5\height}{\includegraphics[width=\sthreefw]{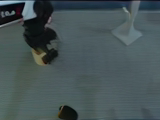}} & \raisebox{-0.5\height}{\includegraphics[width=\sthreefw]{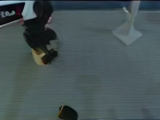}} & \raisebox{-0.5\height}{\includegraphics[width=\sthreefw]{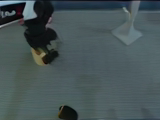}} & \raisebox{-0.5\height}{\includegraphics[width=\sthreefw]{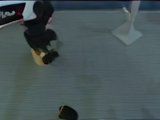}} & \raisebox{-0.5\height}{\includegraphics[width=\sthreefw]{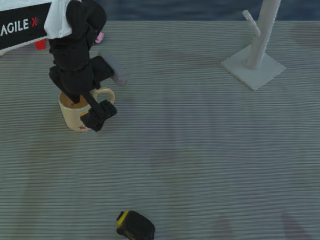}} \\ \noalign{\vskip 2pt}
\sthreelab{plate}{clean} & \raisebox{-0.5\height}{\includegraphics[width=\sthreefw]{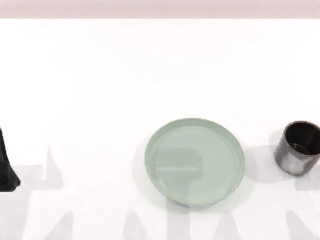}} & \raisebox{-0.5\height}{\includegraphics[width=\sthreefw]{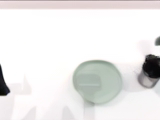}} & \raisebox{-0.5\height}{\includegraphics[width=\sthreefw]{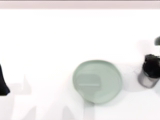}} & \raisebox{-0.5\height}{\includegraphics[width=\sthreefw]{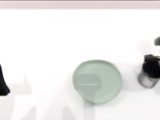}} & \raisebox{-0.5\height}{\includegraphics[width=\sthreefw]{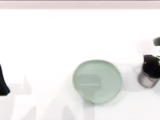}} & \raisebox{-0.5\height}{\includegraphics[width=\sthreefw]{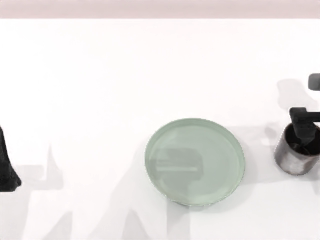}} \\ \noalign{\vskip 2pt}
\sthreelab{plate}{random} & \raisebox{-0.5\height}{\includegraphics[width=\sthreefw]{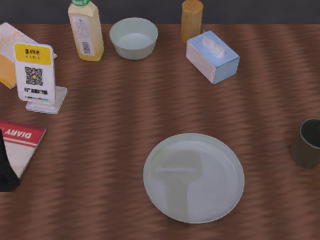}} & \raisebox{-0.5\height}{\includegraphics[width=\sthreefw]{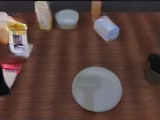}} & \raisebox{-0.5\height}{\includegraphics[width=\sthreefw]{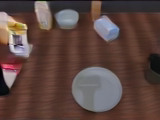}} & \raisebox{-0.5\height}{\includegraphics[width=\sthreefw]{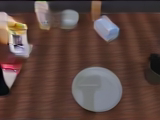}} & \raisebox{-0.5\height}{\includegraphics[width=\sthreefw]{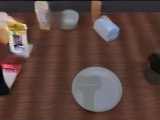}} & \raisebox{-0.5\height}{\includegraphics[width=\sthreefw]{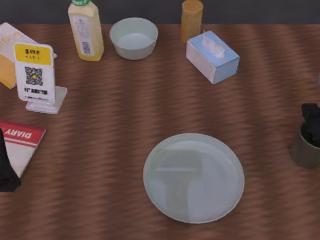}} \\
\end{tabular}
\caption{\textbf{Imagination quality versus video denoising steps.} All ten task--setting pairs of the denoising study (\cref{fig:steps}), without selection: the five tasks of \cref{app:policy-denoise} in the clean and randomized settings. From the current views, Efficient-WAM~\cite{li2026efficientwam}, the compact WAM underlying our policy, imagines the frame at the end of its first 16-command chunk with 1, 2, 4, or 10 video steps and 10 action steps. Only the head view of the three views is shown, at its native $160 \times 120$ pixels, without enhancement. Last column: the frame observed after executing the 2-step chunk. $^\dagger$The 4-step episode of this pair started from another seed.}
\label{fig:budgetall}
\end{figure*}

A pilot study of the denoising steps predates our policy. It measures task success and latency (\cref{fig:steps}) and the quality of the imagined frames (\cref{fig:budgetall}).

\subsubsection{Success and Latency}

The pilot uses the published Efficient-WAM checkpoint~\cite{li2026efficientwam}. It covers five RoboTwin tasks (adjust bottle, blocks ranking by size, handover block, hanging mug, and place container on plate) in the clean and randomized settings, with seed-matched episodes. We measured latency on an RTX 4090 at batch size 1 in bf16 as the median of 30 calls, excluding simulation and video decoding.

With 10 action steps, success is 80\% with 2 video steps (\cref{fig:steps}a). It is 68\%, 66\%, and 68\% with 1, 4, and 10 video steps. A chunk takes 74.8\,ms with 1 video step and 112.7\,ms with 2. It takes 188.6 and 289.3\,ms with 4 and 10 (\cref{fig:steps}b). With 2 video steps, 2 to 20 action steps reach 63--77\% success at 71.8 to 131.4\,ms (\cref{fig:steps}c, d).

One more video refresh costs 38.0\,ms per chunk, and one more action step costs 1.9\,ms, so the video branch dominates the latency. Two video steps are the best operating point we measured.

\subsubsection{Imagination Quality}

For one further episode per task, setting, and video-step count, seed-matched except one 4-step episode (\cref{fig:budgetall}), the study saved the decoded video of every chunk. These 40 episodes are not part of the success counts. We compare the imagined end-of-chunk frame with the decoded observation that starts the next chunk, through the same decoder. The metric is LPIPS with AlexNet features over all three views. The match is specific in time, as the imagined end frame scores 0.13 against the next observation and 0.37 against the one a chunk later.

On the first chunk, LPIPS is 0.133 with 1 video step and 0.129 with 2. It is 0.135 with 4 and 0.130 with 10, against 0.22 for repeating the current frame. One step is slightly worse than two (+0.005 before rounding). Fewer steps change the sample little. Against the 10-step frame, the frames imagined with 1, 2, and 4 steps score 0.035, 0.024, and 0.007 LPIPS.

Beyond the first chunk, closed-loop states diverge, and long failed episodes contribute most chunks. Over all chunks, the error rises to 0.24--0.28 for every step count, above the 0.21--0.26 of repeating the current frame. \Cref{fig:budgetall} shows all ten first-chunk pairs. These frames come from the policy's joint sampling, so they are nominal futures.

\section{Closed-Loop Intervention}
\label{app:closed}

Section~\mainref{sec:exp_closed}{5.3} of the main paper runs the policy with \guard in closed loop on four RoboTwin tasks.
This section gives the protocol, the 400-trial paired run, and where its remaining hazards come from.
It ends with paired runs of from-scratch heads on the same four tasks and on the four unseen tasks, and with Fast-WAM-Joint as the policy.
Finding~2 of Section~\mainref{sec:exp_closed}{5.3} of the main paper reports the from-scratch heads of \cref{app:closed-traced}, and its finding~3 the original checkpoint of \cref{app:closed-paired,app:closed-residual}.

\subsection{Protocol and Replays}
\label{app:closed-protocol}

\subsubsection{Trials and Recovery}

We run the closed loop on the four RoboTwin tasks in the clean and randomized settings, with the same selected checkpoint and threshold. For equal sample sizes, we use the first 20 completed trials per task and setting in ascending seed order, independent of their outcomes (160 in total). Every condition of Table~\mainref{tab:closed}{2} of the main paper uses the same seeds. These trials use expert-feasible starts and the official host configuration.

The perfect detector of Table~\mainref{tab:closed}{2} of the main paper rejects exactly the chunks whose replay shows a hazard. Its remaining hazards therefore start after an accepted chunk ends, for example when an object tips over after a placement.

The native command limits are 1,200 for block-size ranking, 900 for mug hanging, and 800 for bimanual handover and two-block stacking. An accepted candidate executes 16 native commands. SR uses the official task predicate, and a trial may retreat and later complete.

After a rejection, recovery holds the gripper and moves toward the start of the previously accepted chunk. Without such a reference, it executes a bounded inverse-candidate motion. Three consecutive rejected chunks trigger a protective stop. This happened in 3 of the 160 trials with \guard. In all 160, no retreat caused a hazardous event. In the paired run, 2 hazards begin during a retreat or a protective stop (\cref{tab:pairedcause}).

\subsubsection{Rejections and Replays}

Mug hanging, whose hook leaves little clearance, has the most hazards without \guard and the highest rejection rate with it. On the other tasks, \guard removes at least half of the hazards while rejecting fewer than 7\% of the chunks. A trial proposes 24 chunks on average, or 3,850 proposals in the 160 guarded trials.

Replay from a saved state is deterministic. Replaying all 208 rejected chunks shows that 47 of them (23\%) would have caused a hazard. The other 161 are false alarms, and each costs one bounded retreat of 1.4\,s on average.

Replaying 400 randomly chosen accepted chunks finds 2 hazardous ones (0.5\%), in line with the 13 hazardous guarded trials. Ambiguous closed-loop events are reviewed as in \cref{app:labels-audit}, and two of them stay unresolved and count as hazardous.

Sweeping $\tau_r$ in the 160 trials of Table~\mainref{tab:closed}{2} of the main paper, with both settings pooled, trades hazards for interventions. We lower $\tau_r$ from 0.60 to 0.40 and 0.30. The rejected chunks rise from 3.1\% to 5.4\% and 9.8\%. The hazardous trials fall from 12.5\% to 8.1\% and 5.6\%. The retreat time per trial grows from 1.0 to 1.8 and 3.2\,s.
In Table~\mainref{tab:closed}{2} of the main paper, both monitors match the 5.4\% rejection rate of \guard, and random rejection rejects each chunk with this probability.

\subsection{Paired Run}
\label{app:closed-paired}

\begin{table}[t]
\caption{\textbf{Paired closed-loop run.} Randomized setting, 100 trials per task on the four closed-loop tasks, learned monitors matched to \guard's rejection rate. Hazardous trials (\% of 400) by first event: contact (Cont.) or loss of support (Supp.). SR: success (\%). Rej.: realized rejection rate (\%).}
\label{tab:paired}
\centering
\small
\setlength{\tabcolsep}{2pt}
\resizebox{0.6\linewidth}{!}{%
\begin{tabular}{l ccc cc}
\toprule
& \multicolumn{3}{c}{Hazardous trials} & & \\
\cmidrule(lr){2-4}
Execution & All & Cont. & Supp. & SR & Rej. \\
\midrule
Policy only & 108 (27.0\%) & 52 & 56 & 78.5 & -- \\
+ FAIL-Detect-style & 79 (19.8\%) & 33 & 46 & 79.3 & 6.8 \\
+ ContactGuard-style & 62 (15.5\%) & 22 & 40 & 80.0 & 6.6 \\
+ \guard w/o future & 65 (16.3\%) & 22 & 43 & 79.8 & 6.7 \\
+ \guard, nominal future & 56 (14.0\%) & 19 & 37 & 80.3 & 6.8 \\
+ \guard & 39 (9.8\%) & 14 & 25 & 81.8 & 6.7 \\
\midrule
+ analytic layer$^\dagger$ & 68 (17.0\%) & 15 & 53 & 79.5 & 2.9 \\
+ analytic layer$^\dagger$, matched & 60 (15.0\%) & 11 & 49 & 79.3 & 6.8 \\
+ analytic layer$^\dagger$ or \guard & 36 (9.0\%) & 9 & 27 & 81.5 & 6.7 \\
\bottomrule
\end{tabular}}
\par\vspace{2pt}{\footnotesize\raggedright $^\dagger$Privileged geometry.\par}
\end{table}

\Cref{tab:paired} repeats the randomized closed loop with 100 paired trials per task. We set the threshold of each learned monitor so that its rejection rate on 20 separate calibration trials per task matches that of \guard. Rej.\ lists the realized rates (\%). All conditions of \cref{tab:paired} use the same 400 trial seeds.

The bottom rows of \cref{tab:paired} add an analytic safety layer with privileged geometry.
The analytic layer checks the commanded joint positions (2$^\circ$ margin) and velocities against their limits, a table-plane bound, and the forward-kinematics swept volume against the true scene meshes, as the collision check of \cref{app:detect-monitors} does (2\,mm penetration). Its matched version uses a 0.8\,cm clearance instead, set on the calibration trials. In the combination with \guard, a chunk is rejected when the layer at its default margins or \guard flags it, with $\tau_r$ raised from 0.40 to 0.50 to match the rejection rate of \guard alone.
At its default margins, it rejects only 2.9\% of the chunks.
Matched in rejection rate, it leaves fewer contact trials than \guard (11 vs.\ 14), and a motion-limit event occurs in only 1 trial (no monitor: 8; \guard: 4; each after an earlier hazardous event).
But losses of support barely fall (49; policy alone: 56), since kinematics does not show whether a grasp or a support holds (\cref{app:imagine-helps}).
Overall, the layer leaves 60 hazardous trials.

Rejecting a chunk when the layer or \guard flags it, at the same rejection rate, leaves the fewest hazardous trials (36).
Against the matched layer, the combination removes mainly losses of support (27 vs.\ 49; contact: 9 vs.\ 11).
Against \guard alone, it removes contacts (9 vs.\ 14; support: 27 vs.\ 25), and the totals are close (36 vs.\ 39).
On a real robot, this layer would need perceived geometry.

\subsection{Remaining Hazards}
\label{app:closed-residual}

This subsection breaks down the hazardous trials that remain in the paired run, first by type and cause and then chunk by chunk (\cref{tab:pairedcause}).

\begin{table}[t]
\caption{\textbf{Where the remaining hazards come from.} Paired run of \cref{tab:paired}, \guard vs.\ the ContactGuard-style monitor (CG) at the same rejection rate. (a)~Hazardous trials by cause (all, contact/loss of support). (b)~Chunk-level replay (\%).}
\label{tab:pairedcause}
\centering
\small
\setlength{\tabcolsep}{4pt}
\begin{tabular}{l cc}
\toprule
& \guard & CG \\
\midrule
\multicolumn{3}{l}{\textit{(a) Hazardous trials by cause (contact/support)}} \\
Missed detection & 22 (9/13) & 44 (18/26) \\
After an accepted chunk & 12 (3/9) & 14 (3/11) \\
During a retreat or stop & 2 (1/1) & 1 (0/1) \\
Unresolved & 3 (1/2) & 3 (1/2) \\
All & 39 (14/25) & 62 (22/40) \\
\midrule
\multicolumn{3}{l}{\textit{(b) Chunk-level replay}} \\
Proposed chunks & 9,936 & 9,968 \\
Hazardous chunks & 178 & 186 \\
Rejected (hazardous) & 666 (146) & 657 (130) \\
TPR & 82.0 & 69.9 \\
FPR & 5.3 & 5.4 \\
Precision & 21.9 & 19.8 \\
\bottomrule
\end{tabular}
\end{table}

\Cref{tab:pairedcause}(a) assigns each hazardous trial to the cause of its first hazardous event, using replays of the saved states. A missed detection is an executed, accepted chunk that causes the event, which recurs on replay but not when the robot holds still. An event after an accepted chunk also occurs when the robot holds still, so an earlier accepted chunk set it up beyond its 0.96\,s window. Other events begin during a retreat or a protective stop, and unresolved events stay ambiguous after review and count as hazardous. \Cref{tab:pairedcause}(b) replays every chunk proposed in each monitor's 400 trials from its saved state. A chunk is hazardous if its replay records an event that does not occur when the robot holds still, and ambiguous replays count as hazardous. TPR is the share of hazardous chunks that are rejected, FPR the share of safe chunks that are rejected, and precision the share of rejected chunks that are hazardous.

\subsubsection{Contact and Loss of Support}

We assign each hazardous trial to the type of its first hazardous event (\cref{tab:paired}). \guard lowers the contact trials of the policy alone from 52 to 14 and its loss-of-support trials from 56 to 25. The ContactGuard-style monitor lowers them to 22 and 40.

Loss of support remains the harder type, as on the validation split (\cref{app:percat}), but the future frames reduce it the most. It falls from 43 trials without them to 37 with the nominal future and 25 with the candidate-conditioned future, whereas contact falls only from 22 to 14.

Of the 39 hazardous trials with \guard (\cref{tab:pairedcause}(a)), 22 are missed detections. Another 12, 9 of them losses of support, were set up by an earlier accepted chunk. Finally, 2 begin during a retreat or protective stop, and 3 remain ambiguous.

The perfect detector of Table~\mainref{tab:closed}{2} of the main paper also leaves hazards that an earlier accepted chunk set up (3.8\% of its randomized trials). The ContactGuard-style monitor leaves about as many hazards of the three other kinds as \guard (18 vs.\ 17). The gap between the two monitors lies in missed detections (44 vs.\ 22).

\subsubsection{Chunk-Level Replay}

\Cref{tab:pairedcause}(b) gives the chunk-level operating point behind the trial counts. \guard rejects 146 of the 178 hazardous proposals (TPR 82.0\%) and 520 of the 9,758 safe ones (FPR 5.3\%). Of its rejections, 21.9\% are hazardous, close to the 23\% of the 160-trial replay (\cref{app:closed-protocol}).

Of the chunks that \guard accepts, 0.35\% are hazardous, in line with the 2 of the 400 accepted chunks sampled in that replay. Its recall lies between those on the policy proposals and on the validation split at a similar false-alarm rate (77.1\% and 84.4\% at 5\%). This recall counts every proposal. After a retreat, the policy often proposes a similar chunk again, so the 178 hazardous proposals fall in only 105 trials.

At the same rejection rate, the ContactGuard-style monitor rejects 69.9\% of the hazardous proposals (73.1\% on validation at 5\% FPR). It trails \guard by 12.1 points, and the precisions differ by only 2.1 points. Hazardous chunks are rare, so most rejections of either monitor are false alarms.

\subsection{From-Scratch Heads and Unseen Tasks}
\label{app:closed-traced}

\begin{table}[t]
\caption{\textbf{From-scratch heads in closed loop.} Two paired runs in the randomized setting, columns as in \cref{tab:paired}. (a)~The four closed-loop tasks. (b)~The four unseen tasks.}
\label{tab:pairedmore}
\centering
\small
\setlength{\tabcolsep}{2pt}
\resizebox{0.6\linewidth}{!}{%
\begin{tabular}{l ccc cc}
\toprule
& \multicolumn{3}{c}{Hazardous trials} & & \\
\cmidrule(lr){2-4}
Execution & All & Cont. & Supp. & SR & Rej. \\
\midrule
\multicolumn{6}{l}{\textit{(a) Closed-loop tasks}} \\
Policy only & 108 (27.0\%) & 52 & 56 & 78.5 & -- \\
+ ContactGuard-style & 61 (15.3\%) & 21 & 40 & 79.8 & 7.1 \\
+ \guard w/o future & 67 (16.8\%) & 23 & 44 & 79.5 & 7.3 \\
+ \guard, nominal future & 59 (14.8\%) & 20 & 39 & 80.0 & 7.2 \\
+ \guard & 43 (10.8\%) & 15 & 28 & 81.3 & 7.2 \\
\midrule
\multicolumn{6}{l}{\textit{(b) Unseen tasks}} \\
Policy only & 119 (29.8\%) & 48 & 71 & 71.5 & -- \\
+ ContactGuard-style & 84 (21.0\%) & 29 & 55 & 72.3 & 8.0 \\
+ \guard w/o future & 87 (21.8\%) & 27 & 60 & 72.0 & 8.2 \\
+ \guard & 69 (17.3\%) & 25 & 44 & 73.0 & 8.1 \\
\bottomrule
\end{tabular}}
\end{table}

In \cref{tab:pairedmore}, \guard ($\tau_r = 0.35$) and the ContactGuard-style monitor are the selected seeds of the from-scratch heads trained on the fully recorded windows (\cref{app:detect-blind}), and the \guard variants are trained in the same way. The monitor and the variants are matched to \guard's rejection rate. Panel~(a) uses the 400 trial seeds and the calibration trials of \cref{tab:paired}, and its policy-only counts repeat that table. Panel~(b) is a separate run on the four unseen tasks of \cref{tab:general}, with 100 new trial seeds and 20 calibration trials per task.

\subsubsection{Closed-Loop Tasks}

\Cref{tab:pairedmore}(a) repeats the paired run of \cref{tab:paired} on its trial seeds with the from-scratch heads described above.
The from-scratch \guard rejects 7.2\% of the chunks at its selected $\tau_r = 0.35$ and leaves 43 hazardous trials.
Matched to this rate, the ContactGuard-style monitor leaves 61, \guard without future frames 67, and \guard with the nominal future 59.

At the chunk level (as in \cref{tab:pairedcause}(b)), the from-scratch \guard rejects 79.3\% of the hazardous proposals at 5.8\% FPR, with 20.5\% precision.
The other two from-scratch seeds, at their own thresholds (0.30 and 0.45), leave 46 and 47 hazardous trials (three seeds: 11.3$\pm$0.5\%).

\begin{figure*}[t]
\centering
\includegraphics[width=\textwidth]{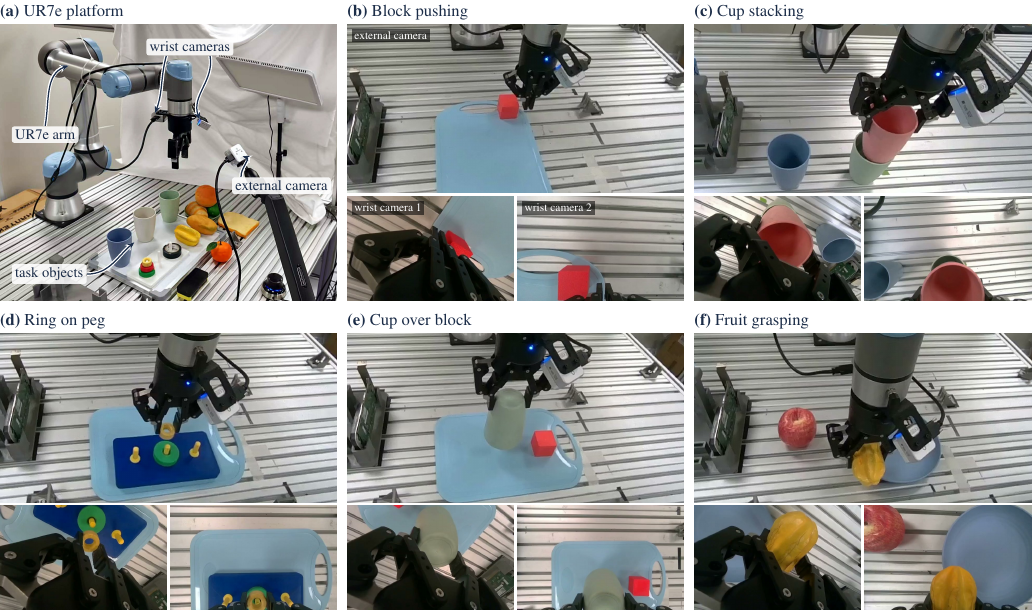}
\caption{\textbf{The UR7e platform and its five tasks.} (a)~A UR7e arm with a parallel gripper, two gripper-mounted wrist cameras, and an external camera. (b)--(f)~One time step of one teleoperated training demonstration per task, from the three cameras the policy reads: the external camera (top, cropped around the gripper) and the two wrist cameras (bottom).}
\label{fig:realtasks}
\end{figure*}

\subsubsection{Unseen Tasks}

\Cref{tab:pairedmore}(b) uses the from-scratch heads of panel~(a), whose training windows exclude these tasks (\cref{tab:general}).
\guard lowers the hazardous trials from 29.8\% to 17.3\%.
The ContactGuard-style monitor leaves 21.0\%, and \guard without future frames 21.8\%.

The lead over the monitor is smaller than on the closed-loop tasks and lies mostly in loss of support (44 vs.\ 55 trials; contact: 25 vs.\ 29).
At the chunk level, \guard rejects 61.5\% of the hazardous proposals at 6.9\% FPR and the monitor 50.2\% at 7.1\%, with precisions of 16.2\% and 13.0\%.
Success changes little (73.0\% vs.\ 71.5\% for the policy alone).

\subsection{Another Host WAM}
\label{app:closed-otherwam}

\begin{table}[t]
\caption{\textbf{\guard on Fast-WAM-Joint.} Blind-set BAcc (\%) and the paired closed loop with Fast-WAM-Joint as the policy (randomized, 400 trials per row), monitors matched to \guard's rejection rate. ContactGuard-style blind BAcc: recipe of \cref{app:detect-monitors} without the 24-configuration search.}
\label{tab:otherwam}
\centering
\small
\setlength{\tabcolsep}{4pt}
\begin{tabular}{l cccc}
\toprule
Execution & Blind BAcc & Hazard$\downarrow$ & SR$\uparrow$ & Rej. \\
\midrule
Fast-WAM-Joint only & -- & 25.8 & 80.3 & -- \\
+ ContactGuard-style & 86.8 & 16.0 & 80.5 & 6.9 \\
+ \guard & 90.4 & 11.3 & 80.8 & 6.9 \\
\bottomrule
\end{tabular}
\end{table}

We attach \guard to Fast-WAM~\cite{yuan2026fastwam}, a 6B WAM that learns from future video but skips video generation at test time.
Its attention mask keeps the future video tokens from attending to the action tokens, so its stock video branch would imagine the same future for every candidate. We therefore use its joint variant, Fast-WAM-Joint~\cite{yuan2026fastwam}, in which the two attend to each other. As for our policy, we fix the action tokens to the candidate at noise level zero and denoise only the future latents, which adds 45\,ms per candidate, and we retrain \guard on the same training windows with the frames it imagines.
On the blind set, this \guard reaches 90.4\% BAcc (three-seed mean of heads trained from scratch, as in \cref{app:detect-blind}), below the 92.1\% it reaches with our video expert (\cref{tab:otherwam}).
With Fast-WAM-Joint as the policy in the paired closed loop of \cref{app:closed-paired}, it lowers hazardous trials from 25.8\% to 11.3\%, against 16.0\% for the ContactGuard-style monitor at the same rejection rate.
Success stays at 80.8\%, against 80.3\% without a monitor (\cref{tab:otherwam}).

\section{Real-Robot Experiments}
\label{app:real}

Section~\mainref{sec:exp_real}{5.6} of the main paper evaluates the policy and \guard on a UR7e arm.
This section describes the hardware and tasks, the success protocol, and the adaptation of \guard with the staged hazard scenarios.

\begin{table*}[t]
\caption{\textbf{Real-robot hazard scenarios.} Hazards per condition in 20 layouts per scenario, false alarms (FA) in 20 control layouts without the hazard, and completed hazard layouts.}
\label{tab:realsafety}
\centering
\footnotesize
\setlength{\tabcolsep}{2.5pt}
\begin{adjustbox}{max width=\linewidth}\begin{tabular}{l c cc cc cc cc cc cc cc}
\toprule
& Direct & \multicolumn{2}{c}{Depth check} & \multicolumn{2}{c}{CG-style} & \multicolumn{2}{c}{No future} & \multicolumn{2}{c}{Real only} & \multicolumn{2}{c}{Sim.\ only} & \multicolumn{2}{c}{\guard} & \multicolumn{2}{c}{Completed} \\
\cmidrule(lr){3-4} \cmidrule(lr){5-6} \cmidrule(lr){7-8} \cmidrule(lr){9-10} \cmidrule(lr){11-12} \cmidrule(lr){13-14} \cmidrule(lr){15-16}
Scenario & Hazards & Hazards & FA & Hazards & FA & Hazards & FA & Hazards & FA & Hazards & FA & Hazards & FA & Direct & \guard \\
\midrule
Bottle in the path & 14/20 & 6/20 & 3/20 & 4/20 & 2/20 & 5/20 & 2/20 & 7/20 & 4/20 & 9/20 & 2/20 & 2/20 & 1/20 & 6/20 & 15/20 \\
Cup near the table edge & 15/20 & 8/20 & 4/20 & 5/20 & 2/20 & 5/20 & 1/20 & 9/20 & 5/20 & 10/20 & 2/20 & 3/20 & 1/20 & 5/20 & 14/20 \\
Lowered approach & 14/20 & 5/20 & 2/20 & 2/20 & 1/20 & 3/20 & 1/20 & 7/20 & 3/20 & 6/20 & 3/20 & 1/20 & 1/20 & 7/20 & 15/20 \\
\midrule
Total & 43/60 & 19/60 & 9/60 & 11/60 & 5/60 & 13/60 & 4/60 & 23/60 & 12/60 & 25/60 & 7/60 & 6/60 & 3/60 & 18/60 & 44/60 \\
\bottomrule
\end{tabular}\end{adjustbox}
\end{table*}

\subsection{Hardware and Tasks}
\label{app:real-hw}

A UR7e arm with a parallel gripper performs five tasks that cover pushing, stacking, assembling, covering, and grasping (\cref{fig:realtasks}).
\emph{Block pushing to target} pushes a red block across a board to its handle cutout.
\emph{Cup stacking} picks one of three cups and nests it into another.

\emph{Ring-on-peg placement} puts two rings onto the same peg of a peg board.
The robot must align each ring with the narrow peg before releasing it.
\emph{Cup-over-block placement} covers a red block with an upside-down cup.
\emph{Fruit grasping} picks a starfruit and places it on a plate next to a distractor apple.

One external RGB-D camera (1280$\times$720) and two wrist cameras (640$\times$480) record at 15\,fps.
The policy and \guard read these three views in training and at inference, in place of the head and wrist views.
\Cref{fig:realtasks} shows all three views at one time step, and \cref{fig:imagined} shows the external view only.
The robot executes commands at 10\,Hz, and the 15\,Hz recordings are the time base of \cref{fig:imagined}.
The policy runs on a workstation GPU and sends 7-dimensional joint-velocity and gripper commands.

\subsection{Task Success}
\label{app:real-success}

All methods are trained on the same 50 teleoperated demonstrations per task.
\methodname is fine-tuned from its RoboTwin checkpoint, $\pi_{0.5}$ from its official base checkpoint, and DiT4DiT and Diffusion Policy are trained with their released code. Fast-WAM is fine-tuned from its released checkpoint.

Each method is trained with four seeds and evaluated in 40 labeled trials per task, ten per seed, from initial layouts drawn from one set shared by all methods. \methodname runs these trials twice, with \guard on (its row in Table~\mainref{tab:real}{6} of the main paper) and with \guard off (the w/o-\guard row).
The object layout changes from trial to trial, so a policy cannot succeed by replaying one trajectory.
Executions that cannot be labeled, for example because a hardware fault interrupts them, are discarded and rerun.

With \guard off, \methodname succeeds in 151 of the 200 trials and $\pi_{0.5}$ in 119.
Fast-WAM succeeds in 97, DiT4DiT in 58, and Diffusion Policy in 27 (Table~\mainref{tab:real}{6} of the main paper).
On block pushing, the two methods differ by one trial (31 vs.\ 30).
With \guard on, \methodname succeeds in 148 of the 200 trials, with 0.4 retreats per trial and no protective stop. This also exceeds the 119 successes of $\pi_{0.5}$. On block pushing, $\pi_{0.5}$ succeeds in one trial more (30 vs.\ 29).

\subsection{Staged Hazard Scenarios}
\label{app:real-staged}

We adapt \guard to the UR7e and test it in the three staged hazard scenarios of \cref{tab:realsafety}.

Each scenario of \cref{tab:realsafety} runs its 20 layouts unmonitored, with a depth-based collision check, with \guard without future frames, and with \guard, paired by layout, plus 20 control layouts without the hazard. The CG-style, real-only, and sim-only conditions are described in \cref{app:real-learned}. A false alarm is a control layout with at least one rejection, and a completed layout is a hazard layout in which the task is completed.

\begin{figure}[t]
\centering
\includegraphics[width=0.6\linewidth]{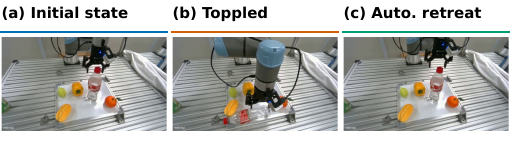}
\caption{\textbf{Bottle scenario on the UR7e.} Representative states: (a) the initial configuration, (b) a bottle-toppling contact under direct execution, and (c) \guard-triggered automatic retreat with the bottle upright.}
\label{fig:realrobot}
\end{figure}

\subsubsection{Adaptation and Scenarios}

Of the 258 human-verified windows of \cref{app:labels-other}, eight trajectories give 196 for training and the other two give 62 for validation.
The adaptation trajectories use a mug and a box in layouts different from the test scenarios, which use a bottle, a cup, and an approach toward the bare table.

Robot-specific encoders embed the measured 7-dimensional state and sixteen 7-dimensional velocity/gripper commands.
Training updates 0.05\,M encoder and head parameters, while the shared visual and fusion layers stay fixed.
For velocity commands, the displacement input of $g_{\Delta}$ is the joint displacement obtained by integrating the commanded velocities over the chunk.
\guard reads the same three views as the policy.

In the bottle and cup scenarios of \cref{tab:realsafety}, the policy performs its task with the obstacle in its path.
\Cref{fig:realrobot} shows the bottle scenario.
In each layout, the four conditions above run in random order, and the three conditions of \cref{app:real-learned} follow in the same layouts.
A hazard is a toppled bottle, a cup that falls off the table, or a gripper--table contact above 15\,N on the wrist force--torque sensor.
Two observers who were blind to the condition judged the hazards from video and the wrist force--torque log (agreement 98\%).

The lowered approach is instead induced like the table collisions of the pipeline (\cref{tab:hazardsfull}).
In one approach chunk per trial, the end-effector targets are lowered by 3\,cm and converted back to joint velocities by inverse kinematics before the monitor scores the candidate.
Executed directly, this chunk drives the fingertips into the table, above 15\,N in 14 of the 20 layouts (\cref{tab:realsafety}).
Its control layouts run the unmodified approach.

The depth-based check sweeps the gripper volume along the forward-kinematics path of the candidate through the point cloud of the external camera.
Its safety margin (1.5\,cm) was tuned on the adaptation trajectories.
With \guard, the hazard trials need 1.9 retreats and 2.6\,s of extra execution time on average, and no protective stop occurs.

\subsubsection{The Imagined Frame on the UR7e}

On the UR7e, \guard reads the last frame that the video expert imagines for each view.
This frame lies 1.07\,s (16 frames at 15\,Hz) after the decision time, the last offset shown in \cref{fig:imagined}.
It falls beyond the 8 commands that an accepted chunk executes and within the 16-command candidate.
At this offset the nominal imagined frames of \cref{fig:imagined} are heavily blurred and keep at most the coarse layout and arm pose, not the contact itself.

\subsubsection{Learned Baseline and Training Sources}
\label{app:real-learned}

We run three further conditions in the same 60 hazard and 60 control layouts, after the runs above and with the same blinded observers.
The ContactGuard-style monitor of \cref{app:detect-monitors} is adapted with the same 258 windows: starting from its RoboTwin weights, its latent world model is fine-tuned on the task demonstrations and the adaptation trajectories, and its probe on the 196 training windows.
The real-only head is \guard trained from random initialization on the 196 training windows, with no simulated data.
For both, the 62 validation windows set the threshold, as for the adapted head.
The simulation-only head is the RoboTwin head without adaptation. A fixed mapping feeds it the UR7e state and the integrated velocity commands as those of one RoboTwin arm, and $\tau_r$ stays at 0.40.

The ContactGuard-style monitor leaves 11 hazards, between \guard (6) and \guard without future frames (13).
The heads trained on one data source leave many more hazards, 23 with the real windows only and 25 with simulation only, and complete the task less often.
The real-only head fits its training windows (98.9\% BAcc) but reaches only 70.6\% on the validation windows, against 88.2\% for the adapted head.
It also raises 12 false alarms in the controls, against 3.
Neither data source alone matches the adapted head, which uses both.

\section{Limitations}
\label{app:limits}

Below are the limits of our evaluation and of \guard.

\subsection{Evaluation Scope}
\label{app:limits-scope}

Hazard labels are rule-defined and were not human-audited in closed loop.

Closed-loop hazard counts cover eight of the 50 RoboTwin tasks.
On the four unseen tasks among them, the lead of \guard over the ContactGuard-style monitor is smaller than on the other four (\cref{tab:pairedmore}(b)).
The UR7e head needs 258 human-verified windows, and neither these windows alone nor simulation alone trains a comparable head (\cref{app:real-learned}).

\subsection{Detection and the Shared Video Expert}
\label{app:limits-method}

Transfer to unseen tasks is partial (\cref{tab:general}), and its scaling with the pipeline's coverage is unmeasured.
Most rejections are false alarms.
\guard gives no safety guarantee, and its gain is not fully separated from the capacity of the video expert that it shares with the policy.
A hazard that this expert cannot imagine may be both proposed and accepted.
On real data, the imagined frames blur beyond about 0.3\,s (\cref{fig:imagined}).

%% file: ref.bib
@inproceedings{agia2024sentinel,
  title = {Unpacking Failure Modes of Generative Policies: Runtime Monitoring of Consistency and Progress},
  author = {Agia, Christopher and Sinha, Rohan and Yang, Jingyun and Cao, Ziang and Antonova, Rika and Pavone, Marco and Bohg, Jeannette},
  booktitle = {Conference on Robot Learning (CoRL)},
  year = {2024},
  url = {https://arxiv.org/abs/2410.04640}
}

@article{akbari2026flashwam,
  title = {{Flash-WAM}: Modality-Aware Distillation for World Action Models},
  author = {Akbari, Arman and Zhang, Ci and Akbari, Arash and Zhao, Lin and Chen, Yixiao and Chen, Weiwei and Zhang, Xuan and Yuan, Geng and Wang, Yanzhi},
  journal = {arXiv preprint arXiv:2606.05254},
  year = {2026},
  doi = {10.48550/arXiv.2606.05254},
  url = {https://arxiv.org/abs/2606.05254}
}

@article{bai2025qwen3vl,
  author  = {Bai, Shuai and Cai, Yuxuan and Chen, Ruizhe and Chen, Keqin and Chen, Xionghui and Cheng, Zesen and Deng, Lianghao and Ding, Wei and Gao, Chang and Ge, Chunjiang and Ge, Wenbin and Guo, Zhifang and Huang, Qidong and Huang, Jie and Huang, Fei and Hui, Binyuan and Jiang, Shutong and Li, Zhaohai and Li, Mingsheng and Li, Mei and Li, Kaixin and Lin, Zicheng and Lin, Junyang and Liu, Xuejing and Liu, Jiawei and Liu, Chenglong and Liu, Yang and Liu, Dayiheng and Liu, Shixuan and Lu, Dunjie and Luo, Ruilin and Lv, Chenxu and Men, Rui and Meng, Lingchen and Ren, Xuancheng and Ren, Xingzhang and Song, Sibo and Sun, Yuchong and Tang, Jun and Tu, Jianhong and Wan, Jianqiang and Wang, Peng and Wang, Pengfei and Wang, Qiuyue and Wang, Yuxuan and Xie, Tianbao and Xu, Yiheng and Xu, Haiyang and Xu, Jin and Yang, Zhibo and Yang, Mingkun and Yang, Jianxin and Yang, An and Yu, Bowen and Zhang, Fei and Zhang, Hang and Zhang, Xi and Zheng, Bo and Zhong, Humen and Zhou, Jingren and Zhou, Fan and Zhou, Jing and Zhu, Yuanzhi and Zhu, Ke},
  title   = {{Qwen3-VL} Technical Report},
  journal = {arXiv preprint arXiv:2511.21631},
  year    = {2025},
  url     = {https://arxiv.org/abs/2511.21631}
}

@inproceedings{bi2025motus,
  title = {{Motus}: A Unified Latent Action World Model},
  author = {Bi, Hongzhe and Tan, Hengkai and Xie, Shenghao and Wang, Zeyuan and Huang, Shuhe and Liu, Haitian and Zhao, Ruowen and Feng, Yao and Xiang, Chendong and Rong, Yinze and Zhao, Hongyan and Liu, Hanyu and Su, Zhizhong and Ma, Lei and Su, Hang and Zhu, Jun},
  booktitle = {Proceedings of the IEEE/CVF Conference on Computer Vision and Pattern Recognition (CVPR)},
  pages = {35101--35113},
  year = {2026},
  url = {https://arxiv.org/abs/2512.13030}
}

@inproceedings{black2024pi0,
  title = {{$\pi_0$}: A Vision-Language-Action Flow Model for General Robot Control},
  author = {Black, Kevin and Brown, Noah and Driess, Danny and Esmail, Adnan and Equi, Michael and Finn, Chelsea and Fusai, Niccolo and Groom, Lachy and Hausman, Karol and Ichter, Brian and Jakubczak, Szymon and Jones, Tim and Ke, Liyiming and Levine, Sergey and Li-Bell, Adrian and Mothukuri, Mohith and Nair, Suraj and Pertsch, Karl and Shi, Lucy Xiaoyang and Smith, Laura and Tanner, James and Vuong, Quan and Walling, Anna and Wang, Haohuan and Zhilinsky, Ury},
  booktitle = {Robotics: Science and Systems (RSS)},
  year = {2025},
  doi = {10.15607/RSS.2025.XXI.010},
  url = {https://arxiv.org/abs/2410.24164}
}

@article{cai2026ahawam,
  title = {{AHA-WAM}: Asynchronous Horizon-Adaptive World-Action Modeling with Observation-Guided Context Routing},
  author = {Cai, Jisong and Ling, Long and Chu, Shiwei and Liu, Zhongshan and Kang, Jiayue and Liang, Zhixuan and Xu, Wenjie and Mao, Yinan and Zhang, Weinan and Yang, Xiaokang and Ying, Ru and Zheng, Ran and Mu, Yao},
  journal = {arXiv preprint arXiv:2606.09811},
  year = {2026},
  doi = {10.48550/arXiv.2606.09811},
  url = {https://arxiv.org/abs/2606.09811}
}

@article{cen2025worldvla,
  title = {{WorldVLA}: Towards Autoregressive Action World Model},
  author = {Cen, Jun and Yu, Chaohui and Yuan, Hangjie and Jiang, Yuming and Huang, Siteng and Guo, Jiayan and Li, Xin and Song, Yibing and Luo, Hao and Wang, Fan and Zhao, Deli and Chen, Hao},
  journal = {arXiv preprint arXiv:2506.21539},
  year = {2025},
  url = {https://arxiv.org/abs/2506.21539}
}

@article{cheang2024gr2,
  title = {{GR-2}: A Generative Video-Language-Action Model with Web-Scale Knowledge for Robot Manipulation},
  author = {Cheang, Chi-Lam and Chen, Guangzeng and Jing, Ya and Kong, Tao and Li, Hang and Li, Yifeng and Liu, Yuxiao and Wu, Hongtao and Xu, Jiafeng and Yang, Yichu and Zhang, Hanbo and Zhu, Minzhao},
  journal = {arXiv preprint arXiv:2410.06158},
  year = {2024},
  url = {https://arxiv.org/abs/2410.06158}
}

@inproceedings{chen2025robotwin2,
  title = {{RoboTwin} 2.0: A Scalable Data Generator and Benchmark with Strong Domain Randomization for Robust Bimanual Robotic Manipulation},
  author = {Chen, Tianxing and Chen, Zanxin and Chen, Baijun and Cai, Zijian and Liu, Yibin and Li, Zixuan and Liang, Qiwei and Lin, Xianliang and Ge, Yiheng and Gu, Zhenyu and Deng, Weiliang and Guo, Yubin and Nian, Tian and Xie, Xuanbing and Chen, Qiangyu and Su, Kailun and Xu, Tianling and Liu, Guodong and Hu, Mengkang and Gao, Huan-ang and Wang, Kaixuan and Liang, Zhixuan and Qin, Yusen and Yang, Xiaokang and Luo, Ping and Mu, Yao},
  booktitle = {International Conference on Machine Learning (ICML)},
  year = {2026},
  url = {https://arxiv.org/abs/2506.18088}
}

@inproceedings{chi2023diffusion,
  title = {{Diffusion Policy}: Visuomotor Policy Learning via Action Diffusion},
  author = {Chi, Cheng and Feng, Siyuan and Du, Yilun and Xu, Zhenjia and Cousineau, Eric and Burchfiel, Benjamin and Song, Shuran},
  booktitle = {Robotics: Science and Systems (RSS)},
  year = {2023},
  url = {https://arxiv.org/abs/2303.04137}
}

@inproceedings{collins2026tapestry,
  title = {{TAPESTRY}: A Single Backbone for Video Generation and Robot Control},
  author = {Collins, Jeremy A. and Lee, Seungjae and Wachakorn, Rachanon and Srinivasan, Krishnan and Garg, Animesh and Guizilini, Vitor Campagnolo and Shah, Paarth},
  booktitle = {Conference on Robot Learning (CoRL)},
  year = {2026},
  url = {https://arxiv.org/abs/2610.36823}
}

@inproceedings{dai2025racer,
  author    = {Dai, Yinpei and Lee, Jayjun and Fazeli, Nima and Chai, Joyce},
  title     = {{RACER}: Rich Language-Guided Failure Recovery Policies for Imitation Learning},
  booktitle = {IEEE International Conference on Robotics and Automation (ICRA)},
  pages     = {15657--15664},
  year      = {2025},
  doi       = {10.1109/ICRA55743.2025.11127799},
  url       = {https://arxiv.org/abs/2409.14674}
}

@inproceedings{du2023unipi,
  title = {Learning Universal Policies via Text-Guided Video Generation},
  author = {Du, Yilun and Yang, Sherry and Dai, Bo and Dai, Hanjun and Nachum, Ofir and Tenenbaum, Josh and Schuurmans, Dale and Abbeel, Pieter},
  booktitle = {Advances in Neural Information Processing Systems (NeurIPS)},
  volume = {36},
  pages = {9156--9172},
  year = {2023},
  doi = {10.52202/075280-0403},
  url = {https://arxiv.org/abs/2302.00111}
}

@inproceedings{duan2024aha,
  title = {{AHA}: A Vision-Language-Model for Detecting and Reasoning Over Failures in Robotic Manipulation},
  author = {Duan, Jiafei and Pumacay, Wilbert and Kumar, Nishanth and Wang, Yi Ru and Tian, Shulin and Yuan, Wentao and Krishna, Ranjay and Fox, Dieter and Mandlekar, Ajay and Guo, Yijie},
  booktitle = {International Conference on Learning Representations (ICLR)},
  year = {2025},
  url = {https://arxiv.org/abs/2410.00371}
}

@article{ebert2018visualforesight,
  title = {Visual Foresight: Model-Based Deep Reinforcement Learning for Vision-Based Robotic Control},
  author = {Ebert, Frederik and Finn, Chelsea and Dasari, Sudeep and Xie, Annie and Lee, Alex and Levine, Sergey},
  journal = {arXiv preprint arXiv:1812.00568},
  year = {2018},
  url = {https://arxiv.org/abs/1812.00568}
}

@inproceedings{finn2017foresight,
  title = {Deep Visual Foresight for Planning Robot Motion},
  author = {Finn, Chelsea and Levine, Sergey},
  booktitle = {IEEE International Conference on Robotics and Automation (ICRA)},
  pages = {2786--2793},
  year = {2017},
  doi = {10.1109/ICRA.2017.7989324},
  url = {https://arxiv.org/abs/1610.00696}
}

@inproceedings{gu2025safe,
  title = {{SAFE}: Multitask Failure Detection for Vision-Language-Action Models},
  author = {Gu, Qiao and Ju, Yuanliang and Sun, Shengxiang and Gilitschenski, Igor and Nishimura, Haruki and Itkina, Masha and Shkurti, Florian},
  booktitle = {Advances in Neural Information Processing Systems (NeurIPS)},
  year = {2025},
  doi = {10.52202/085713-1337},
  url = {https://arxiv.org/abs/2506.09937}
}

@inproceedings{guo2024pad,
  title = {Prediction with Action: Visual Policy Learning via Joint Denoising Process},
  author = {Guo, Yanjiang and Hu, Yucheng and Zhang, Jianke and Wang, Yen-Jen and Chen, Xiaoyu and Lu, Chaochao and Chen, Jianyu},
  booktitle = {Advances in Neural Information Processing Systems (NeurIPS)},
  year = {2024},
  url = {https://arxiv.org/abs/2411.18179}
}

@article{hinton2015distilling,
  title = {Distilling the Knowledge in a Neural Network},
  author = {Hinton, Geoffrey and Vinyals, Oriol and Dean, Jeff},
  journal = {arXiv preprint arXiv:1503.02531},
  year = {2015},
  url = {https://arxiv.org/abs/1503.02531}
}

@inproceedings{hu2022lora,
  title = {{LoRA}: Low-Rank Adaptation of Large Language Models},
  author = {Hu, Edward J. and Shen, Yelong and Wallis, Phillip and Allen-Zhu, Zeyuan and Li, Yuanzhi and Wang, Shean and Wang, Lu and Chen, Weizhu},
  booktitle = {International Conference on Learning Representations (ICLR)},
  year = {2022},
  url = {https://arxiv.org/abs/2106.09685}
}

@inproceedings{hu2025vpp,
  title = {Video Prediction Policy: A Generalist Robot Policy with Predictive Visual Representations},
  author = {Hu, Yucheng and Guo, Yanjiang and Wang, Pengchao and Chen, Xiaoyu and Wang, Yen-Jen and Zhang, Jianke and Sreenath, Koushil and Lu, Chaochao and Chen, Jianyu},
  booktitle = {International Conference on Machine Learning (ICML)},
  year = {2025},
  url = {https://arxiv.org/abs/2412.14803}
}

@inproceedings{intelligence2025pi05,
  title = {{$\pi_{0.5}$}: a Vision-Language-Action Model with Open-World Generalization},
  author = {Black, Kevin and Brown, Noah and Darpinian, James and Dhabalia, Karan and Driess, Danny and Esmail, Adnan and Equi, Michael and Finn, Chelsea and Fusai, Niccolo and Galliker, Manuel Y. and Ghosh, Dibya and Groom, Lachy and Hausman, Karol and Ichter, Brian and Jakubczak, Szymon and Jones, Tim and Ke, Liyiming and LeBlanc, Devin and Levine, Sergey and Li-Bell, Adrian and Mothukuri, Mohith and Nair, Suraj and Pertsch, Karl and Ren, Allen Z. and Shi, Lucy Xiaoyang and Smith, Laura and Springenberg, Jost Tobias and Stachowicz, Kyle and Tanner, James and Vuong, Quan and Walke, Homer and Walling, Anna and Wang, Haohuan and Yu, Lili and Zhilinsky, Ury},
  booktitle = {Conference on Robot Learning (CoRL)},
  year = {2025},
  url = {https://arxiv.org/abs/2504.16054}
}

@inproceedings{kim2025openvlaoft,
  title = {Fine-Tuning Vision-Language-Action Models: Optimizing Speed and Success},
  author = {Kim, Moo Jin and Finn, Chelsea and Liang, Percy},
  booktitle = {Robotics: Science and Systems (RSS)},
  year = {2025},
  doi = {10.15607/RSS.2025.XXI.017},
  url = {https://arxiv.org/abs/2502.19645}
}

@inproceedings{kim2026cosmospolicy,
  title = {{Cosmos Policy}: Fine-Tuning Video Models for Visuomotor Control and Planning},
  author = {Kim, Moo Jin and Gao, Yihuai and Lin, Tsung-Yi and Lin, Yen-Chen and Ge, Yunhao and Lam, Grace and Liang, Percy and Song, Shuran and Liu, Ming-Yu and Finn, Chelsea and Gu, Jinwei},
  booktitle = {International Conference on Learning Representations (ICLR)},
  year = {2026},
  url = {https://arxiv.org/abs/2601.16163}
}

@inproceedings{ko2024avdc,
  title = {Learning to Act from Actionless Videos through Dense Correspondences},
  author = {Ko, Po-Chen and Mao, Jiayuan and Du, Yilun and Sun, Shao-Hua and Tenenbaum, Joshua B.},
  booktitle = {International Conference on Learning Representations (ICLR)},
  year = {2024},
  url = {https://arxiv.org/abs/2310.08576}
}

@inproceedings{li2025uva,
  title = {Unified Video Action Model},
  author = {Li, Shuang and Gao, Yihuai and Sadigh, Dorsa and Song, Shuran},
  booktitle = {Robotics: Science and Systems (RSS)},
  year = {2025},
  url = {https://arxiv.org/abs/2503.00200}
}

@article{li2026efficientwam,
  title = {{Efficient-WAM}: A {1B}-Parameter World-Action Model with Low-Cost Future Imagination},
  author = {Li, Jiajun and Guo, Tiecheng and Ye, Yifan and Zhang, Rongyu and Chi, Xiaowei and Sun, Qianpu and Li, Ying and Lou, Yunfan and Wang, Jinqiao and Lu, Zhihe and Guo, Meng and Zhang, Shanghang},
  journal = {arXiv preprint arXiv:2606.10040},
  year = {2026},
  doi = {10.48550/arXiv.2606.10040},
  url = {https://arxiv.org/abs/2606.10040}
}

@inproceedings{li2026lightwam,
  title = {{Light-WAM}: Efficient World Action Models with State-Fusion Action Decoding},
  author = {Li, Ziang and Cheng, Dongzhou and Wang, Yibin and Wang, Shiyue and Xu, Xiaoyang and Weng, Lingxuan and Wang, Juan and Wang, Jiaqi},
  booktitle = {Conference on Robot Learning (CoRL)},
  year = {2026},
  url = {https://arxiv.org/abs/2606.08242}
}

@inproceedings{li2026lingbotva,
  title = {Causal World Modeling for Robot Control},
  author = {Li, Lin and Zhang, Qihang and Luo, Yiming and Yang, Shuai and Wang, Ruilin and Zhang, Luyao and Yu, Mingrui and Gao, Zelin and Xue, Nan and Zhou, Boyu and Zhu, Xing and Ding, Mingyu and Shen, Yujun and Xu, Yinghao},
  booktitle = {Robotics: Science and Systems (RSS)},
  year = {2026},
  doi = {10.15607/RSS.2026.XXII.016},
  url = {https://arxiv.org/abs/2601.21998}
}

@article{liang2025videopolicy,
  title = {Video Generators are Robot Policies},
  author = {Liang, Junbang and Tokmakov, Pavel and Liu, Ruoshi and Sudhakar, Sruthi and Shah, Paarth and Ambrus, Rares and Vondrick, Carl},
  journal = {arXiv preprint arXiv:2508.00795},
  year = {2025},
  url = {https://arxiv.org/abs/2508.00795}
}

@inproceedings{liao2026genie,
  title = {{Genie Envisioner}: A Unified World Foundation Platform for Robotic Manipulation},
  author = {Liao, Yue and Zhou, Pengfei and Huang, Siyuan and Yang, Donglin and Chen, Shengcong and Jiang, Yuxin and Hu, Yue and Liu, Si and Luo, Jianlan and Chen, Liliang and Yan, Shuicheng and Yao, Maoqing and Ren, Guanghui},
  booktitle = {International Conference on Learning Representations (ICLR)},
  year = {2026},
  url = {https://arxiv.org/abs/2508.05635}
}

@inproceedings{ma2026dit4dit,
  title = {{DiT4DiT}: Jointly Modeling Video Dynamics and Actions for Generalizable Robot Control},
  author = {Ma, Teli and Zheng, Jia and Wang, Zifan and Jiang, Chunli and Cui, Andy and Liang, Junwei and Yang, Shuo},
  booktitle = {Advances in Neural Information Processing Systems (NeurIPS)},
  year = {2026},
  url = {https://arxiv.org/abs/2603.10448}
}

@inproceedings{nakamura2025latentsafety,
  author    = {Nakamura, Kensuke and Peters, Lasse and Bajcsy, Andrea},
  title     = {Generalizing Safety Beyond Collision-Avoidance via Latent-Space Reachability Analysis},
  booktitle = {Robotics: Science and Systems (RSS)},
  year      = {2025},
  doi       = {10.15607/RSS.2025.XXI.113},
  url       = {https://arxiv.org/abs/2502.00935}
}

@article{nvidia2025cosmos,
  title = {{Cosmos} World Foundation Model Platform for Physical {AI}},
  author = {{NVIDIA} and Agarwal, Niket and Ali, Arslan and Bala, Maciej and Balaji, Yogesh and Barker, Erik and Cai, Tiffany and Chattopadhyay, Prithvijit and Chen, Yongxin and Cui, Yin and Ding, Yifan and Dworakowski, Daniel and Fan, Jiaojiao and Fenzi, Michele and Ferroni, Francesco and Fidler, Sanja and Fox, Dieter and Ge, Songwei and Ge, Yunhao and Gu, Jinwei and Gururani, Siddharth and He, Ethan and Huang, Jiahui and Huffman, Jacob and Jannaty, Pooya and Jin, Jingyi and Kim, Seung Wook and Kl{\'a}r, Gergely and Lam, Grace and Lan, Shiyi and Leal-Taixe, Laura and Li, Anqi and Li, Zhaoshuo and Lin, Chen-Hsuan and Lin, Tsung-Yi and Ling, Huan and Liu, Ming-Yu and Liu, Xian and Luo, Alice and Ma, Qianli and Mao, Hanzi and Mo, Kaichun and Mousavian, Arsalan and Nah, Seungjun and Niverty, Sriharsha and Page, David and Paschalidou, Despoina and Patel, Zeeshan and Pavao, Lindsey and Ramezanali, Morteza and Reda, Fitsum and Ren, Xiaowei and Sabavat, Vasanth Rao Naik and Schmerling, Ed and Shi, Stella and Stefaniak, Bartosz and Tang, Shitao and Tchapmi, Lyne and Tredak, Przemek and Tseng, Wei-Cheng and Varghese, Jibin and Wang, Hao and Wang, Haoxiang and Wang, Heng and Wang, Ting-Chun and Wei, Fangyin and Wei, Xinyue and Wu, Jay Zhangjie and Xu, Jiashu and Yang, Wei and Yen-Chen, Lin and Zeng, Xiaohui and Zeng, Yu and Zhang, Jing and Zhang, Qinsheng and Zhang, Yuxuan and Zhao, Qingqing and Zolkowski, Artur},
  journal = {arXiv preprint arXiv:2501.03575},
  year = {2025},
  doi = {10.48550/arXiv.2501.03575},
  url = {https://arxiv.org/abs/2501.03575}
}

@article{nvidia2026cosmos3,
  title = {{Cosmos 3}: Omnimodal World Models for Physical {AI}},
  author = {{NVIDIA} and others},
  journal = {arXiv preprint arXiv:2606.02800},
  year = {2026},
  doi = {10.48550/arXiv.2606.02800},
  url = {https://arxiv.org/abs/2606.02800}
}

@inproceedings{pai2026mimicvideo,
  title = {{mimic-video}: Video-Action Models for Generalizable Robot Control Beyond {VLAs}},
  author = {Pai, Jonas and Achenbach, Liam and Sanchez, Oliver and Charalambous, Stefanos and Montesinos, Victoriano and Forrai, Benedek and Mees, Oier and Nava, Elvis},
  booktitle = {Robotics: Science and Systems (RSS)},
  year = {2026},
  doi = {10.15607/RSS.2026.XXII.077},
  url = {https://arxiv.org/abs/2512.15692}
}

@article{qiu2026vid2wam,
  title = {{Vid2WAM}: Distilling Video Diffusion Priors into World Action Models},
  author = {Qiu, Chenhao and Wang, Ruixiang and Zhao, Runyi and Lin, Sixu and Gu, Songen and Nan, Shufeng and Liu, Guiliang and Jia, Kui and Fu, Yanwei and Wu, Simo},
  journal = {arXiv preprint arXiv:2608.08558},
  year = {2026},
  doi = {10.48550/arXiv.2608.08558},
  url = {https://arxiv.org/abs/2608.08558}
}

@inproceedings{shen2025videovla,
  title = {{VideoVLA}: Video Generators Can Be Generalizable Robot Manipulators},
  author = {Shen, Yichao and Wei, Fangyun and Du, Zhiying and Liang, Yaobo and Lu, Yan and Yang, Jiaolong and Zheng, Nanning and Guo, Baining},
  booktitle = {Advances in Neural Information Processing Systems (NeurIPS)},
  year = {2025},
  url = {https://arxiv.org/abs/2512.06963}
}

@inproceedings{tao2024maniskill3,
  title = {Demonstrating {GPU} Parallelized Robot Simulation and Rendering for Generalizable Embodied {AI} with {ManiSkill3}},
  author = {Tao, Stone and Xiang, Fanbo and Shukla, Arth and Qin, Yuzhe and Hinrichsen, Xander and Yuan, Xiaodi and Bao, Chen and Lin, Xinsong and Liu, Yulin and Chan, Tse-kai and Gao, Yuan and Li, Xuanlin and Mu, Tongzhou and Xiao, Nan and Gurha, Arnav and Rajesh, Viswesh Nagaswamy and Choi, Yong Woo and Chen, Yen-Ru and Huang, Zhiao and Calandra, Roberto and Chen, Rui and Luo, Shan and Su, Hao},
  booktitle = {Robotics: Science and Systems (RSS)},
  year = {2025},
  url = {https://arxiv.org/abs/2410.00425}
}

@article{thananjeyan2021recoveryrl,
  title = {{Recovery RL}: Safe Reinforcement Learning With Learned Recovery Zones},
  author = {Thananjeyan, Brijen and Balakrishna, Ashwin and Nair, Suraj and Luo, Michael and Srinivasan, Krishnan and Hwang, Minho and Gonzalez, Joseph E. and Ibarz, Julian and Finn, Chelsea and Goldberg, Ken},
  journal = {IEEE Robotics and Automation Letters},
  volume = {6},
  number = {3},
  pages = {4915--4922},
  year = {2021},
  doi = {10.1109/LRA.2021.3070252},
  url = {https://arxiv.org/abs/2010.15920}
}

@inproceedings{tian2025seer,
  title = {Predictive Inverse Dynamics Models are Scalable Learners for Robotic Manipulation},
  author = {Tian, Yang and Yang, Sizhe and Zeng, Jia and Wang, Ping and Lin, Dahua and Dong, Hao and Pang, Jiangmiao},
  booktitle = {International Conference on Learning Representations (ICLR)},
  year = {2025},
  url = {https://arxiv.org/abs/2412.15109}
}

@article{wan2025wan,
  title = {{Wan}: Open and Advanced Large-Scale Video Generative Models},
  author = {{Team Wan} and Wang, Ang and Ai, Baole and Wen, Bin and Mao, Chaojie and Xie, Chen-Wei and Chen, Di and Yu, Feiwu and Zhao, Haiming and Yang, Jianxiao and Zeng, Jianyuan and Wang, Jiayu and Zhang, Jingfeng and Zhou, Jingren and Wang, Jinkai and Chen, Jixuan and Zhu, Kai and Zhao, Kang and Yan, Keyu and Huang, Lianghua and Feng, Mengyang and Zhang, Ningyi and Li, Pandeng and Wu, Pingyu and Chu, Ruihang and Feng, Ruili and Zhang, Shiwei and Sun, Siyang and Fang, Tao and Wang, Tianxing and Gui, Tianyi and Weng, Tingyu and Shen, Tong and Lin, Wei and Wang, Wei and Wang, Wei and Zhou, Wenmeng and Wang, Wente and Shen, Wenting and Yu, Wenyuan and Shi, Xianzhong and Huang, Xiaoming and Xu, Xin and Kou, Yan and Lv, Yangyu and Li, Yifei and Liu, Yijing and Wang, Yiming and Zhang, Yingya and Huang, Yitong and Li, Yong and Wu, You and Liu, Yu and Pan, Yulin and Zheng, Yun and Hong, Yuntao and Shi, Yupeng and Feng, Yutong and Jiang, Zeyinzi and Han, Zhen and Wu, Zhi-Fan and Liu, Ziyu},
  journal = {arXiv preprint arXiv:2503.20314},
  year = {2025},
  url = {https://arxiv.org/abs/2503.20314}
}

@inproceedings{wu2024gr1,
  title = {Unleashing Large-Scale Video Generative Pre-training for Visual Robot Manipulation},
  author = {Wu, Hongtao and Jing, Ya and Cheang, Chilam and Chen, Guangzeng and Xu, Jiafeng and Li, Xinghang and Liu, Minghuan and Li, Hang and Kong, Tao},
  booktitle = {International Conference on Learning Representations (ICLR)},
  year = {2024},
  url = {https://arxiv.org/abs/2312.13139}
}

@inproceedings{wu2025forewarn,
  author    = {Wu, Yilin and Tian, Ran and Swamy, Gokul and Bajcsy, Andrea},
  title     = {From Foresight to Forethought: {VLM}-In-the-Loop Policy Steering via Latent Alignment},
  booktitle = {Robotics: Science and Systems (RSS)},
  year      = {2025},
  doi       = {10.15607/RSS.2025.XXI.076},
  url       = {https://arxiv.org/abs/2502.01828}
}

@article{wu2026lingbotvla,
  title = {A Pragmatic {VLA} Foundation Model},
  author = {Wu, Wei and Lu, Fan and Wang, Yunnan and Yang, Shuai and Liu, Shi and Wang, Fangjing and Zhu, Qian and Sun, He and Wang, Yong and Ma, Shuailei and Ren, Yiyu and Zhang, Kejia and Yu, Hui and Zhao, Jingmei and Zhou, Shuai and Qiu, Zhenqi and Xiong, Houlong and Wang, Ziyu and Wang, Zechen and Cheng, Ran and Li, Yong-Lu and Huang, Yongtao and Zhu, Xing and Shen, Yujun and Zheng, Kecheng},
  journal = {arXiv preprint arXiv:2601.18692},
  year = {2026},
  url = {https://arxiv.org/abs/2601.18692}
}

@inproceedings{xu2025faildetect,
  author    = {Xu, Chen and Nguyen, Tony Khuong and Dixon, Emma and Rodriguez, Christopher and Miller, Patrick and Lee, Robert and Shah, Paarth and Ambrus, Rares and Nishimura, Haruki and Itkina, Masha},
  title     = {Can We Detect Failures Without Failure Data? {U}ncertainty-Aware Runtime Failure Detection for Imitation Learning Policies},
  booktitle = {Robotics: Science and Systems (RSS)},
  year      = {2025},
  doi       = {10.15607/RSS.2025.XXI.073},
  url       = {https://arxiv.org/abs/2503.08558}
}

@inproceedings{yan2026svam,
  title = {{S-VAM}: Shortcut Video-Action Model by Self-Distilling Geometric and Semantic Foresight},
  author = {Yan, Haodong and Zhong, Zhide and Zhu, Jiaguan and He, Junjie and Yuan, Weilin and Song, Wenxuan and Gong, Xin and Cai, Yingjie and Zhao, Guanyi and Yan, Xu and Liu, Bingbing and Chen, Ying-Cong and Li, Haoang},
  booktitle = {European Conference on Computer Vision (ECCV)},
  year = {2026},
  url = {https://arxiv.org/abs/2603.16195}
}

@article{yang2026abotm0,
  title = {{ABot-M0}: {VLA} Foundation Model for Robotic Manipulation with Action Manifold Learning},
  author = {Yang, Yandan and Zeng, Shuang and Lin, Tong and Chang, Xinyuan and Qi, Dekang and Xiao, Junjin and Liu, Haoyun and Chen, Ronghan and Chen, Yuzhi and Huo, Dongjie and Xiong, Feng and Wei, Xing and Ma, Zhiheng and Xu, Mu},
  journal = {arXiv preprint arXiv:2602.11236},
  year = {2026},
  doi = {10.48550/arXiv.2602.11236},
  url = {https://arxiv.org/abs/2602.11236}
}

@inproceedings{ye2026dreamzero,
  title = {World Action Models are Zero-shot Policies},
  author = {Ye, Seonghyeon and Ge, Yunhao and Zheng, Kaiyuan and Gao, Shenyuan and Yu, Sihyun and Kurian, George and Indupuru, Suneel and Tan, You Liang and Zhu, Chuning and Xiang, Jiannan and Malik, Ayaan Naveed and Lee, Kyungmin and Liang, William and Ranawaka Arachchige, Nadun and Gu, Jiasheng and Xu, Yinzhen and Wang, Guanzhi and Hu, Fengyuan and Narayan, Avnish and Bjorck, Johan and Wang, Jing and Kim, Gwanghyun and Niu, Dantong and Zheng, Ruijie and Xie, Yuqi and Wu, Jimmy and Wang, Qi and Julian, Ryan and Xu, Danfei and Du, Yilun and Chebotar, Yevgen and Reed, Scott and Kautz, Jan and Zhu, Yuke and Fan, Linxi and Jang, Joel},
  booktitle = {Conference on Robot Learning (CoRL)},
  year = {2026},
  url = {https://arxiv.org/abs/2602.15922}
}

@article{ye2026gigaworldpolicy,
  title = {{GigaWorld-Policy}: An Efficient Action-Centered World--Action Model},
  author = {Ye, Angen and Wang, Boyuan and Ni, Chaojun and Huang, Guan and Zhao, Guosheng and Li, Hao and Li, Hengtao and Li, Jie and Lv, Jindi and Liu, Jingyu and Cao, Min and Li, Peng and Deng, Qiuping and Mei, Wenjun and Wang, Xiaofeng and Chen, Xinze and Zhou, Xinyu and Wang, Yang and Chang, Yifan and Li, Yifan and Zhou, Yukun and Ye, Yun and Liu, Zhichao and Zhu, Zheng},
  journal = {arXiv preprint arXiv:2603.17240},
  year = {2026},
  doi = {10.48550/arXiv.2603.17240},
  url = {https://arxiv.org/abs/2603.17240}
}

@inproceedings{ye2026starvla,
  title = {{StarVLA-$\alpha$}: Reducing Complexity in Vision-Language-Action Systems},
  author = {Ye, Jinhui and Gao, Ning and Yang, Senqiao and Zheng, Jinliang and Wang, Zixuan and Chen, Yuxin and Chen, Pengguang and Chen, Yilun and Liu, Shu and Jia, Jiaya},
  booktitle = {European Conference on Computer Vision (ECCV)},
  year = {2026},
  pages = {351--369},
  doi = {10.1007/978-3-032-37467-7_20},
  url = {https://arxiv.org/abs/2604.11757}
}

@article{yuan2026fastwam,
  title = {{Fast-WAM}: Do World Action Models Need Test-time Future Imagination?},
  author = {Yuan, Tianyuan and Dong, Zibin and Liu, Yicheng and Zhao, Hang},
  journal = {arXiv preprint arXiv:2603.16666},
  year = {2026},
  url = {https://arxiv.org/abs/2603.16666}
}

@inproceedings{zhang2025dreamvla,
  title = {{DreamVLA}: A Vision-Language-Action Model Dreamed with Comprehensive World Knowledge},
  author = {Zhang, Wenyao and Liu, Hongsi and Qi, Zekun and Wang, Yunnan and Yu, Xinqiang and Zhang, Jiazhao and Dong, Runpei and He, Jiawei and Wang, He and Zhang, Zhizheng and Yi, Li and Zeng, Wenjun and Jin, Xin},
  booktitle = {Advances in Neural Information Processing Systems (NeurIPS)},
  year = {2025},
  url = {https://arxiv.org/abs/2507.04447}
}

@inproceedings{zhang2025safevla,
  title = {{SafeVLA}: Towards Safety Alignment of Vision-Language-Action Model via Constrained Learning},
  author = {Zhang, Borong and Zhang, Yuhao and Ji, Jiaming and Lei, Yingshan and Dai, Juntao and Chen, Yuanpei and Yang, Yaodong},
  booktitle = {Advances in Neural Information Processing Systems (NeurIPS)},
  year = {2025},
  doi = {10.52202/085713-5128},
  url = {https://arxiv.org/abs/2503.03480}
}

@inproceedings{zhang2026w2a,
  title = {{W2A}: Distilling World Priors into Action Models for Real-Time Robotic Control},
  author = {Zhang, Xiaopeng and Mu, Yongjin and Weng, Yueyang and Zhao, Chengzhi and Li, Yanjie},
  booktitle = {Conference on Robot Learning (CoRL)},
  year = {2026},
  url = {https://arxiv.org/abs/2611.02600}
}

@inproceedings{zhao2025cotvla,
  title = {{CoT-VLA}: Visual Chain-of-Thought Reasoning for Vision-Language-Action Models},
  author = {Zhao, Qingqing and Lu, Yao and Kim, Moo Jin and Fu, Zipeng and Zhang, Zhuoyang and Wu, Yecheng and Li, Zhaoshuo and Ma, Qianli and Han, Song and Finn, Chelsea and Handa, Ankur and Lin, Tsung-Yi and Wetzstein, Gordon and Liu, Ming-Yu and Xiang, Donglai},
  booktitle = {Proceedings of the IEEE/CVF Conference on Computer Vision and Pattern Recognition (CVPR)},
  year = {2025},
  url = {https://arxiv.org/abs/2503.22020}
}

@inproceedings{zheng2026contactguard,
  title = {{ContactGuard}: Pre-Contact Execution Monitoring with Action-Conditioned Latent World Models},
  author = {Zheng, Gehan and Johnson-Roberson, Matthew and Zhi, Weiming},
  booktitle = {Conference on Robot Learning (CoRL)},
  year = {2026},
  url = {https://arxiv.org/abs/2608.13438}
}

@article{zheng2026rewindil,
  title = {{Rewind-IL}: Online Failure Detection and State Respawning for Imitation Learning},
  author = {Zheng, Gehan and Seenivasan, Sanjay and Johnson-Roberson, Matthew and Zhi, Weiming},
  journal = {IEEE Robotics and Automation Letters},
  volume = {11},
  number = {11},
  pages = {13327--13334},
  year = {2026},
  doi = {10.1109/LRA.2026.3734897},
  url = {https://arxiv.org/abs/2604.16683}
}

@inproceedings{zhou2025codeasmonitor,
  author    = {Zhou, Enshen and Su, Qi and Chi, Cheng and Zhang, Zhizheng and Wang, Zhongyuan and Huang, Tiejun and Sheng, Lu and Wang, He},
  title     = {{Code-as-Monitor}: Constraint-aware Visual Programming for Reactive and Proactive Robotic Failure Detection},
  booktitle = {Proceedings of the IEEE/CVF Conference on Computer Vision and Pattern Recognition (CVPR)},
  pages     = {6919--6929},
  year      = {2025},
  url       = {https://arxiv.org/abs/2412.04455}
}

@inproceedings{zhu2025uwm,
  title = {Unified World Models: Coupling Video and Action Diffusion for Pretraining on Large Robotic Datasets},
  author = {Zhu, Chuning and Yu, Raymond and Feng, Siyuan and Burchfiel, Benjamin and Shah, Paarth and Gupta, Abhishek},
  booktitle = {Robotics: Science and Systems (RSS)},
  year = {2025},
  url = {https://arxiv.org/abs/2504.02792}
}

@article{starvla2026,
  title = {{StarVLA}: A {Lego}-like Codebase for Vision-Language-Action Model Developing},
  author = {{StarVLA Community}},
  journal = {arXiv preprint arXiv:2604.05014},
  year = {2026},
  url = {https://arxiv.org/abs/2604.05014}
}

@article{oquab2024dinov2,
  title = {{DINOv2}: Learning Robust Visual Features without Supervision},
  author = {Oquab, Maxime and Darcet, Timoth{\'e}e and Moutakanni, Th{\'e}o and Vo, Huy and Szafraniec, Marc and Khalidov, Vasil and Fernandez, Pierre and Haziza, Daniel and Massa, Francisco and El-Nouby, Alaaeldin and others},
  journal = {Transactions on Machine Learning Research},
  year = {2024},
  url = {https://arxiv.org/abs/2304.07193}
}

@inproceedings{zhang2018lpips,
  title = {The Unreasonable Effectiveness of Deep Features as a Perceptual Metric},
  author = {Zhang, Richard and Isola, Phillip and Efros, Alexei A. and Shechtman, Eli and Wang, Oliver},
  booktitle = {Proceedings of the IEEE/CVF Conference on Computer Vision and Pattern Recognition (CVPR)},
  year = {2018},
  url = {https://arxiv.org/abs/1801.03924}
}

@article{guo2026unified,
  title = {Unified {4D} World Action Modeling from Video Priors with Asynchronous Denoising},
  author = {Guo, Jun and Li, Qiwei and Li, Peiyan and Chen, Zilong and Sun, Nan and Su, Yifei and Wang, Heyun and Zhang, Yuan and Li, Xinghang and Liu, Huaping},
  journal = {arXiv preprint arXiv:2604.26694},
  year = {2026},
  url = {https://arxiv.org/abs/2604.26694}
}

@article{li2025worldeval,
  title = {{WorldEval}: World Model as Real-World Robot Policies Evaluator},
  author = {Li, Yaxuan and Zhu, Yichen and Wen, Junjie and Shen, Chaomin and Xu, Yi},
  journal = {arXiv preprint arXiv:2505.19017},
  year = {2025},
  url = {https://arxiv.org/abs/2505.19017}
}

@article{quevedo2025worldgym,
  title = {{WorldGym}: World Model as An Environment for Policy Evaluation},
  author = {Quevedo, Julian and Sharma, Ansh Kumar and Sun, Yixiang and Suryavanshi, Varad and Liang, Percy and Yang, Sherry},
  journal = {arXiv preprint arXiv:2506.00613},
  year = {2025},
  url = {https://arxiv.org/abs/2506.00613}
}

@article{guo2025ctrlworld,
  title = {{Ctrl-World}: A Controllable Generative World Model for Robot Manipulation},
  author = {Guo, Yanjiang and Shi, Lucy Xiaoyang and Chen, Jianyu and Finn, Chelsea},
  journal = {arXiv preprint arXiv:2510.10125},
  year = {2025},
  url = {https://arxiv.org/abs/2510.10125}
}

@inproceedings{liu2023libero,
  title = {{LIBERO}: Benchmarking Knowledge Transfer for Lifelong Robot Learning},
  author = {Liu, Bo and Zhu, Yifeng and Gao, Chongkai and Feng, Yihao and Liu, Qiang and Zhu, Yuke and Stone, Peter},
  booktitle = {Advances in Neural Information Processing Systems (NeurIPS), Datasets and Benchmarks Track},
  year = {2023},
  url = {https://arxiv.org/abs/2306.03310}
}

@inproceedings{xiang2020sapien,
  title = {{SAPIEN}: A {SimulAted} Part-based Interactive {ENvironment}},
  author = {Xiang, Fanbo and Qin, Yuzhe and Mo, Kaichun and Xia, Yikuan and Zhu, Hao and Liu, Fangchen and Liu, Minghua and Jiang, Hanxiao and Yuan, Yifu and Wang, He and Yi, Li and Chang, Angel X. and Guibas, Leonidas J. and Su, Hao},
  booktitle = {Proceedings of the IEEE/CVF Conference on Computer Vision and Pattern Recognition (CVPR)},
  year = {2020},
  url = {https://arxiv.org/abs/2003.08515}
}

@article{deepseekai2026v41flash,
  title = {{DeepSeek-V4.1-Flash}: Pushing the Limits of {KV} Cache Compression},
  author = {{DeepSeek-AI}},
  journal = {arXiv preprint arXiv:2609.19969},
  year = {2026},
  url = {https://arxiv.org/abs/2609.19969}
}

@inproceedings{shan2026dvg,
  title={{DVG-WM}: Disentangled video generation enables efficient embodied world model for robotic manipulation},
  author={Shan, Ziyu and Wu, Zhenyu and Wang, Xiaofeng and Zhu, Zheng and Wang, Ziwei},
  booktitle={European Conference on Computer Vision},
  pages={306--323},
  year={2026},
  organization={Springer}
}
